\documentclass{article}

\PassOptionsToPackage{numbers,compress}{natbib}

\usepackage[preprint]{conf_style}

\usepackage{multirow}

\usepackage[utf8]{inputenc} 
\usepackage[T1]{fontenc}    
\usepackage{hyperref}       
\usepackage{url}            
\usepackage{booktabs}       
\usepackage{amsfonts}       
\usepackage{nicefrac}       
\usepackage{microtype}      
\usepackage{xcolor}         
\usepackage{esvect} 
\usepackage{subcaption}
\usepackage{tikz}

\usepackage{algorithm}
\usepackage{algorithmic}
\usepackage[utf8]{inputenc} 
\usepackage[T1]{fontenc}    
\usepackage{hyperref}       
\usepackage{url}            
\usepackage{booktabs}       
\usepackage{amsfonts}       
\usepackage{nicefrac}       
\usepackage{microtype}      
\usepackage{xcolor}         
\usepackage{pifont}
\usepackage{amssymb}

\newcommand{\beeq}{\begin{equation}}
\newcommand{\eneq}{\end{equation}}
\usepackage{amsmath,amssymb}

\usepackage{bm}
\usepackage{natbib}
\usepackage{graphicx}
\usepackage{xcolor}
\usepackage{comment}

\usepackage{multirow}

\NewDocumentCommand{\grad}{e{_^}}{%
  \mathop{}\!
  \nabla
  \IfValueT{#1}{_{\!#1}}
  \IfValueT{#2}{^{#2}}
}

\newcommand\RR{\mathbb{R}}

\usepackage{todonotes}

\usepackage{multirow}

\def\bmzero{{\bm{0}}}

\newcommand{\bmI}{\bm{\mathrm{I}}}

\newcommand{\bu}{\boldsymbol{\mu}}

\newcommand{\bx}{\mathbf{x}}

\def\rvepsilon{{\bm{\epsilon}}}

\def\eqref#1{Eq.~(\ref{#1})}

\def\gN{{\mathcal{N}}}

\title{On the Constrained Time-Series Generation Problem}

\author{%
  Andrea Coletta\\
  J.P. Morgan AI Research\\
  London, UK\\
 \And
 Sriram Gopalakrishan\\
J.P. Morgan AI Research \\
New York, USA\\
\And
 Daniel Borrajo\\
J.P. Morgan AI Research \\
Madrid, ESP
\And
   Svitlana Vyetrenko\\
  J.P. Morgan AI Research\\
  New York, USA\\
}

\begin{document}

\maketitle

\begin{abstract}
Synthetic time series are often used in practical applications to augment the historical time series dataset for better performance of machine learning algorithms, amplify the occurrence of rare events, and also create counterfactual scenarios described by the time series. Distributional-similarity (which we refer to as realism) as well as the satisfaction of certain numerical constraints are common requirements in counterfactual time series scenario generation requests. For instance, the US Federal Reserve publishes synthetic market stress scenarios given by the constrained time series for financial institutions to assess their performance in hypothetical recessions.
Existing approaches for generating constrained time series usually penalize training loss to enforce constraints, and reject non-conforming samples. However, these approaches would require re-training if we change constraints, and rejection sampling can be computationally expensive, or impractical for complex constraints.
In this paper, we propose a novel set of methods to tackle the constrained time series generation problem and provide efficient sampling while ensuring the realism of generated time series.  
In particular, we frame the problem using a constrained optimization framework and then we propose a set of generative methods including ``GuidedDiffTime'', a guided  diffusion model to generate realistic time series. 
Empirically, we evaluate our work on several datasets for financial and energy data, where incorporating constraints is critical. We show that our approaches outperform existing work both qualitatively and quantitatively. Most importantly, we show that our ``GuidedDiffTime'' model is the only solution where re-training is not necessary for new constraints, resulting in a significant carbon footprint reduction, up to $92\%$ w.r.t. existing deep learning methods. 
\end{abstract}

\section{Introduction}
In recent years, synthetic time series (TS) have gained popularity in various applications such as data augmentation, forecasting, and imputation of missing values ~\cite{timeGAN,brophy2023generative,li2022generative,jordon2021hide,tashiro2021csdi}. Additionally, synthetic TS are extremely useful to generate unseen and counterfactual scenarios, where we can test hypotheses and algorithms before employing them in real settings~\cite{coletta2021towards}. For example, in financial markets, it can be very useful to test trading strategies on unseen hypothetical markets scenarios, as poorly tested algorithms can lead to large losses for investors, as well as to overall market instability~\cite{bouchaud2018trades,kirilenko2017flash}. In order to be useful, such hypothetical market stress scenarios need to be realistic - i.e., the synthetic market TS need to have statistical properties similar to the historical ones. They also need to satisfy certain constraints supplied by experts on how hypothetical market shock scenarios can potentially unfold. For instance, in order to ensure financial market stability, the US Federal Reserve annually assesses the market conditions and publishes a set of constrained market stress scenarios that financial institutions must subject their portfolios to, in order to estimate and adjust for their losses in case of market downturns~\cite{federalreserve}.

Our work targets the problem of generating constrained TS that are both statistically similar to historical times series and match a given set of constraints. These constraints can be imposed by the underlying physical process that generates the data. For example, synthetic energy data should adhere to the principle of 'energy conservation'~\cite{seo2021controlling}. Or, as in the preceding example of the US Federal Reserve stress scenarios, constraints can be used to generate counterfactual synthetic TS with some given conditions, e.g., a stock market index decreases by 5\%~\cite{federalreserve, kinlay2023synthetic}.

\textbf{Related work}: Existing work employs deep generative models (DGMs) to capture the statistical data properties and temporal dynamics of TS~\cite{timeGAN, jeon2022gt, seyfi2022generating, jarrett2021time}, and additional constraints are usually introduced by penalizing the generative model proportionally to the mass it allocates to invalid data~\cite{xu2018semantic,guimaraes2017objective,di2020efficient}; or by adding a regularization term to the loss function~\cite{ganchev2010posterior,takeishi2021knowledge}. Other approaches condition the generative process by encoding constraints and feeding them into the model~\cite{seo2021controlling}; or they reject and re-sample TS that do not match the constraints~\cite{patki2016synthetic}. Finally, a different line of work proposes special-purpose solutions, with ad-hoc architectures or sampling methods, which however tackle specific applications (not TS generation)~\cite{wang2021multi,torrado2020bootstrapping,xue2019embedding,de2018molgan}. In general, while most of these models are able to reproduce the real data statistics, complex constraints can still be challenging to be guaranteed. Most importantly, as DGMs incorporate constraints during training, a change to the constraints may require re-training since the learned distribution of a DGM may no longer cover the target distribution, and thus even rejection sampling would not be effective~\cite{peng2018advanced}.

\begin{figure}[t]
\centering
\subcaptionbox{Unconstrained}{\includegraphics[width=0.20\textwidth]{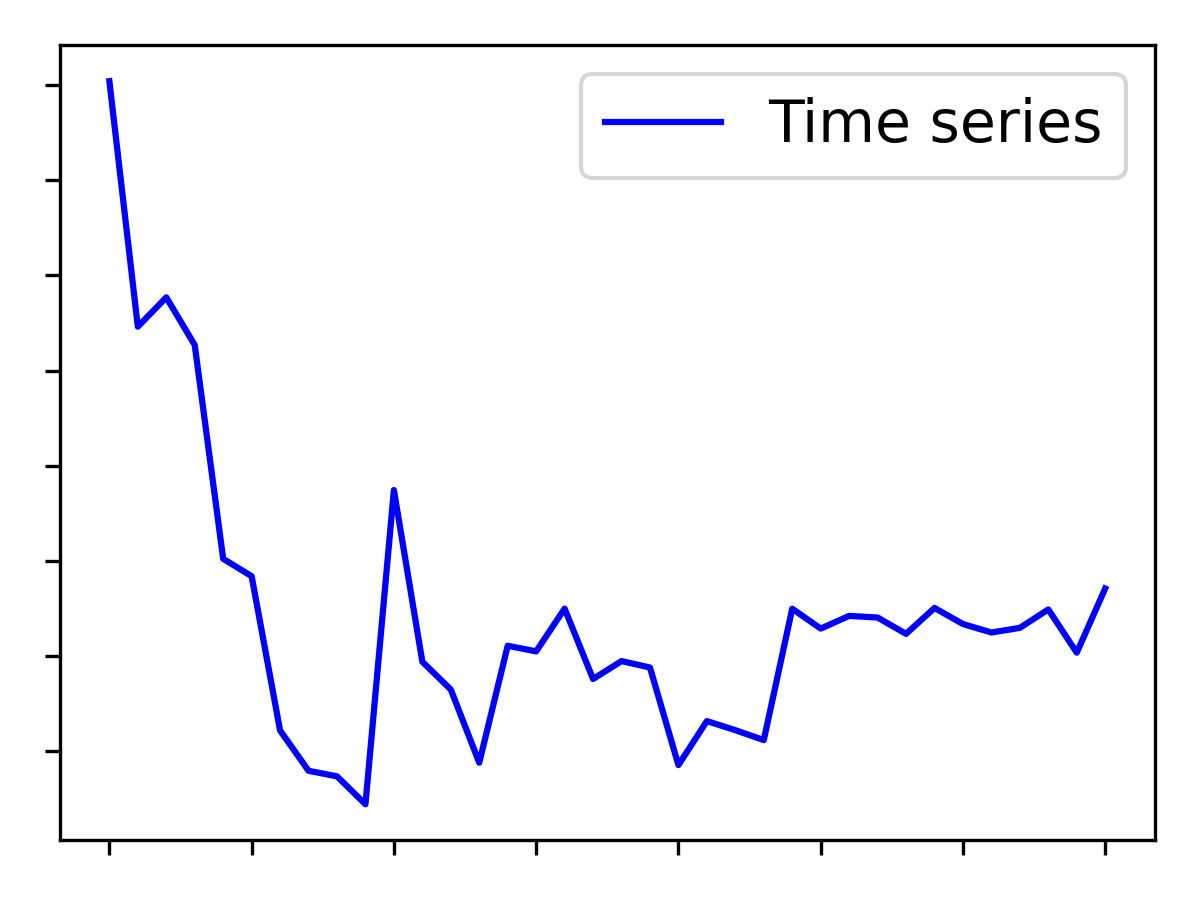}}
\hfill
\subcaptionbox{Trend}{\includegraphics[width=0.20\textwidth]{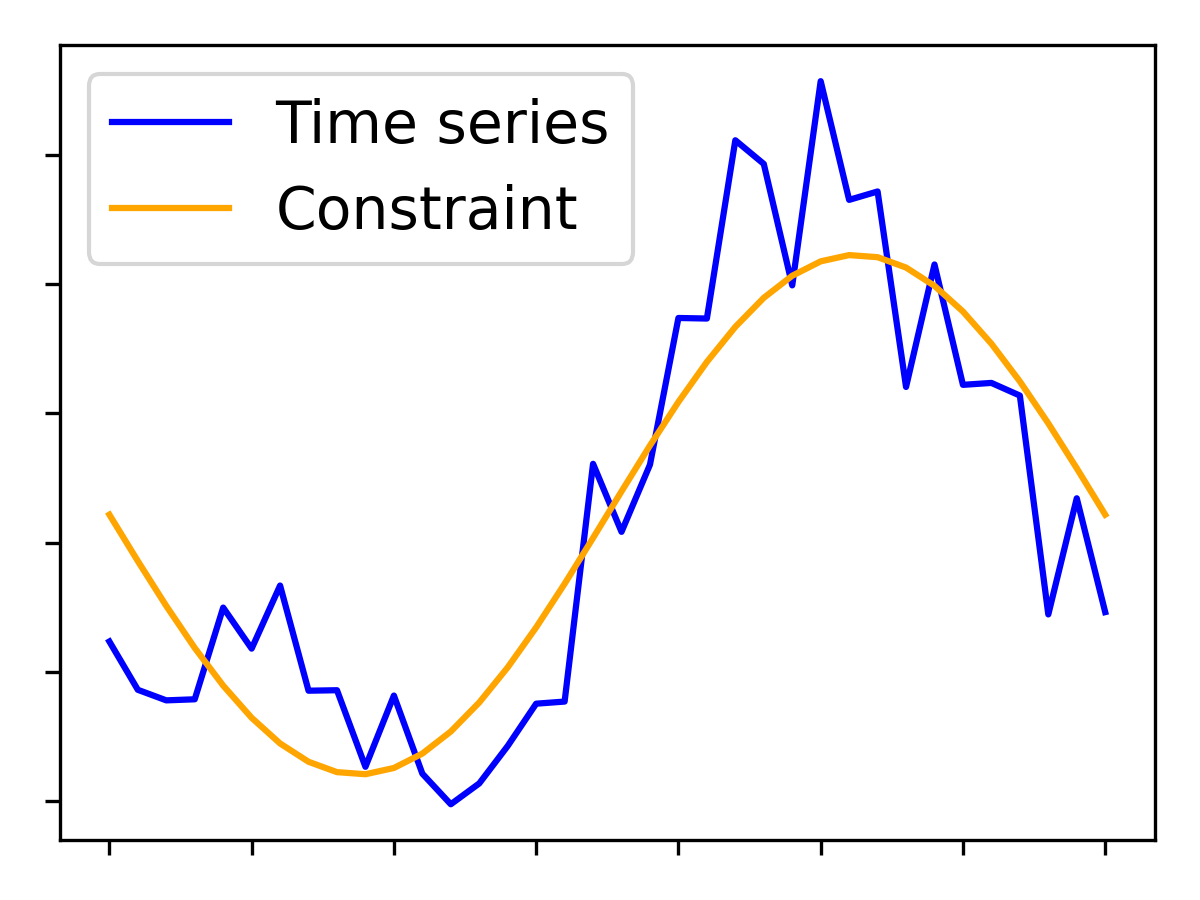}}
\hfill
\subcaptionbox{Fixed value}{\includegraphics[width=0.20\textwidth]{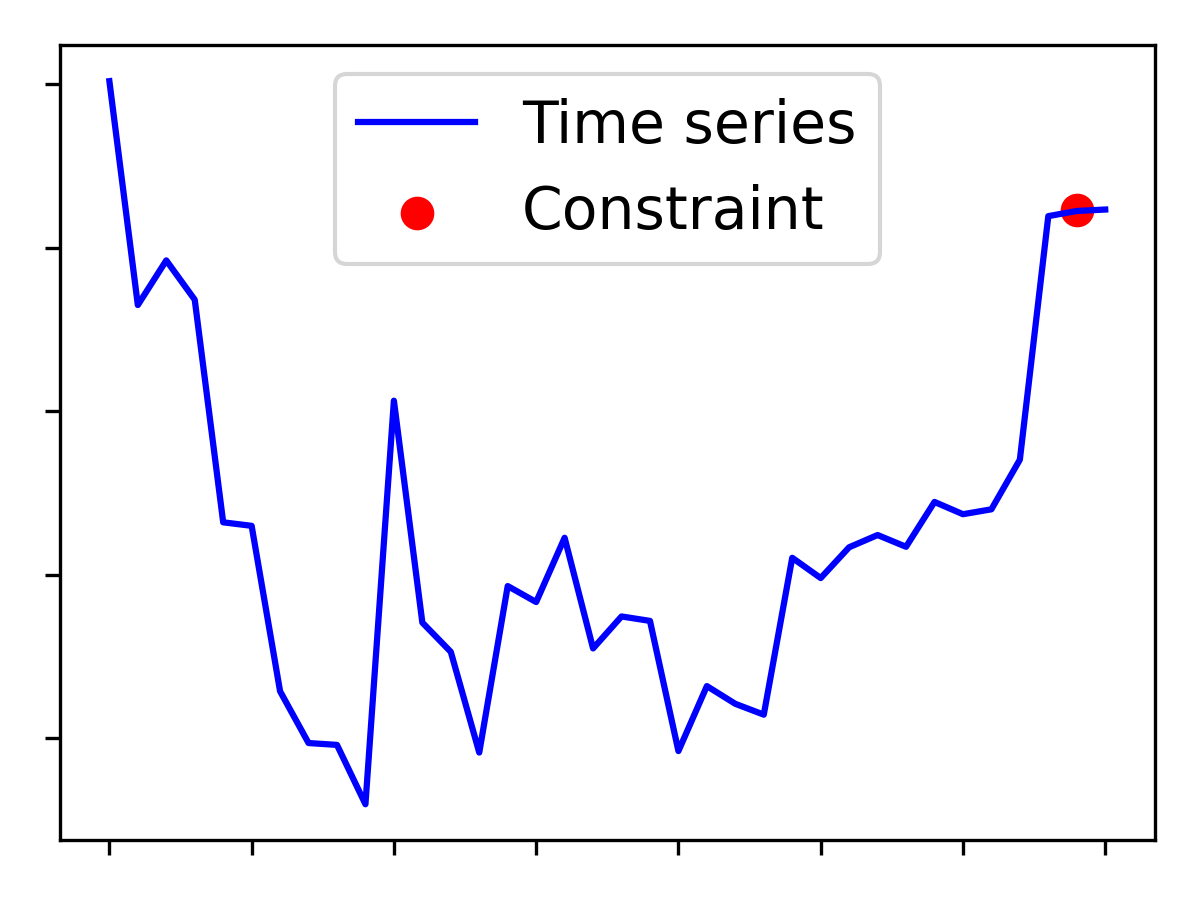}}
\hfill
\subcaptionbox{Global minimum}{\includegraphics[width=0.20\textwidth]{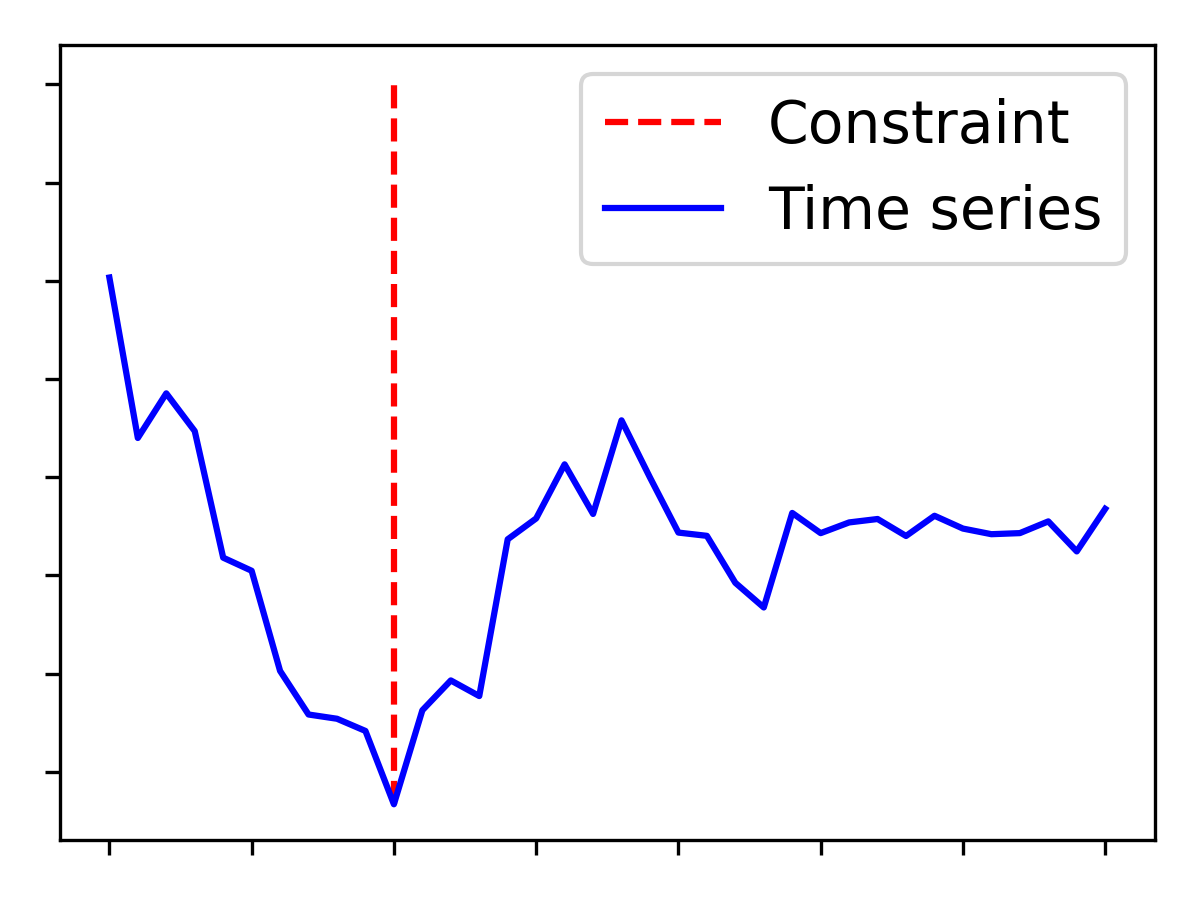}}
\hfill
\subcaptionbox{Multivariate}{\includegraphics[width=0.20\textwidth]{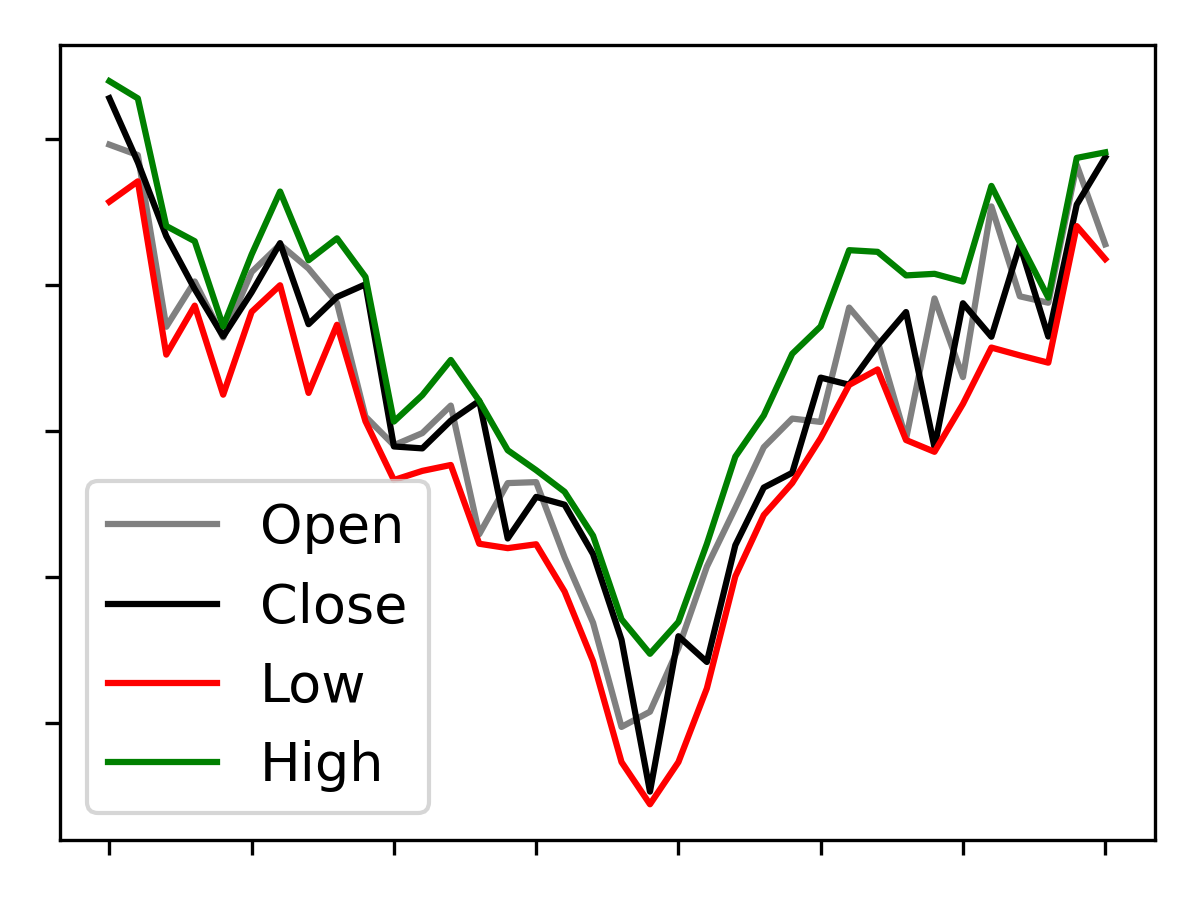}}
\caption{ An example of synthetic stock market time-series under different constraints: (a) unconstrained generation; (b) a time-series following a trend constraint; (c) the final value of the TS has to hold a specific value; (d) the global minimum must be at a given time point; (e) multi-variate TS where the \textit{High} and \textit{Low} dimensions have the maximum and minimum values, respectively.
}\label{fig:example_constraints}
\end{figure}

In this paper, we tackle the constrained TS generation problem for different constraint types, and we present and compare a set of methods that can incorporate constraints into TS generation. 
First, we present an optimization-based approach in which we compile the problem into a constrained optimization problem (\textbf{COP})~\cite{jorge2006numerical} and generate synthetic data from an explicit formulation of data properties and constraints. 
To overcome the need for an explicit definition of data properties, we employ a DGM that can implicitly capture data properties from training data. In particular, we introduce \textbf{DiffTime}, an approach based on conditional denoising diffusion models~\cite{ho2020denoising} where several constraints can condition the data generation.
In addition, we show that any kind of constraint can be applied to diffusion models by penalizing the model proportionally to the constraint violation during training. This approach, called \textbf{Loss-DiffTime}, shows good performance with efficient sampling, but requires re-training upon a new constraint. 
Finally, to increase computational efficiency and reduce the carbon footprint of the model~\cite{dhar2020carbon}, we propose a guided diffusion model \textbf{Guided-DiffTime} that does not require re-training upon changes in the constraints ---- at inference, it can adjust the TS generation  process based on the input constraints.

Our main contributions can be summarized as follows: 
\begin{itemize}
    \item We formally define the constrained TS problem, and characterize different types of constraints for synthetic TS.
\item We propose and compare different approaches to generate synthetic TS. We evaluate their advantages and disadvantages with respect to different measures of performance and constraint types. We show how COP can also be used for post-hoc fine-tuning of TS, such that synthetic TS generated by any DGM can be adjusted to guarantee constraint satisfaction.
\item We empirically demonstrate that our approaches outperform existing work in many TS domains, both qualitatively and quantitatively. We show that \textit{DiffTime} outperforms existing state-of-art models in unconstrained TS generation, while \textbf{\textit{Guided-DiffTime} is the only solution where re-training for new constraints is not necessary, resulting in a significant carbon footprint reduction.} 
\end{itemize}

\section{Definitions and Problem Formulation}\label{sec:prob_formulation}

The constrained TS generation problem requires generating synthetic TS data, where each TS is defined in the sample space $ \chi = \mathbb{R}^{L \times K}$ where $L$ is the length of the TS and $K$ is the number of features. Our goal is to generate synthetic data such that the synthetic distribution approximates the input data distribution, \textit{and} each TS also conforms to user-specified constraints. The problem input is a tuple  $\langle \mathbb{D} = \{\mathbf{x}^i\}_{i=1}^N,C\rangle$ consisting of a dataset $\mathbb{D}$ of $N$ time series $\bx^i\in\chi, i\in[1..N]$ and a list of constraints $C$ that a synthetic TS should conform to. The constraints include realism constraints (see Section~\ref{sec:cop}). Henceforth, we will drop the sample index for $\bx$ unless needed, and only keep the position and feature indices. We also shortly denote $[K] \triangleq \{0, \ldots,  K\}$ and $[L] \triangleq \{0, \ldots, L\}$.

Constraints --- like those in Figure \ref{fig:example_constraints} ---  are defined as tuples of the form $\langle t,f\rangle \in C$, where $t$ can be either {\it soft} or {\it hard} and a differentiable function $f$. If the constraint type is {\it hard}, then $f$ can be an inequality or an equality constraint. An inequality constraint is of the form $f(\hat{\bx}) \leq 0$ where $\hat{\bx}$ is the generated synthetic TS. An equality constraint is of the form $f(\hat{\bx}) = 0$. Hard constraints are required to hold in a generated TS. Otherwise, the TS is rejected. Soft constraints are of the form $f: \chi\rightarrow \mathbb{R}$ whose value we would like to optimize (minimize) for. Therefore, by definition, soft constraints do not require sample rejection. The constraints can be defined with respect to individual synthetic TS samples $\hat{\bx}$, or at the dataset level (distribution-related constraints). 
As a type of soft constraint, we define \textit{trend-lines} (see Figure~\ref{fig:example_constraints}.b) as a time series $\mathbf{s} \in \chi$. This constraint tells a generative method to minimize the L2 distance between the trend and the corresponding points of the synthetic TS. Formally, the synthetic TS $\hat{\bx}$ would be optimized as to minimize $\left\| \mathbf{s} - \hat{\bx} \right\|_{2}^{2}$. 

Additionally, both soft and hard constraints can be categorized into \textit{local} and \textit{global} constraints.  Global constraints are those that compare across all the points in the TS. For example, we can enforce that $x_{i,j} \leq x_{3,0}, \ \forall (i,j) \in [L] \times [K]$ such that the maximum value is at $x_{3,0}$. Local constraints are those that only refer only to a subset of points. For instance, requiring ($x_{i,j} = 2.5$) for a given point $(i,j) \in [L] \times [K]$. We refer to this kind of constraint as \textit{fixed-point constraints} (see Figure~\ref{fig:example_constraints}.c) since they require that the value of the TS is fixed at that point to a specific value. The set of all fixed-point constraints is $\mathcal{R}$, where each element $r_{i,j} \in \mathbb{R}$ and $(i,j) \in [L] \times [K]$.

The aforementioned types of constraints are explicit. Additionally, the problem of synthetic data generation requires statistical similarity between the input and the synthetic datasets, which can either be built-in into the data generating method (e.g., by design GANs  generate data that is distributionally similar to the input~\cite{brophy2023generative}) or specified explicitly as constraints in the model (e.g., autocorrelation similarity can be an explicit constraint).
The methods presented herein assume that the constraints are differentiable. This is needed for deep generative methods, such as diffusion models, where constraints need to be incorporated into the training or inference process. If the functions are differentiable, then a straightforward approach~\cite{Zico_DC3_hard_constraints_solver} to incorporate them into the loss is:
\begin{equation}
    loss(\bx, \hat{\bx}) = objective\_loss(\bx, \hat{\bx}) + \lambda_{g} \text{ReLU}(g(\hat{\bx})) + \lambda_{h} h(\hat{\bx})^2
    \label{eq:constraint_loss}
\end{equation}
where $g(\hat{\bx})$ and $h(\hat{\bx})$ are the inequality and an equality constraint respectively, which are added as soft constraints into the loss function with penalty terms $ \lambda_{g}$ and $ \lambda_{h}$. However, incorporating constraints into the loss function may not guarantee constraint-conforming solutions, but good candidate or starting solutions that we can then fine-tune (i.e., adjust to guarantee constraints). If the constraints are not differentiable, one can use approaches that compute the loss for such a ``rule'' using perturbations~\cite{seo2021controlling}.

\section{Constrained Time-Series Generation - Proposed Approaches}

In this section, we introduce several approaches to tackle the constrained TS generation problem. In particular, we discuss their advantages and disadvantages, and how they handle different scenarios and constraints.

\subsection{Constrained Optimization Problem (COP)}\label{sec:cop}

Our first model tackles the synthetic TS generation problem as a Constrained Optimization Problem (COP) in which we treat each point $x_{i,j} \in \bx$ as a decision variable to be optimized. We will refer to this method simply as ``\textit{COP-method}''. A COP problem is defined by an objective function that a solution is optimized for, and set of constraints that need to be satisfied by the solution.

We can use COP in two ways, as a generative method, and as a fine-tuning method. If COP is used for generating synthetic TS, then we take as input a real sample $\bx$ from $\mathbb{D}$ as the starting TS (i.e., seed) for generation, and set the objective to maximize the difference between the seed and the synthetic TS. Formally, we maximize the L2 norm of the TS difference: $obj(\bx,\hat{\bx}) = \left\| (\bx - \hat{\bx}) \right\|_2$, where $\bx$ is the seed TS and $\hat{\bx}$ is the generated TS. COP can also be used for fine-tuning candidate solutions generated by other methods, such as by a diffusion model. When using COP as a fine-tuning method, the candidate solution generated by the other method becomes the seed TS, and the objective simply changes from maximizing the L2 difference to minimizing it; this is to preserve the information from the candidate TS and just search in the space of nearby solutions for one that satisfies all the constraints (if any failed). This can be helpful in fixing almost correct solutions, rather than using rejection sampling. 

The constraints for the COP formulation come from $C$. Additionally, when using COP as a generative process we need to add constraints to satisfy the desired distributional properties that methods like GANs would implicitly handle, such as preserving the distribution of the autocorrelation of returns for stock data.
This is done by constraining the COP solver to try and match the desired statistical properties of the seed TS; by matching the property at the sample level, we seek to match the distribution of that property at the dataset level. 
We do this by computing the target value from the seed TS and comparing it with the corresponding value from the synthetic TS. Specifically, we constrain the COP to limit the magnitude of the error between the property value computed for $\hat{\bx}$ and $\bx$ within an allowed amount (a budget for error tolerance). This is done by using an inequality constraint as follows: $e(z(\hat{\bx}),z(\bx)) - b \leq 0$, where $b$ is the budget we set ($b=0.1$ in our experiments), $z(.)$ is the function that computes the desired property, and $e(.)$ is the error function that measures the error between the target and generated values.
For example, for autocorrelation of returns in stock data, the TS property is a vector, so the $e(.)$ is the L2-norm of the difference. If the COP solver cannot find a solution within the allowed error tolerance, we double the budget and repeat the process for up to a fixed number of $\eta$ repeats (we set $\eta=10$ in our experiments). In Appendix~\ref{app:wgan_cop}, we discuss how distributional constraints can be learned directly from the input data, by training a Wasserstein-GAN~\cite{goodfellow2020generative,arjovsky2017wasserstein} and using the GAN \textit{critic} in the objective function.

Once we define the objective function and constraints for COP, we can employ one of the many solvers available to compute the synthetic TS. In our experiments, we use the Sequential Least Squares Programming (SLSQP) solver~\cite{jorge2006numerical} in Scipy's optimization module~\cite{2020SciPy-NMeth}. 

\subsection{\textit{DiffTime} - Conditional Diffusion Model for Time Series generation}\label{sec:difftime}
In the previous section, we introduced using COP to generate synthetic TS while guaranteeing the input constraints and data properties. However, such COP problems may be non-linear, and solving a non-linear problem is in general difficult and computationally expensive, especially with multi-variate and long time-series (see Section~\ref{sec:exp}).  In this section, we introduce a conditional diffusion model named \textbf{DiffTime} that leverages the latest advancements in score-based diffusion models~\cite{tashiro2021csdi,rasul2021autoregressive,ho2020denoising,sohl2015deep} to generate synthetic TS. Our model can generate realistic TS and cope with \textit{Trend} and \textit{Fixed Points} constraints by conditioning the generative model.

\paragraph{Denoising diffusion models}
Denoising Diffusion models are latent variable models which are trained to generate samples by gradually removing noise --- denoising ---  from samples corrupted by Gaussian noise~\cite{ho2020denoising}. These models approximate a real data distribution $q(\bx_0)$ by learning a model distribution $p_{\theta}(\bx_0) := \int p_\theta(\bx_{0:T}) \, d\bx_{1:T}$,  where the latent variables $\bx_{1:T}$ are defined in the same space $\mathcal{X}$ of the sample $\bx_0$. The training follows: a \textit{forward process} that progressively adds noise to the sample $\bx_0$; and a \textit{reverse process} where the generative process gradually denoises a noisy observation. The forward process is described with the following Markov chain with Gaussian transitions parameterized by $\beta_{1:T}$:
\begin{equation}
        q(\bx_{1:T} \mid \bx_0) := \prod_{t=1}^{T} q(\bx_t \mid \bx_{t-1}), \ \ \ \ \ \ \ \ \ \  q(\bx_t \mid \bx_{t-1}) := \mathcal{N}\left(\sqrt{1-\beta_t} \bx_{t-1}, \beta_t \bmI\right) 
    \label{eq:forward}
\end{equation}
It admits the following close form $q(\bx_t \mid \bx_0) = \mathcal{N}(\bx_t ; \sqrt{\hat{\alpha}_t}\bx_0, (1-\hat{\alpha}_t)\bmI)$, where $\alpha_t := 1 - \beta_t$ and $\hat{\alpha}_t := \prod_{i=1}^{t} \alpha_i$, which allows sampling $\bx_t$ at any arbitrary diffusion step $t$. The generation is performed by the \textit{reverse process} defined as a Markov Chain starting at $p(\bx_T) = \mathcal{N}(\bx_T; \mathbf{0}, \mathbf{I})$:
\begin{equation}    
  p_\theta(\bx_{0:T}) := p(\bx_T)\prod_{t=1}^T p_\theta(\bx_{t-1}|\bx_t), \ \ \ \ \ \ \ \ \ \ 
  p_\theta(\bx_{t-1}|\bx_t) := \mathcal{N}(\bx_{t-1}; \bu_\theta(\bx_t, t), \boldsymbol{\Sigma}_\theta(\bx_t, t) )
\end{equation}

Following the formulation of Denoising Diffusion Probabilistic Models (DDPM)~\cite{ho2020denoising} we parameterize the \textit{reverse process} as follows:

\begin{equation}
  \bu_\theta(\bx_t,t)=\frac{1}{\sqrt{\hat{\alpha}_t}} \left(\bx_t-\frac{\beta_t}{\sqrt{1-\hat{\alpha}_t}} \rvepsilon_\theta (\bx_t,t) \right), \  \ \
 \  \boldsymbol{\Sigma}_\theta(\bx_t,t) = \sigma^2 \mathbf{I}, \  \text{where} \  \sigma^2 = \sqrt{\beta_t}
  \label{eq:mu}
\end{equation}
where $\rvepsilon_\theta$ is a trainable denoising function that predicts $\rvepsilon$ from $\bx_t$, and the choice of $\beta$ corresponds to the upper bound on the reverse process entropy~\cite{sohl2015deep}. This function is approximated through a deep neural network trained according to the following objective:   
\begin{equation}    
 L(\theta) := \mathbb{E}_{t, \bx_0, \rvepsilon}{ \left\| \rvepsilon - \rvepsilon_\theta(\sqrt{\hat{\alpha}_t} \bx_0 + \sqrt{1-\hat{\alpha}_t}\rvepsilon, t) \right\|^2} \label{eq:training_objective_simple}
\end{equation}
where $t$ is uniformly sampled between 1 and $T$, and the noise is Gaussian $\rvepsilon \sim \mathcal{N}(0, \boldsymbol{I})$. The diffusion steps $T$ and variances $\beta_t$ control the expressiveness of the diffusion process and they are important hyperparameters to guarantee that the forward and reverse processes have the same functional form~\cite{sohl2015deep}. 

\paragraph{Conditional diffusion models}
Our \textit{DiffTime} model -- which is a conditional diffusion model --- supports both trend and fixed point constraints that were defined in Section~\ref{sec:prob_formulation}.  To constrain a particular trend, we condition the diffusion process using a trend TS $\mathbf{s}\in\chi$. Following recent work on conditional diffusion models~\cite{tashiro2021csdi}, we define the following model distribution:
\begin{equation}
    p_\theta(\bx_{0:T} | \mathbf{s}) := p(\bx_T) \prod_{t=1}^{T} p_\theta(\bx_{t-1} | \bx_t, \mathbf{s}), \ \ \ \, p_\theta(\bx_{t-1} \mid \bx_t, \mathbf{s}) := \mathcal{N}(\bx_{t-1}; \bu_\theta(\bx_t, t | \mathbf{s}), \boldsymbol{\Sigma}_\theta(\bx_t, t | \mathbf{s})).
\label{eq:cond_eq}
\end{equation}
which we learn by extending the parametrization in Eq.~\ref{eq:mu} with a conditional denoising function $\rvepsilon_{\theta}$:
\begin{equation}
  \bu_\theta(\bx_t,t \mid \mathbf{s}) = \frac{1}{\sqrt{\hat{\alpha}_t}} \left(\bx_t-\frac{\beta_t}{\sqrt{1-\hat{\alpha}_t}} \rvepsilon_\theta(\bx_t,t \mid \mathbf{s}) \right)
  \label{eq:cond_mu}
\end{equation}
where the $\boldsymbol{\Sigma}_\theta(\bx_t,t \mid \mathbf{s}) = \sigma^2 \mathbf{I}$. In this formulation, the trend is provided during each diffusion step $t$, without any noise added to the conditioning trend. During the training, we extract the trend $\mathbf{s}$ directly from the input TS $\bx_0$, which can be a simple linear or polynomial interpolation; during inference, the trend can be defined by the user at inference time. We recall that this is a \textit{soft constraint}, meaning that we do not expect the generated TS to exactly retrace the trend. In particular, during training, we provide a trend that is a low-order polynomial approximation of $\bx_0$ to avoid the model from copying the trend $\mathbf{s}$. Figure~\ref{fig:example_constraints}.b shows an example of the trend constraint.

Thus, the \textit{DiffTime} training procedure minimizes the following revised loss function: 
\begin{equation}
 L(\theta) := \mathbb{E}_{t, \bx_0, \rvepsilon}{ \left\| \rvepsilon - \rvepsilon_\theta(\sqrt{\hat{\alpha}_t} \bx_0 + \sqrt{1-\hat{\alpha}_t}\rvepsilon, t \mid \mathbf{s}) \right\|^2}      
 \label{eq:constr_diff_time} 
\end{equation}

\paragraph{Fixed Points.} To satisfy the \textit{fixed point} constraints, which are hard constraints, we modify the \textit{reverse process} of \textit{DiffTime} to explicitly include them in the latent variables $\bx_{1:T}$. We recall that $\mathcal{R}$ is the set of fixed point constraints, such that a fixed point constraint $r_{i,j} \in \mathcal{R}$ with $(i,j) \in [L] \times [K]$.
Thus, at each diffusion step $t$ we explicitly enforce the fixed-points values in the noisy time-series $\bx_t$, such that
$\forall \ r_{i,j} \in \mathcal{R}, \ x_{i,j} = r_{i,j}$ where $x_{i,j} \in \bx_t$. This approach would guarantee that the generated TS have the desired fixed-point values. Most importantly, we experimentally validated that the forward process generates consistent neighboring points (around the constrained fixed-points) which means that the synthetic samples are conditioned by the fixed points, and preserve the realism of the original input data.   
During training, we \textit{randomly} sample the \textit{fixed points} from the input TS($\bx_0$) and require the diffusion process to conform to those fixed points. At inference, the \textit{fixed points} can be provided by the user. Figure~\ref{fig:example_constraints}.c shows an example of a fixed point at the end of the TS, where the TS adapts to deal with the fixed point. 

In the Appendix, we provide additional details, network architecture, and the algorithm pseudo-codes.

\subsection{\textit{Loss-DiffTime} - Constrained generation with diffusion models}\label{sec:lossdiff_time}
In \textit{DiffTime}, we leverage conditional diffusion models to support trend and fixed values for generating TS. However, just by conditioning the model generation is not possible to encode all the constraints. A common solution is to penalize the generative model proportionally to how much the generated TS violates the input constraint~\cite{bengio2017deep}.

In this section, we propose \textit{Loss-DiffTime} where a \textit{constraint penalty} is applied to deal with any kind of constraint. The penalty function $f_{c}: \mathcal{X} \rightarrow \RR$ is added to the learning objective of the diffusion model, and it evaluates whether the generated TS $\hat{\bx}$ meets the input constraint. We discuss the penalty function $f_{c}$ for constraints in Section \ref{sec:prob_formulation} and in Equation \ref{eq:constraint_loss}. With $f_{c}$ in the loss, the greater the constraint violation is, the greater the model loss during training will be. However, the optimization problem in Eq.~\ref{eq:training_objective_simple} predicts the noise component for the sample $\bx_0$, into which we cannot directly feed to our penalty function. Moreover, we cannot apply $f_c(\bx)$ to a noisy sample $\bx_t$ as the constraints may be evaluated only on the final sample. Therefore, to apply our penalty function, we re-parametrize the optimization problem and force the diffusion model to explicitly model the final sample $\hat{\bx}_0$ at every step as follows:
\begin{equation}
 L(\theta) := \mathbb{E}_{t, \bx_0, \rvepsilon} \left[ { \left\| \rvepsilon - \rvepsilon_\theta(\bx_t, t \mid s)  \right\|^2} + \rho f_{c}(\hat{\bx}_0) \right] \
 \label{eq:loss_diff_time}
\end{equation}
where $\bx_t = \sqrt{\hat{\alpha}_t} \bx_0 + \sqrt{1-\hat{\alpha}_t}\rvepsilon$ and $\hat{\bx}_0 = \frac{1}{\sqrt{\alpha_t}}\left( \bx_t - \frac{1-\alpha_t}{\sqrt{1-\hat{\alpha_t}}}  \rvepsilon_\theta(\bx_t, t) \right)$. We consider that any constraint in $C$ can be differentiable (as discussed in Section \ref{sec:prob_formulation}). So, we can train our diffusion model following Eq.~\ref{eq:loss_diff_time} where $\rho$ is a scale parameter used to adjust the importance of the constraint loss. The conditional information of the trend  $\mathbf{s}$ can be removed if we do not need to enforce any trend constraint. Figure~\ref{fig:example_constraints}.d and Figure~\ref{fig:example_constraints}.e show two examples of more complex constraints with \textit{Loss-DiffTime}. 

\subsection{\textit{Guided-DiffTime} - Guided Diffusion models for constrained generation}
The \textit{Loss-DiffTime} model is now able to generate real TS while dealing with any constraint. However, we notice two major drawbacks: 1) since we translate constraints to penalty terms in the loss, we need to re-train the model for new constraints; and 2) the diffusion models usually require several iterative steps $T$ which can make it slower and expensive for TS generation.     
Our final proposed approach, namely \textit{Guided-DiffTime}, solves these two problems and can dramatically reduce the carbon footprint when using DGM for constrained TS generation. In particular, it adopts a Denoising Diffusion Implicit Model (DDIM)~\cite{song2020denoising} which requires fewer diffusion steps at inference. Moreover, by following the groundbreaking work of~\cite{dhariwal2021diffusion,song2020score}, which shows how to guide a diffusion model using a noisy classifier, we demonstrate how a pre-trained diffusion model can be guided (conditioned) using gradients from differentiable constraints. 

DDIM is a class of non-Markovian diffusion processes with the same training objective of classic DDPMs~\cite{ho2020denoising}, but fewer diffusion steps to generate high-quality samples. 
In particular, DDIMs keep the same training procedure as DDPMs defined in Section~\ref{sec:difftime} while the sampling can be accelerated by using the following re-parametrization of the \textit{reverse process}:
\begin{equation}    
    \bx_{t-1} = \sqrt{\hat{\alpha}_{t-1}} {\left(\frac{\bx_t - \sqrt{1 - \hat{\alpha}_t} \cdot \epsilon_\theta(\bx_t, t)}{\sqrt{\hat{\alpha}_t}}\right)} + {\sqrt{1 - \hat{\alpha}_{t-1} - \sigma_t^2} \cdot \epsilon_\theta(\bx_t, t)} + {\sigma_t \epsilon} 
    \label{eq:sample-eq-gen}
\end{equation}
where $\hat{\alpha}_0 :=1$ and different parametrizations of $\sigma_t$ lead to different generative processes. We set $\sigma_t = 0, \forall t \in [0, T]$ to have a deterministic forward process from latent variables to the sample $\bx_0$ (since the noise term $\sigma_t \epsilon$ is zeroed out). This deterministic forward process defines the DDIM which can use fewer diffusion steps to generate realistic samples. This diffusion steps are defined by a sequence $\tau$ of length $V$ which is a sub-sequence of $[1, \ldots, T]$ with the last value as $T$, i.e., $\tau_V = T$~\cite{song2020denoising}. For example, $\tau=[1,4,9,...,T]$. Moreover, this parametrization is a generalization of DDPM as setting $\sigma_t = \sqrt{(1 - \alpha_{t-1}) / (1 - \alpha_t)} \sqrt{1 - \alpha_t / \alpha_{t-1}}$ describes the original DDPM~\cite{song2020denoising} and the DDIM work showed that re-training of the DDPM model is unnecessary when we change the value of $\sigma$ or the diffusion steps $\tau$.

\begin{algorithm}[hbt]
    \caption{\textit{Guided-DiffTime}}
    \label{alg:guidingddim}
    \begin{algorithmic}
        \STATE Input: differentiable constraint $f_c : \mathcal{X} \rightarrow \RR$, scale parameter $\rho$
        \STATE Output: new TS, $\bx_0$
        \STATE $\bx_T \gets \text{sample from } \mathcal{N}(0, \mathbf{I})$
        \FORALL{$t$ from $T$ to 1}
            \STATE $\hat\epsilon \gets \epsilon_{\theta}(\bx_t, t)$
            \STATE $\hat\epsilon \gets \hat\epsilon - \rho \sqrt{1-\hat{\alpha}_t} \grad_{\bx_t} f_c(\frac{1}{\sqrt{\hat{\alpha}_t}}(\bx_t - \hat\epsilon \sqrt{ 1 - \hat{\alpha}_t}))$
            \STATE $\bx_{t-1} \gets \sqrt{\hat{\alpha}_{t-1}} \left( \frac{\bx_t - \sqrt{1-\hat{\alpha}_t} \hat{\epsilon}}{\sqrt{\hat{\alpha}_t}} \right) + \sqrt{1-\hat{\alpha}_{t-1}} \hat{\epsilon}$
        \ENDFOR
        \RETURN $\bx_0$
    \end{algorithmic}
\end{algorithm}

Given the DDIM, we can then apply the recent results from guided diffusion models~\cite{dhariwal2021diffusion,song2020score} to condition each sampling step with the information given by the gradients of the differentiable constraint $f_c$ (see Section~\ref{sec:lossdiff_time}). Algorithm~\ref{alg:guidingddim} shows the sampling procedure which computes the gradients w.r.t. to the input TS $\bx_t$. We recall that the constraint is applied on the final sample $\hat{\bx}_0$, computed according to the DDIM reverse process. Again, this approach does not require re-training of the original diffusion model to deal with new constraints, which can be applied just at inference time. Hence, we reduce the carbon footprint of the model, and get a faster time-series generation.

\section{Experiments}\label{sec:exp}
In this section we evaluate our approaches, showing their advantages and disadvantages when applied to different domains and constraints. In particular, we follow the five scenarios shown in Figure~\ref{fig:example_constraints} while considering multiple real-world and synthetic datasets. For COP we use a subset of the original TS as starting solution (seed), we leave in Appendix~\ref{sec:cop_initial_seed} the analysis of different seeds.

\paragraph{Baselines} We compare our approaches against existing TS generative models, including GT-GAN\cite{jeon2022gt}, TimeGAN~\cite{timeGAN}, RCGAN~\cite{esteban2017real},  C-RNN-GAN~\cite{mogren2016c}, a Recurrent Neural Networks (RNN)~\cite{timeGAN} trained with T-Forcing and P-Forcing~\cite{lamb2016professor,graves2013generating}, WaveNET~\cite{oord2016wavenet}, and WaveGAN~\cite{donahue2018adversarial}.
For the constrained scenarios, we extend the benchmark architectures to cope with constraints, by introducing a penalty loss~\cite{di2020efficient,xu2018semantic} or by conditioning the generation process.
We also employ \textit{rejection-sampling} and \textit{fine-tuning} with COP on their generated synthetic TS. 

\paragraph{Datasets} We consider three datasets with different characteristics such as periodicity, noise, correlation, and number of features: \textbf{1) daily stocks} which uses daily historical Google stock data from 2004 to 2019 with \textit{open, high, low, close, adjusted close, and volume} features~\cite{timeGAN}; 
\textbf{2) energy data} from the UCI Appliances energy prediction dataset~\cite{candanedo2017data} containing 28 features with noisy periodicity and correlation; \textbf{3) sines} a synthetic multivariate sinusoidal TS with different frequencies and phases~\cite{timeGAN}.

\paragraph{Evaluation metrics} For each experimental scenario, we evaluate the generative models and TS along different quantitative and qualitative dimensions. First, we evaluate the \textbf{realism} through a \textit{discriminative score}~\cite{timeGAN}, which measures how much the generated samples resemble (i.e., are indistinguishable from) the real data using a post-hoc RNN trained to distinguish between real and generated samples. We evaluate the \textbf{distributional-similarity} between the synthetic data and real data by applying t-SNE~\cite{van2008visualizing} on both real and synthetic samples; t-SNE shows (in a 2-dimensional space) how well the synthetic distribution covers the original input distribution. Then, we evaluate the \textbf{usefulness} of generated samples --- how the synthetic data supports a downstream task such as prediction ---- by training an RNN on synthetic data and testing its prediction performance on real data (i.e., \textit{predictive-score}~\cite{timeGAN}). 
To evaluate how the model satisfies different constraints, we introduce the following metrics: \textbf{Perc. error distance} which measures how much the synthetic data follows a trend constraint by evaluating the L2 distance between the TS and the trend; \textbf{satisfaction rate} which measures the percentage of time a synthetic TS meets the input constraints; the \textbf{inference time} measured as the average seconds required to generate a new sample with a given constraint; and finally the \textbf{fine-tuning time} which is the average time, in seconds, needed to enforce constraints over a generated sample, using COP to fine-tune it.

We provide further experiments, including details on the baselines, datasets, metrics, and algorithm hyperparameters in the Appendix.

\subsection{Unconstrained Generation}
\begin{figure}[hbt]
\centering
\subcaptionbox{\textbf{COP}}{\includegraphics[width=0.16\textwidth]{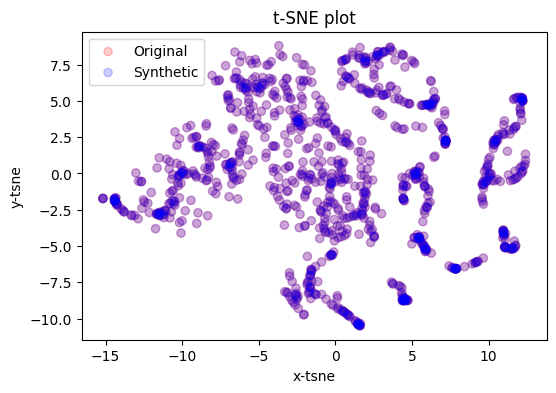}}%
\hfill
\subcaptionbox{\textbf{DiffTime}}{\includegraphics[width=0.16\textwidth]{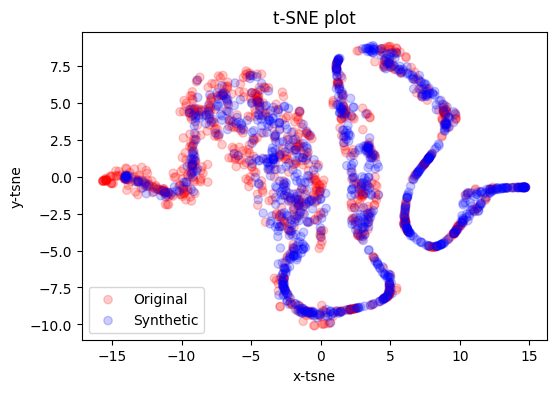}}%
\hfill
\subcaptionbox{GT-GAN}{\includegraphics[width=0.175\textwidth]{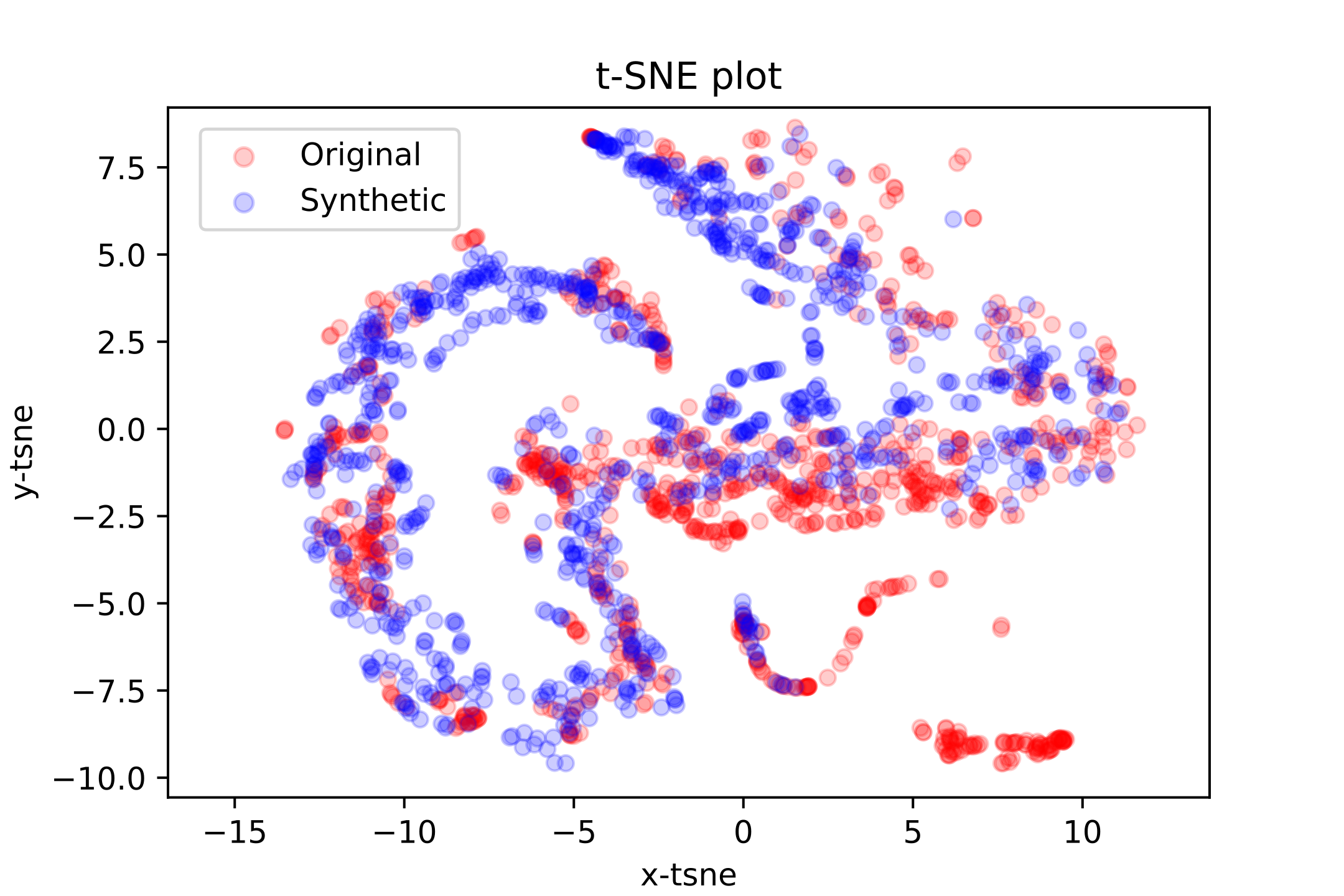}}%
\hfill
\subcaptionbox{TimeGAN}{\includegraphics[width=0.16\textwidth]{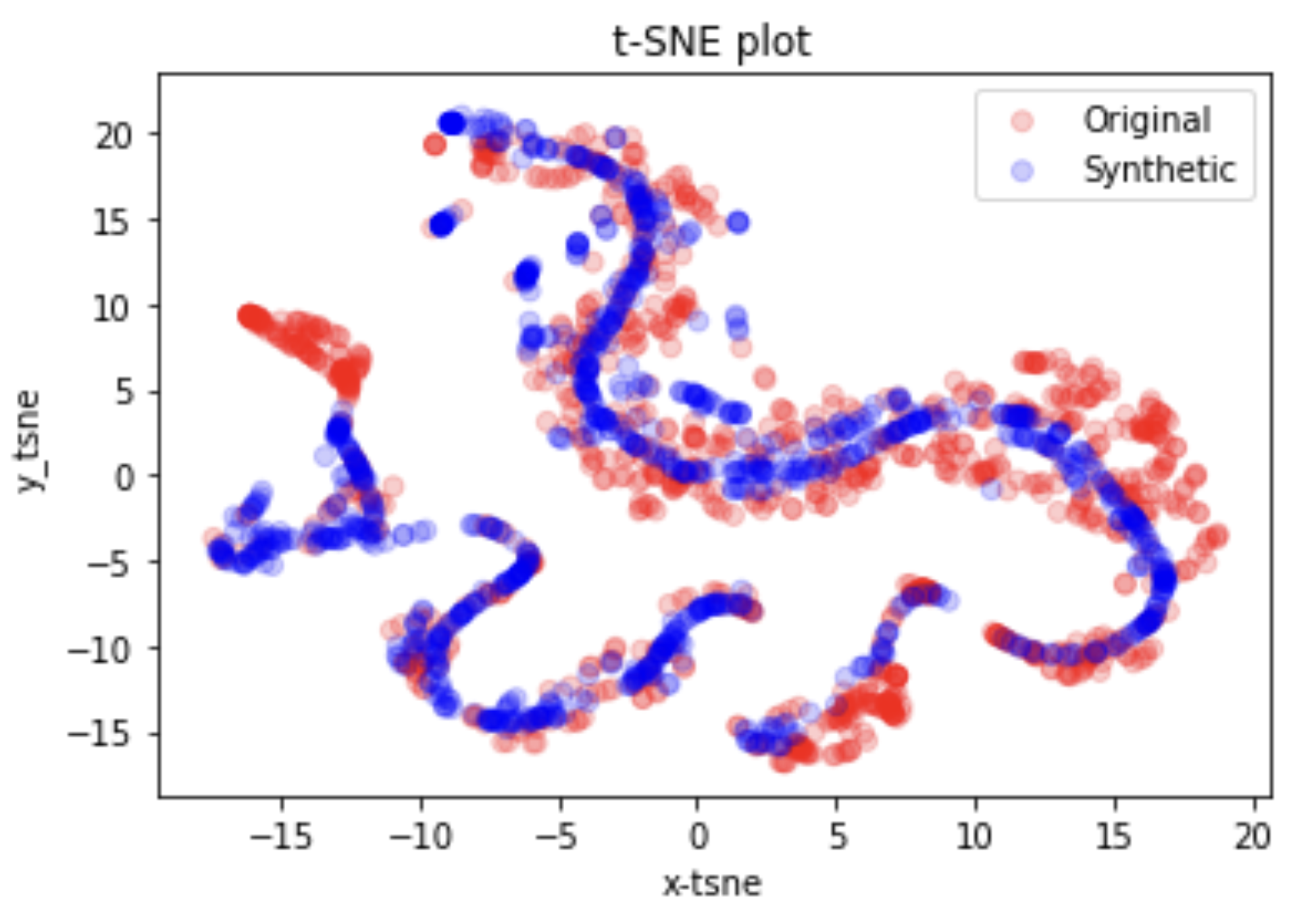}}%
\hfill
\subcaptionbox{RCGAN}{\includegraphics[width=0.16\textwidth]{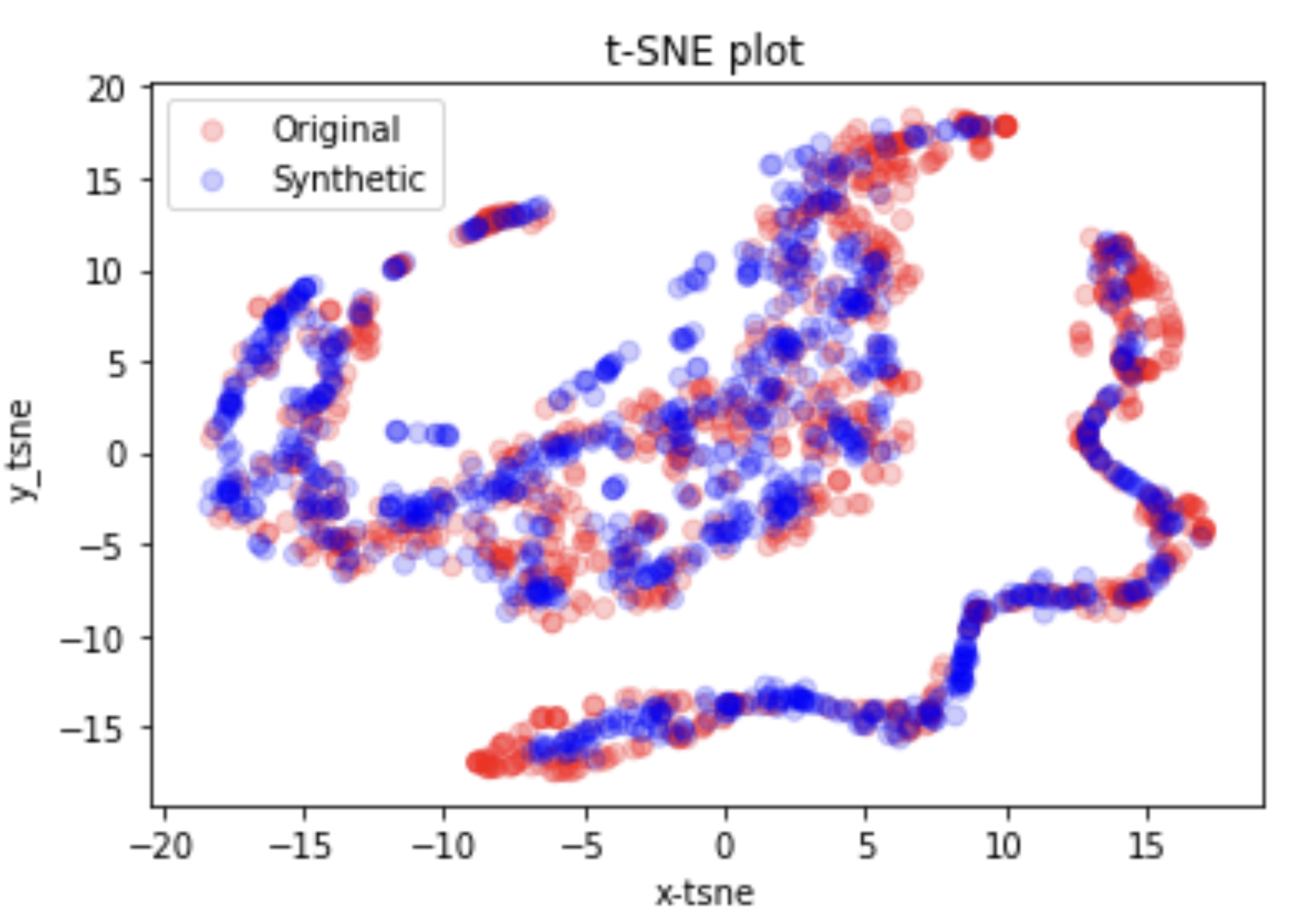}}%
\hfill
\subcaptionbox{C-RNN-GAN}{\includegraphics[width=0.16\textwidth]{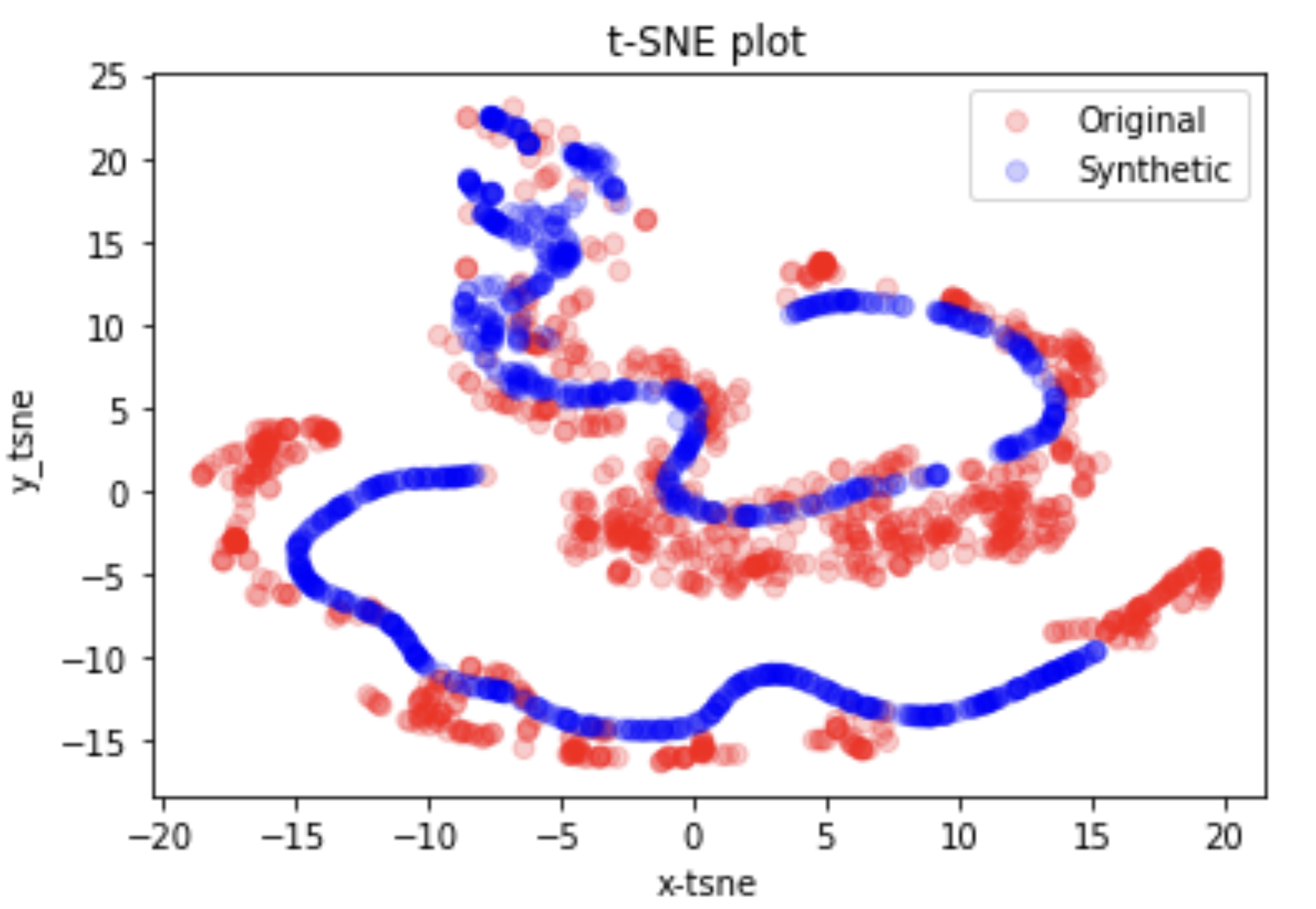}}%
\caption{t-SNE visualizations on multivariate stock data, where a greater overlap of blue and red dots shows a better distributional-similarity between the generated data and original data. Our approaches show the best performance.}\label{fig:tsne}
\end{figure}

First, we compare the ability of \textit{DiffTime} and \textit{COP} to generate unconstrained TS against existing benchmark datasets and algorithms. In Figure~\ref{fig:tsne} we evaluate the realism with respect to the distributional-similarity, where red dots represent the original TS and blue dots the generated TS. The figure shows that our approaches have significantly better performance with better overlap between red and blue samples.\footnote{We report only the top 6 models, leaving the full evaluation to Appendix~\ref{sec:unconstrained_append}.} 

In Table~\ref{tab:unconditional} we measure the \textit{usefulness} and \textit{realism} through the predictive and discriminative scores, respectively. \textit{DiffTime} consistently generates the most useful data with the best predictive score for both Sines and Stocks datasets, while keeping remarkable realism (i.e., discriminative score). \textit{COP} generates excellent synthetic samples as well, indistinguishable from real data with the best discriminative score for both Stocks and Energy. However, we acknowledge COP is advantaged by the original time series as an input seed.

\begin{table}[h]
\centering
\caption{Unconstrained Time-Series Generation (Bold indicates best performance). }\label{tab:unconditional}
\resizebox{0.7\linewidth}{!}{
\footnotesize
\begin{tabular}{|c|c|l|l|l|}
\hline
\multicolumn{1}{|c|}{ Metric}                                                                            & \multicolumn{1}{l|}{Method}    & \multicolumn{1}{l|}{Sines} & \multicolumn{1}{l|}{Stocks} & \multicolumn{1}{l|}{Energy} \\ \hline
\multirow{8}{*}{\begin{tabular}[c]{@{}c@{}}Discriminative\\ Score\\ \\ (Lower the Better)\end{tabular}} & \multicolumn{1}{l|}{DiffTime (Ours)}  & .013 $\pm$ .006                       & .097 $\pm$ .016               & .445 $\pm$ .004 \\

 & \multicolumn{1}{l|}{COP (Ours)}    & .020 $\pm$ .001 & \textbf{.050 $\pm$ .017} & \textbf{.101 $\pm$  .019} \\
 & \multicolumn{1}{l|}{GT-GAN}   & .012 $\pm$ .014 & .077 $\pm$ .031      & .221 $\pm$  .068 \\
 & \multicolumn{1}{l|}{TimeGAN}   & \textbf{.011 $\pm$ .008} & .102 $\pm$ .021          & .236 $\pm$ .012  \\
 & \multicolumn{1}{l|}{RCGAN}     & .022 $\pm$ .008  & .196  $\pm$ .027 & .336 $\pm$  .017        \\
 & \multicolumn{1}{l|}{C-RNN-GAN} & .229 $\pm$ .040 & .399  $\pm$  .028 & .449 $\pm$  .001        \\
 & \multicolumn{1}{l|}{T-Forcing} & .495 $\pm$ .001 & .226  $\pm$  .035 & .483 $\pm$  .004        \\
 & \multicolumn{1}{l|}{P-Forcing} & .430 $\pm$ .227 & .257  $\pm$  .026 & .412 $\pm$  .006        \\
 & \multicolumn{1}{l|}{WaveNet}   & .158 $\pm$ .011 & .232  $\pm$  .028  & .397 $\pm$ .010         \\
 & \multicolumn{1}{l|}{WaveGAN}   & .277 $\pm$ .013 & .217  $\pm$  .022 & .363  $\pm$ .012         \\ \hline
\multirow{9}{*}{\begin{tabular}[c]{@{}c@{}}Predictive\\ Score\\ \\ (Lower the Better)\end{tabular}}     & \multicolumn{1}{l|}{DiffTime (Ours)}  & \textbf{.093 $\pm$ .000}      & \textbf{.038 $\pm$ .001}  & .252 $\pm$ .000  \\
 & \multicolumn{1}{l|}{COP (Ours)}    & .095 $\pm$ .002 & .041 $\pm$ .001 & \textbf{.250 $\pm$  .003} \\

 & \multicolumn{1}{l|}{GT-GAN}   & .097 $\pm$ .000 & .040 $\pm$ .000           & .312 $\pm$ .002           \\
 & \multicolumn{1}{l|}{TimeGAN}   & .093 $\pm$ .019 & \textbf{.038   $\pm$ .001 }      & .273 $\pm$ .004          \\
 & \multicolumn{1}{l|}{RCGAN}     & .097  $\pm$  .001  & .040  $\pm$ .001 & .292  $\pm$ .004 \\
 & \multicolumn{1}{l|}{C-RNN-GAN} & .127 $\pm$  .004  & \textbf{.038  $\pm$ .000} & .483  $\pm$ .005   \\
 & \multicolumn{1}{l|}{T-Forcing} & .150 $\pm$  .022 & \textbf{.038  $\pm$ .001} & .315   $\pm$ .005  \\
 & \multicolumn{1}{l|}{P-Forcing} & .116 $\pm$ .004   & .043  $\pm$ .001 & .303  $\pm$ .005   \\
 & \multicolumn{1}{l|}{WaveNet}   & .117 $\pm$  .008  & .042  $\pm$ .001  & .311  $\pm$ .006   \\
 & \multicolumn{1}{l|}{WaveGAN}   & .134 $\pm$  .013  & .041  $\pm$ .001 & .307    $\pm$ .007              \\ \cline{2-5} 
 & Original                       & .094 $\pm$ .001              & .036 $\pm$ .001                       & .250 $\pm$ .003              \\  \hline \cline{1-1}
\end{tabular}
}
\end{table}

\subsection{Constrained Generation}
We now evaluate the performance of our approaches against the constraints shown in Figure~\ref{fig:example_constraints} using daily stock data. For univariate constraints (i.e., trend, fixed values, and global minimum) we consider only the \textit{Open} value from the daily stock dataset. We consider as benchmarks the best three SoA approaches from Table~\ref{tab:unconditional} (i.e., GT-GAN, TimeGAN, and RCGAN). For trend- and fixed-values constraints we condition their generative process so that different trends and values can be used at inference time. For the other constraints, we add a penalty term in the training loss~\cite{di2020efficient,xu2018semantic}.

\paragraph{Soft Constraints} In table~\ref{table:soft_constr} we constrain the synthetic TS to follow a given trend, computed as a 3-degree polynomial approximation from the original samples. Our approaches generate synthetic data that are closer to the input trend, with the smallest relative distance (i.e., \textit{perc. error distance}). Moreover, our approaches are among the best in terms of realism and usefulness. In Appendix~\ref{app:trend_sine} we investigate the use of sinusoidal trends, including additional evaluation metrics.

\begin{table}[hbt]
\centering
\caption{Soft Constraints (Trend) Time-Series Generation (Bold indicates best performance).}\label{table:soft_constr}
\resizebox{0.6\linewidth}{!}{
\begin{tabular}{|l|c|c|c|c|}
\hline
Algo & Discr-Score & Pred-Score & Inference-Time & Perc. Error Distance \\ \hline
COP (Ours) & \textbf{0.01±0.01} & \textbf{0.20±0.00} & 0.73±0.05 & \textbf{0.015±0} \\
DiffTime (Ours) & \textbf{0.01±0.01} & \textbf{0.20±0.00} & 0.02±0.00 & 0.018±0 \\
GT-GAN & 0.04±0.03 & 0.22±0.00 & \textbf{0.00±0.00} & 1.378±2 \\
TimeGAN & 0.02±0.02 & \textbf{0.20±0.00} & \textbf{0.00±0.00} & 0.073±0 \\ 
RCGAN & 0.02±0.01 &\textbf{0.20±0.00}& \textbf{0.00±0.00} & 0.071±0 \\ \hline
\end{tabular}}
\end{table}

\begin{figure}[hbt]
\subcaptionbox{\scriptsize \textbf{COP}}{\includegraphics[width=0.16\textwidth]{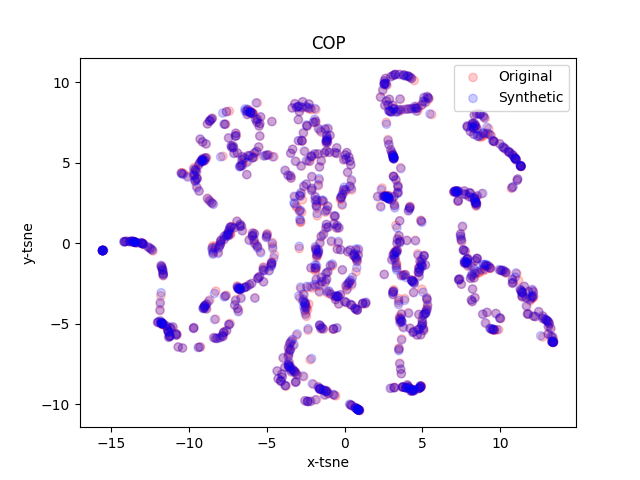}}%
\hfill
\subcaptionbox{\scriptsize \textbf{Guided DiffTime}}{\includegraphics[width=0.16\textwidth]{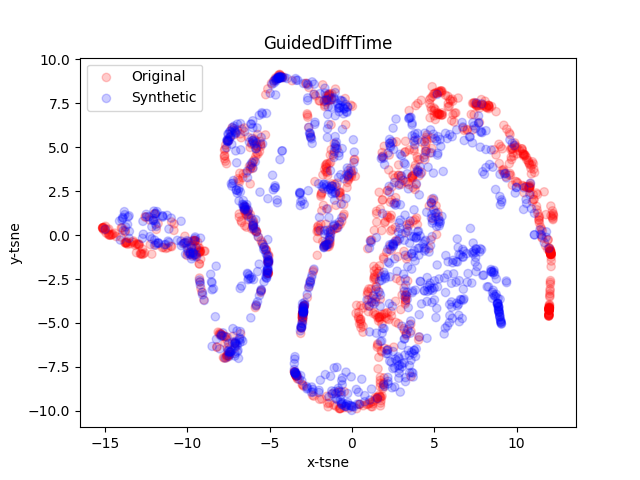}}%
\hfill
\subcaptionbox{\scriptsize \textbf{LossDiffTime}}{\includegraphics[width=0.16\textwidth]{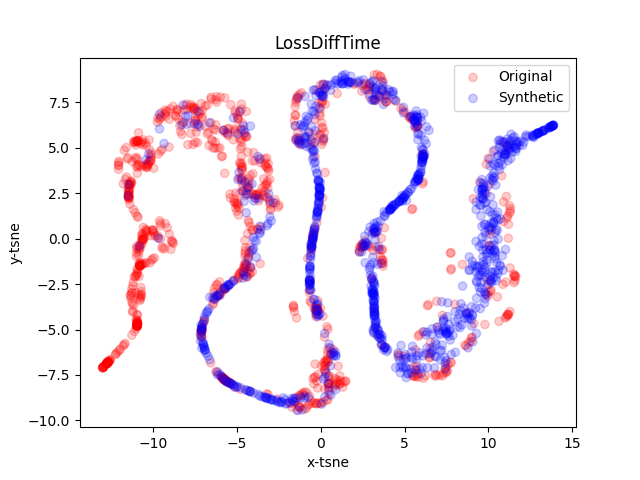}}%
\hfill
\subcaptionbox{\scriptsize GT-GAN}{\includegraphics[width=0.16\textwidth]{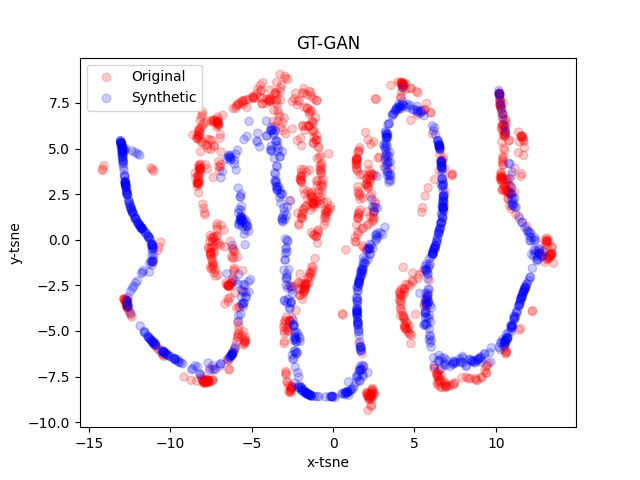}}%
\hfill
\subcaptionbox{\scriptsize TimeGAN}{\includegraphics[width=0.16\textwidth]{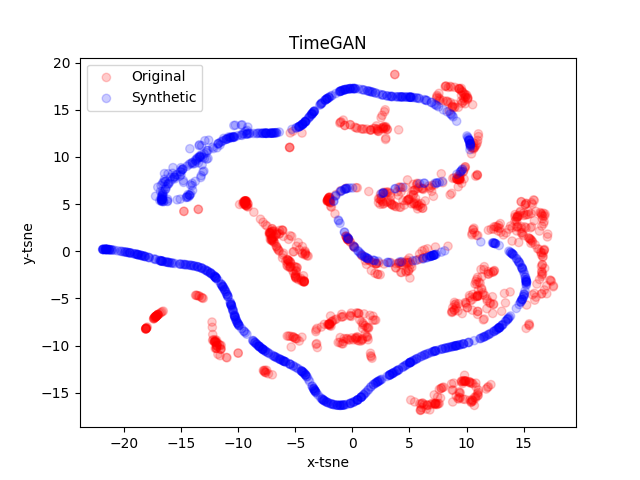}}%
\hfill
\subcaptionbox{\scriptsize RCGAN}{\includegraphics[width=0.16\textwidth]{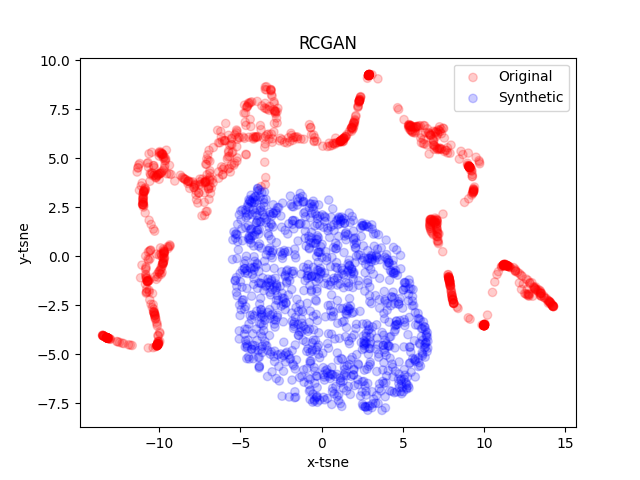}}%
\caption{t-SNE visualizations on \textit{Global-Min} constrained data, where a greater overlap of blue and red dots implies a better distributional-similarity between the generated data and original data. Our approaches show the best performance.}\label{fig:tsnemin}
\end{figure}

\paragraph{Hard Constraints} In Table~\ref{table:hard_constr} we evaluate all the approaches against hard constraints (see Fixed Points, Global Min, and Multivariate in Figure~\ref{fig:example_constraints}). For \textit{Global Min} almost all approaches have a great \textit{satisfaction rate}. However, our approaches are above $0.90$ \textit{and} have the best discriminative and predictive score. Additionally, while $100\%$ of the synthetic time-series generated by TimeGAN and RCGAN guarantee this type of constraint, they do not approximate the input distribution well(see Figure~\ref{fig:tsnemin}). For most complex constraints like the multivariate one, the satisfaction rate drops for most of the benchmarks while for our \textit{GuidedDiffTime} and \textit{COP} the satisfaction rate is still very high, with great realism and usefulness. Finally, when we employ the fixed point constraints, we fix the values of the points at index $6$ and $18$. All the benchmarks fail to satisfy these constraints, while we show instead that \textit{DiffTime} is able to always guarantee this constraint, by enforcing it during the diffusion steps. Most importantly, it achieves very good discriminative and predictive scores with low inference time. To summarize our results: COP achieves almost always the best realism and usefulness scores, but with higher inference time and using original input TS as seed (which makes the generated TS very similar to the input data); diffusion models are also very powerful with lower inference time and use random noise as input seed as opposed to a real TS (this gives us better variety in TS compared to COP); and \textit{GuidedDiffTime} is able to enforce new constraints without any re-training yet achieving excellent performance. 

\renewrobustcmd{\bfseries}{\fontseries{b}\selectfont}
\renewrobustcmd{\boldmath}{}
\newrobustcmd{\B}{\bfseries}

\begin{table}[h]
\caption{Hard Constraints Time-Series Generation (Bold indicates best performance).}\label{table:hard_constr}
\centering
\resizebox{\linewidth}{!}{
\begin{tabular}{|l|l|c|c|c|c|c|}
\hline
\textbf{Constraint} & \textbf{Algo} & Discr-Score & Pred-Score & Inference-Time & Satisfaction Rate  & Fine-Tuning Time  \\ \hline
\multirow[c]{7}{*}{Global Min} 
 & COP (Ours) & \B 0.02±0.01 & \B 0.20±0.00 & 19.1±1.01 & \B 1.00±0.00 & \B \B 0.00±0.00 \\
  & GuidedDiffTime (Ours) & 0.03±0.02 & 0.21±0.00 & 0.03±0.00 & 0.90±0.01 &  3.01±0.10 \\
 & LossDiffTime (Ours) & 0.22±0.03 & 0.38±0.00 & 0.02±0.00 & 0.99±0.00 &  6.00±0.60 \\
 & GT-GAN & 0.04±0.02 & 0.22±0.00 & \B 0.00±0.00 & 0.87±0.02 &  
 9.30±1.30 \\
 & TimeGAN & 0.03±0.02 & 0.21±0.00 & \B 0.00±0.00 & \B 1.00±0.00 &  \B 0.00±0.00 \\
 & RCGAN & 0.23±0.03 & \B 0.20±0.00 & \B 0.00±0.00 & \B 1.00±0.00 &  \B 0.00±0.00 \\ \hline
\multirow[c]{7}{*}{Multivariate (OHLC)} 
 & COP (Ours) & \B 0.04±0.02 & \B 0.04±0.00 & 2.17±0.10 & \B 1.00±0.00 & \B 0.00±0.00 \\
  & GuidedDiffTime (Ours) & 0.08±0.00 & \textbf{0.04±0.10} & 0.15±0.00 &  0.72±0.02 &  31.0±1.50 \\
 & LossDiffTime (Ours) & 0.35±0.04 & \B 0.04±0.01 & 0.14±0.00 &  0.69±0.01  & 57.5±5.01 \\
 & GT-GAN & 0.22±0.07 & 0.05±0.00 & \B 0.00±0.00 & 0.05±0.01 &  44.5±3.01 \\
 & TimeGAN & 0.24±0.03 & 0.05±0.00 & \B 0.00±0.00 & 0.51±0.02 &  16.1±1.30 \\ 
 & RCGAN & 0.35±0.04 & \B 0.04±0.00 & \B 0.00±0.00 &  0.00±0.00 &  95.1±4.03 \\ \hline
\multirow[c]{6}{*}{Two Fixed Points}
 & COP (Ours) & \B 0.02±0.02 & \B 0.20±0.00 & 0.56±0.11 & \B 1.00±0.00 & \B 0.00±0.00 \\
 & DiffTime (Ours) & 0.04±0.03 & 0.21±0.00 & 0.01±0.00 & \B 1.00±0.00 & \B 0.00±0.00 \\
 & GT-GAN & 0.04±0.03 & 0.21±0.00 & \B 0.00±0.00 &  0.00±0.00 & 0.99±0.10 \\
 & TimeGAN & 0.03±0.01 & \B 0.20±0.00 & \B 0.00±0.00 &  0.00±0.00 &  0.84±0.00 \\ 
 & RCGAN &  \B 0.02±0.02 &  \B 0.20±0.00 & \B 0.00±0.00 & 0.00±0.00 &  0.87±0.20 \\ \hline
\end{tabular}
}
\end{table}

\section{Conclusions}
In summary, we defined the problem of generating synthetic TS data with both soft and hard constraints, and we presented a set of novel methods. We evaluated our approaches on different datasets and we compared their performance against existing state-of-art methods. We showed that our approaches outperform existing work both qualitatively and quantitatively. Most importantly, we introduced \textit{GuidedDiffTime} to handle new constraints without re-training, and we showed that the \textit{COP-method} can be used to fine-tune candidate solutions. Please refer to the Appendix for more details on experiments comparing the methods presented herein.

\section*{Disclaimer}
This paper was prepared for informational purposes by
the Artificial Intelligence Research group of JPMorgan Chase \& Co. and its affiliates (``JP Morgan''),
and is not a product of the Research Department of JP Morgan.
JP Morgan makes no representation and warranty whatsoever and disclaims all liability,
for the completeness, accuracy or reliability of the information contained herein.
This document is not intended as investment research or investment advice, or a recommendation,
offer or solicitation for the purchase or sale of any security, financial instrument, financial product or service,
or to be used in any way for evaluating the merits of participating in any transaction,
and shall not constitute a solicitation under any jurisdiction or to any person,
if such solicitation under such jurisdiction or to such person would be unlawful.

\bibliography{main}
\bibliographystyle{unsrt}

\newpage

\appendix

\section{Datasets}
In our experiments we consider two publicly available datasets and one synthetic dataset. All the datasets have different characteristics such as periodicity, noise, correlation, and number of features. 
In particular, the \textbf{daily stock} dataset uses daily historical Google stock data from 2004 to 2019 with 6 features, namely \textit{open, high, low, close, adjusted close, and volume}.
When running univariate experiments, we only used the \textbf{open} feature from the daily-stock dataset. 
The \textbf{energy data} from the UCI Appliances energy prediction dataset~\cite{candanedo2017data} contains 28 features, 
at 10-minute resolution, with noisy periodicity and correlation. 
Finally, the synthetic \textbf{sine} dataset contains multivariate sinusoidal time-series, 
where each dimension is created independently, sampling the frequencies and phases according the following equation:
\begin{equation}\label{eq:sine}
    x_i(t)=sin(2\pi \eta_it+\theta_i), \ s.t. \ \eta_i\sim \mathcal{U}[0,1] \land \theta_i\sim\mathcal{U}[-\pi,\pi], \ \forall i \in \{1,...,5\}
\end{equation}
This synthetic dataset comes from prior work ~\cite{timeGAN}.
Where not-otherwise stated, we consider time-series of length 24. 
In table \ref{tab:dataset_summary} we summarize the dataset properties, 
while in Table~\ref{tab:goog}, Table~\ref{tab:energy} and Table~\ref{tab:sine_data} we report all the detailed statistics. 

\begin{table}[ht]
\centering
\caption{\label{tab:dataset_summary}Dataset Description}
\begin{tabular}{l|ccccc}
\hline
Dataset Name & Data type & Samples & $\dim(\mathbf{x})$ & Data Resolution & Link  \\ \hline %
Stocks  & Real & 3,773     & 6    & 1-day    &     \href{http://finance.yahoo.com/quote/GOOG/history?p=GOOG}{Link}     \\ 
Energy  & Real & 19,711    & 28   & 10-minutes  & \href{http://archive.ics.uci.edu/ml/datasets/Appliances+energy+prediction }{Link}  \\ 
Sines   & Synthetic & 10,000     & 5    & data-point &    -                  \\ \hline 
\end{tabular}
\end{table}

All the datasets are normalized between [-1,1] for the diffusion models.

\section{Benchmarks}
We compare our approaches against existing time-series generative models, i.e., 
GT-GAN~\cite{jeon2022gt}, TimeGAN~\cite{timeGAN}, RCGAN~\cite{esteban2017real}, 
C-RNN-GAN~\cite{mogren2016c},a Recurrent Neural Networks (RNN) trained with T-Forcing 
and P-Forcing~\cite{lamb2016professor,graves2013generating}, WaveNET~\cite{oord2016wavenet}, 
and WaveGAN~\cite{donahue2018adversarial}.
We use and modify the publicly available source code for each of the methods:
\begin{itemize}
    \item GT-GAN~\cite{jeon2022gt} :  \href{https://openreview.net/attachment?id=ez6VHWvuXEx&name=supplementary_material}{openreview}
    \item TimeGAN~\cite{timeGAN} : \href{https://github.com/jsyoon0823/TimeGAN}{github}
    \item RCGAN~\cite{esteban2017real} : \href{https://github.com/ratschlab/RGAN}{github}
    \item  C-RNN-GAN~\cite{mogren2016c} : \href{https://github.com/olofmogren/c-rnn-gan}{github}
    \item T-Forcing~\cite{graves2013generating} : \href{https://github.com/snowkylin/rnn-handwriting-generation}{github}
    \item P-Forcing~\cite{lamb2016professor} : \href{https://github.com/anirudh9119/LM_GANS}{github}
    \item WaveNET~\cite{oord2016wavenet} : \href{https://github.com/ibab/tensorflow-wavenet}{github}
    \item WaveGAN~\cite{donahue2018adversarial} :\href{https://github.com/chrisdonahue/wavegan}{github}
\end{itemize}

In particular for T-forcing and P-forcing we use a 3-layer GRUs with hidden dimensions four times the size of input features, as suggested in~\cite{timeGAN}. 

For constrained time-series generation scenarios, we restrict our analysis to the top three performing benchmarks, and we adapt their architectures as follows:
\begin{itemize}
    \item \textbf{Trend} and \textbf{Fixed-Values} constraints. We condition the generators by introducing a new dimension that contains the \textbf{trend} or the \textbf{fixed-values}. During the training the trend and the fixed-values are extracted directly from the input time-series to let the models rely on this additional information.  
    \item \textbf{Hard-Constraints}. We follow the recent work in~\cite{di2020efficient,xu2018semantic,bengio2017deep}, and we extend the benchmark architectures by introducing a penalty loss. This loss penalizes the generative models proportional to how much the generated time-series violate the input constraint, and it is added to the original model loss by a scale factor $\lambda_{c}$, which we consider as a hyper-parameter tuned in the experiments. Most importantly, for GT-GAN and TimeGAN we add an optimization step, which lasts $1/4$ of the total epochs, and it optimizes the generator alone w.r.t. to the constraint loss. We found that these models benefit from this additional optimization step.
\end{itemize}

We also employ \textit{rejection-sampling} and \textit{fine-tuning} using COP-method on all their generated synthetic TS. 

\section{Implementation details}
We implement our work in Python. Specifically, we use PyTorch~\cite{paszke2019pytorch} to implement diffusion models, and we use the Sequential Least Squares Programming (SLSQP) solver~\cite{jorge2006numerical} in Scipy's optimization module~\cite{2020SciPy-NMeth} for COP-method. 

All the deep generative models are trained on an NVIDIA T4 GPU, with 4 CPU and 16gb or RAM. To compare the computational times, the inference is done on a 4 CPU 3rd generation AMD EPYC processors for all the models including the COP-method.
The default hyper-parameters for the diffusion model are reported in Table~\ref{tab:parameters}; we specify in each section when different hyper-parameters are used to compute the results.

\begin{table}[hbt]
\centering
\caption{Diffusion Model default hyper-Parameters}\label{tab:parameters}
\vspace{0.2in}
\begin{subtable}{0.48\textwidth}
\centering
\begin{tabular}{|l|c|}
\toprule
\textbf{Hyper-parameter} & \textbf{Value} \\
\midrule
batch-size & 16 \\
$\beta_1$ & 1.0e-06 \\
$\beta_T$ & 0.5 \\
channels & 64 \\
diffusion-embedding-dim & 128 \\
epochs & 10000 \\
kernel-size & 2 \\
\bottomrule
\end{tabular}
\end{subtable}
\hfill
\begin{subtable}{0.48\textwidth}
\centering
\begin{tabular}{|l|c|}
\toprule
\textbf{Hyper-parameter} & \textbf{Value} \\
\midrule
layers & 4 \\
learning-rate & 0.0001 \\
n-heads & 8 \\
noise-steps $T$ & 50 \\
noise schedule & quadratic \\
weight-decay & 1.0e-06 \\
constraint & None \\
\bottomrule
\end{tabular}
\end{subtable}
\end{table}

\section{Constrained Optimization Method (COP)}

\subsection{Algorithm and Details}

\begin{figure}[ht]
    \centering
    \includegraphics[trim={0.3in 0.3in 0.3in 0.3in}, width=1\textwidth]{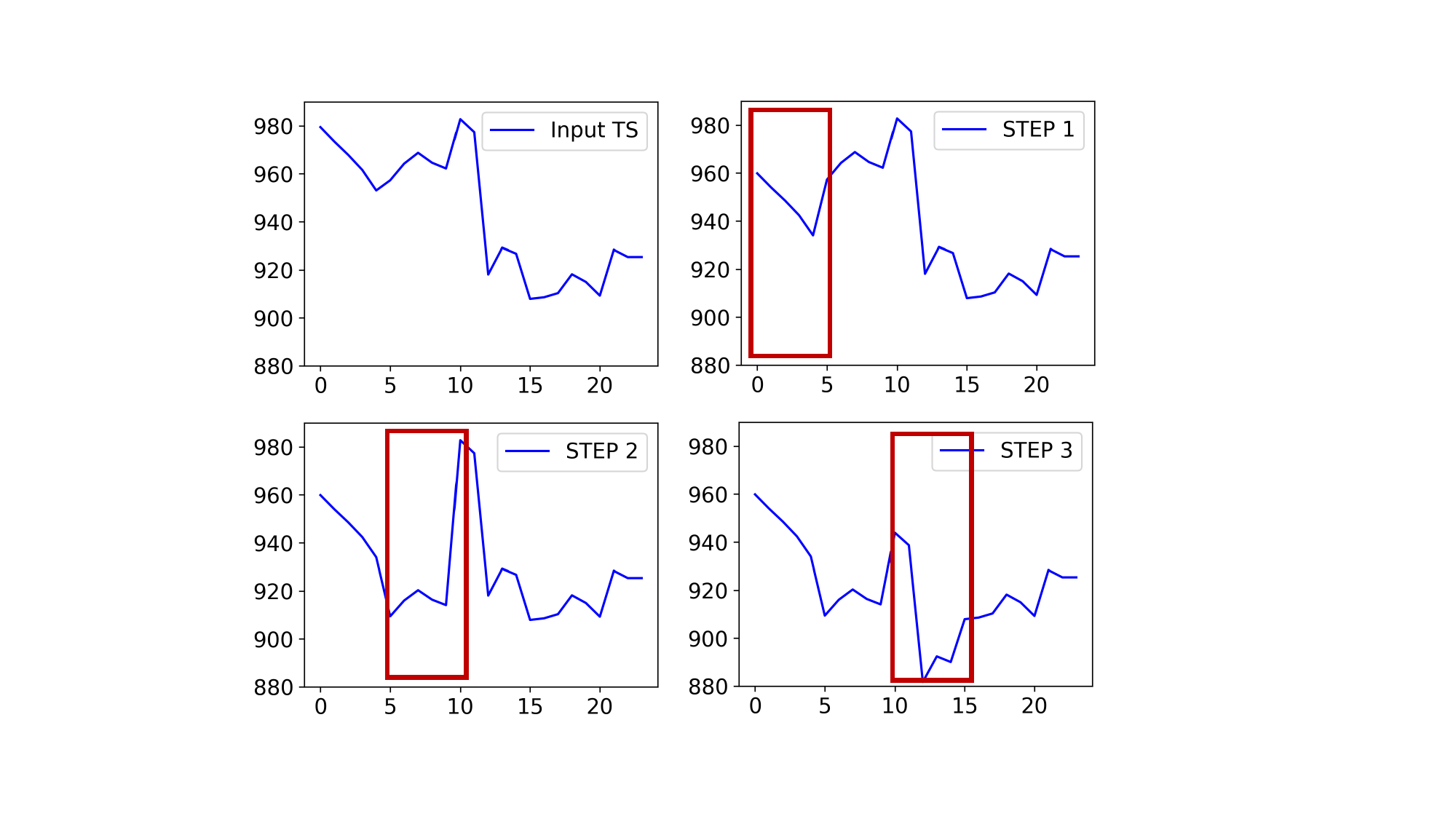}
    \vspace{-0.1in}
    \caption{An example of a TS being altered (fine-tuning or generation) by the \textit{COP-method}; changes are made within a window that convolves over the TS. Top-left is the original TS and the window shifts to the right (follow images clockwise) and changes the TS from the starting data in order to respect constraints.}
    \label{fig:COP_sliding_window}
\end{figure}

In Algorithm \ref{alg:COP_repeated_optimization_search}, we present the procedure of the \textit{COP-method} to generate or fine-tune TS such that they conform to constraints. An illustration of the method is shown in Figure~\ref{fig:COP_sliding_window}. 
{\color{black} We recall that COP frames the task of generating a TS sample as optimizing the value of a set of ordered points that make up the TS sample, such that the sample satisfies domain properties (like auto-correlation). In particular, it starts from an initial sample TS (taken from the dataset, or randomly generated) and optimizes its values according to its objective (e.g., maximize the distance between the generated and initial sample), while respecting some constraints (which could be statistical properties or additional structural constraints). With this problem formulation, we can use existing COP solvers (e.g., SLSQP) to get new samples by solving those non-linear constraints and objectives. Thus the COP solver using the specified constraints and objective function becomes the generative process. }

\begin{algorithm}[hbt]    
    \caption{\textit{Constrained Optimization Search for TS Generation/Fine-Tuning}}
    \label{alg:COP_repeated_optimization_search}
    \begin{algorithmic}[1]
        \STATE \textbf{Input:} Seed TS $\bx_0$, Constraints $C$ , Objective Function $f$, budgets for constraints $b$, window size $\theta_w$, window overlap ratio $\theta_v$, number of retries $\eta_r$, number of iterations per sample $\eta_i$
        \STATE \textbf{Output:} $\hat{\bx}$
        
        \STATE len\_TS $\gets$ get\_length($\bx_0$)
        \STATE bounds $\gets$ (0.98*min($\bx_0$),1.02*max($\bx_0$))
        
        \FOR{$r \gets 1$ to $\eta_r$ }
            \STATE $\hat{\bx}_{seed}$ $\gets$ $\bx_0$ \\
            \COMMENT { the seed here is a dataset sample, but it can be initialized in different ways; for example, it can also be sampled using brownian motion and rescaled using a sampled real TS' mean and variance}          
            \STATE $\hat{\bx}$ $\gets$ copy($\hat{\bx}_{seed}$)
            \STATE $W_{pos}$ $\gets$ get\_all\_window\_positions(len\_TS,$\theta_w$, $\theta_v$) \\
            \COMMENT{ Window positions with overlap would be (for example) [0, 4], [2, 6], ...}
            \STATE $b_r$ $\gets$ $b \times 2^r$ \\
            \STATE $C_b$ $\gets$ update\_constraints\_with\_budget(C,$b_r$) \\
            
            \FOR{$i \gets 1$ to $\eta_i$}
                \STATE $v_f^*$ $\gets \infty$ 
                \STATE $w_{pos}^*$ $\gets \emptyset $ 
                \STATE $\hat{\bx}_{i,best}$ $\gets$ copy($\hat{\bx}$)
                \FOR{($\phi_{start}$,$\phi_{end}$) $\in$ $W_{pos}$}
                    \STATE (status, $\hat{v}_f$, $\hat{\bx}_{i,\phi_{start},\phi_{end}})$ $\gets$ SLSQP\_optimize(f, $C_b$, $\hat{\bx_i}$[$\phi_{start}$:$\phi_{end}$], bounds) \\
                    \COMMENT{solver returns the updated points. Then we insert them into the candidate solution to get the updated solution}
                    \STATE $\hat{\bx}_{i,temp}$ $\gets$ $\hat{\bx_i}$[0:$\phi_{start}$] | $\hat{\bx}_{i,\phi_{start},\phi_{end}}$ | $\hat{\bx_i}$[$\phi_{end}$:]
                    \IF{ (status = success) \& ($\hat{v}_f < v_f^*$)}
                        \STATE $v_f^*$ $\gets$ $\hat{v}_f$
                        \STATE $\hat{\bx}_{i,best}$ $\gets$ $\hat{\bx}_{i,temp}$
                        \STATE $w_{pos}^*$ $\gets$ ($\phi_{start}$,$\phi_{end}$)
                    \ENDIF
                \ENDFOR
                
                \IF{$v_f^*$ $\neq \infty$}
                    \STATE $W_{pos} \gets W_{pos}\backslash w_{pos}^*$ 
                    \STATE $\hat{\bx}$ $\gets$ $\hat{\bx}_{i,best}$
                \ENDIF
            \ENDFOR
            
            \IF{$\hat{\bx}$ $\neq$ $\hat{\bx}_{seed}$}
                \RETURN $\hat{\bx}$
            \ENDIF
        \ENDFOR
        
        \RETURN $\emptyset$
    \end{algorithmic}
\end{algorithm}

For Algorithm \ref{alg:COP_repeated_optimization_search}, the parameters we used in our experiments are as follows: $b = 0.1$, $\theta_w = 3$, $\theta_r = 0.5$, $\eta_r = 10$, $\eta_i = 2$. If the task is to match a trend, then we set $\theta_w = L$ to be the length of the TS. For very long time series, using a $\theta_w < L$ can help make it easier for the COP solver by breaking the problem into chunks. However, for some constraints, we may have to solve for all points (i.e. $\theta_w = L$) at once. 
The initial seed TS ($\bx_0$) is a sample from the dataset (Line 6 of Algorithm \ref{alg:COP_repeated_optimization_search}), but we also do experiments with different seed TS, and present the results in Section~\ref{sec:cop_initial_seed}.
With respect to the constraints that we input into the COP-method, they come from the hard constraints in $C$. The objective used in COP-method for generating TS is to maximize L2-norm of the difference with the seed TS. If a soft-constraint in $C$ is a trend to be followed, then the objective is updated to encourage the COP solver to minimize the L2-norm of the generated TS with the trend; the objective becomes a weighted combination as follows: $f(\hat{\bx},\bx,\mathbf{s}) = (1-\omega)*\left\| \mathbf{x} - \hat{\bx} \right\|_{2}^{2} - \omega*\left\| \mathbf{s} - \hat{\bx} \right\|_{2}^{2}$ where $\omega \in [0,1]$. By using different $\omega$ values one can trade-off between matching the trend and pushing the generated TS to be different from the seed TS. For our experiments that have a trend in constraints $C$, we set $\omega = 1.0$.
\\
Finally, if COP-method is used to fine-tune a TS to fit constraints, then the objective is changed to minimize L2-norm with the input TS. This is to help prevent changing the TS significantly when fine-tuning. We found L2-norm to work well for our experiments, but other distance functions, like L1 norm or percentage difference, can be used as well.

The constraints used in the COP-method are the same as those input to the other methods, with one exception. We need to additionally constrain the solver to match the statistical properties of real data. 

\subsection{Realism constraints for Time Series}\label{sup:Realism_constraints}
To give an example of constraints one might like to impose for the realism of a generated dataset, let us consider stock prices TS in \textbf{daily stocks} dataset. It is typically required for synthetic stock price TS to preserve stylized facts (a term in economics ~\cite{sewell2011characterization}) about the financial markets. These include the distributions of returns, and 
return auto-correlations~\cite{bouchaud2018trades, vyetrenko2019real}.  
The return at each point in time is defined as the percentage change in the value. For auto-correlation, we use the discrete auto-correlation function for real data. Equations for returns, and autocorrelation (for 1 dimensional TS) are shown in Equation~\ref{eq:autocorr_and_return} on the left and right side respectively.

\begin{equation}
    r_{\bx}(t) = \frac{\bx_{t} - \bx_{t-1}}{\bx_{t-1}}, \quad \rho_{\bx\bx}(\tau) = \frac{E[(\bx_{t+\tau}-\mu)(\bx_{t}-\mu)]}{\sigma_{\bx}^2}
    \label{eq:autocorr_and_return}
\end{equation}

where $\tau$ is the auto-correlation lag parameter and $\mu$ is the mean value of the TS. 

In the COP method we use auto-correlation of the returns as a constraint -- we task the solver to match this property-- for the \textit{daily stocks} dataset. For the \textit{Energy} dataset and \textit{Sines} data, we used the autocorrelation of the TS (not the autocorrelation of the returns) as the constraint to improve the realism of the data generated. 
In our experiments, the auto-correlation lag parameter $\tau$ is set to 5. We constrain the solver by taking the L2-norm of the difference in the autocorrelation vectors and limiting that error. The initial budget for this constraint is set to 0.1, which can increase if the solver cannot find a solution (see line 9 in Algorithm \ref{alg:COP_repeated_optimization_search}), and then the solver will iteratively try again.

\subsection{WGAN-based Constrained Optimization (WGAN-COP)}\label{app:wgan_cop}

One of the main limitations of the presented COP-method is the need to explicitly define all data properties that generated synthetic data must have. In the \textit{COP} method we had to add constraints to match the autocorrelation of the TS signal as a way of capturing desired TS properties. We would ideally like a function $f^* : \mathcal{X} \rightarrow \mathbb{R}$ able to evaluate how much the synthetic data resembles the real one. However, a single domain-agnostic metric to evaluate synthetic data does not exist yet~\cite{alaa2022faithful}.

Following the recent advances in generative adversarial networks however, we notice that the \textit{critic} $f_w$ of a WGAN~\cite{arjovsky2017wasserstein,gulrajani2017improved} matches the description of our function $f^*$, in the sense that it will return a higher value if an input sample resembles real data, i.e. it looks like it was sampled from the real distribution of data. The critic also has additional interesting properties, as it is trained using the Wasserstein, or Earth-Mover (EM), distance $W(q, p)$ between the real distribution $q(\bx)$ and the synthetic one $p(\bx)$.  This distance is continuous everywhere and differentiable almost everywhere under mild assumptions~\cite{arjovsky2017wasserstein}, and the critic $f_w$ must lie within the space of 1-Lipschitz function $||f_w||_L \leq 1$. This means that: 1) we can train the critic till convergence to get a reliable approximation of the Wasserstein distance~\cite{arjovsky2017wasserstein}; 2) the critic value correlates with sample realism or quality~\cite{gulrajani2017improved}.  

Therefore we can first train the WGAN architecture, and then we can replace the constraints used to represent desired TS data properties (e.g. auto-correlation) with the critic function $f_w$ which we put into the objective function. We call this adaptation of COP-method with WGAN as \textit{WGAN-COP}. WGAN-COP tries to maximize $f_w$ while guaranteeing any additional explicit constraints. A COP-solver can get gradients w.r.t. sample $\mathbf{x}$ via back-propagation over the critic's neural network $\nabla_\theta f_w(\bx)$. In general, this approach does not hold for all the GAN architectures (e.g., those that minimize KL divergence), as the gradients can saturate with no guide for the COP. 

Initial experiments show that \textit{WGAN-COP} does not need to explicitly define realism constraints with comparable performance, however the training of the WGAN is in general expensive and unstable, especially with high-dimensional data. 

\section{Diffusion Models}
We now introduce the pseudo-code algorithms for the proposed diffusion-based approaches, 
namely \textit{DiffTime}, \textit{Loss-DiffTime} and \textit{Guided-DiffTime}. 
For all the models we keep the same choice of diffusion steps $T=50$, and 
we compute the noise variance $\boldsymbol{\beta}$ using a quadratic scheduler with a start value of 
$\beta = $1.0e-06 and end value of $\beta_T = 0.5$. We evaluate the impact of different choices of $T$ and $\boldsymbol{\beta}$ in Section~\ref{sec:diff_step}; and different noise scheduler in Section~\ref{sec:noise_var}. We recall also that $\alpha_t := 1 - \beta_t$ and $\hat{\alpha}_t := \prod_{i=1}^{t} \alpha_i$.

\subsection{DiffTime}
\textit{DiffTime} is our base diffusion model approach, which can be trained to incorporate both \textit{trend} and \textit{fixed-values} constraints. 
The basic model (i.e., without any constraint) follows the standard diffusion model procedures. 
Algorithm~\ref{alg:diff_time_training} shows \textit{DiffTime} training process, while Algorithm~\ref{alg:diff_time_sampling} shows the inference process to generate new synthetic time-series.

\begin{figure}[htbp]
\begin{minipage}[t]{0.49\textwidth}
\begin{algorithm}[H]
  \caption{Unconstrained \textit{DiffTime} Training} \label{alg:diff_time_training}
\begin{algorithmic}[1]
   \STATE \textbf{Input:} input TS distribution $q(\bx_0)$, number of epochs $\mathbf{E}$
   \STATE \textbf{Output:} trained diffusion function $\rvepsilon_\theta$ 
   \FOR{$i=1$ \textbf{to} $\mathbf{E}$}
   \STATE $t \sim \textrm{Uniform}(\{1,\ldots,T\})$
   \STATE $\bx_0 \sim q(\bx_0) \ $; $\ \rvepsilon \sim \gN(\bmzero,\bmI)$ 
    \STATE Take gradient step on \\  $\grad_\theta \left\| \rvepsilon - \rvepsilon_\theta(\sqrt{\hat{\alpha}_t} \bx_0 + \sqrt{1-\hat{\alpha}_t}\rvepsilon, t) \right\|^2$
\ENDFOR
\end{algorithmic}
\end{algorithm}
\end{minipage}
\hfill
\begin{minipage}[t]{0.51\textwidth}
\begin{algorithm}[H]
  \caption{Unconstrained \textit{DiffTime} Sampling} \label{alg:diff_time_sampling}
\begin{algorithmic}[1]
   \STATE \textbf{Input:} trained diffusion function $\rvepsilon_\theta$ 
    \STATE \textbf{Output:} synthetic time-series $\hat{\bx}$    
    \STATE $\hat{\bx}_{T} \sim \mathcal{N}(\bmzero, \bmI)$
   \FOR{$t=\mathbf{T}$ \textbf{to} $1$}
    \STATE  $\boldsymbol{z} \sim \mathcal{N}(\bmzero, \bmI)$ if $t > 1$, else $\boldsymbol{z} = \bmzero$
      \STATE \mbox{$\hat{\bx}_{t-1} = \frac{1}{\sqrt{\alpha_t}}\left( \hat{\bx}_t - \frac{1-\alpha_t}{\sqrt{1-\hat{\alpha}_t}} \rvepsilon_\theta(\hat{\bx}_t, t) \right) + \sigma_t \boldsymbol{z}$}
\ENDFOR
\STATE \textbf{Return} $\hat{\bx}_0$ 
\vspace{0.55em}
\end{algorithmic}
\end{algorithm}
\end{minipage}
\end{figure}

\begin{figure}[H]
\centering
\includegraphics[width=0.8\textwidth]{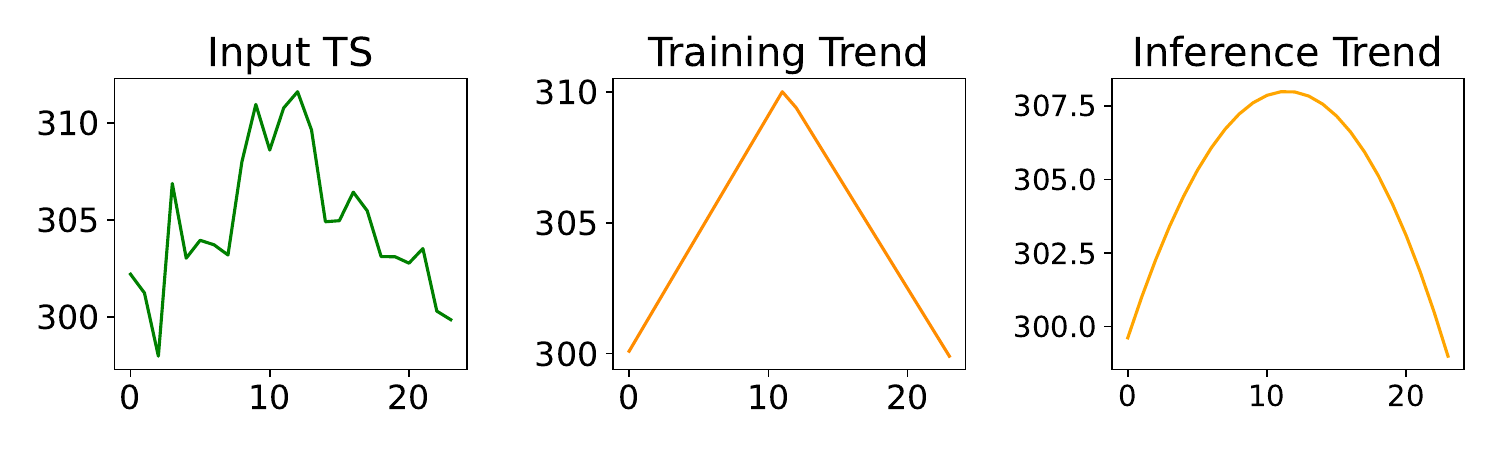}
\caption{An example of trend-constraints.}\label{fig:trend_example}
\end{figure}

\paragraph{Trend Constraint.} To constrain a particular trend, we condition the diffusion process using a trend time-series $\mathbf{s}\in\chi$. 
We follow the recent work of \cite{tashiro2021csdi} to define our conditional diffusion model, and 
we show the training procedure in Algorithm~\ref{alg:diff_trend_training}.
At each training iteration, a trend $\boldsymbol{s}$, extracted 
directly from the input time-series $\bx_0 \sim q(\bx_0)$, and is used to condition the generative model. 
The trend  $\boldsymbol{s}$ can be any interpolation of the input time-series $\bx_0$.
In our experiments, during training we compute the trend by dividing the time-series in two, 
and fitting each half with a linear interpolation. We combine the linear interpolations
to obtain a very simple trend $\boldsymbol{s}$, and retain the model from just copying the trend.
During inference, we test the ability of the model to generalize using instead a low-order (i.e., 3) polynomial approximation. 
In Figure~\ref{fig:trend_example} we show an example of time-series and its trends, used respectively for training and inference, while in Section~\ref{sec:trend_exp} we show some examples of generated time-series.

\begin{figure}[hbt]
\begin{minipage}[t]{0.51\textwidth}
\begin{algorithm}[H]
  \caption{ \mbox{Trend-Constrained \textit{DiffTime} Training}} \label{alg:diff_trend_training}
\begin{algorithmic}[1]
   \STATE \textbf{Input:} input TS distribution $q(\bx_0)$, number of epochs $\mathbf{E}$
   \STATE \textbf{Output:} trained diffusion function $\rvepsilon_\theta$ 
   \FOR{$i=1$ \textbf{to} $\mathbf{E}$}
   \STATE $t \sim \textrm{Uniform}(\{1,\ldots,T\})$
   \STATE $\bx_0 \sim q(\bx_0) \ $; $\ \rvepsilon \sim \gN(\bmzero,\bmI)$
   \STATE $\boldsymbol{s}$ = poly-interpolation($\bx_0$)
    \STATE Take gradient step on \\  $\grad_\theta \left\| \rvepsilon - \rvepsilon_\theta(\sqrt{\hat{\alpha}_t} \bx_0 + \sqrt{1-\hat{\alpha}_t}\rvepsilon, t | \boldsymbol{s}) \right\|^2$
\ENDFOR
\end{algorithmic}
\end{algorithm}
\end{minipage}
\hfill
\begin{minipage}[t]{0.51\textwidth}
\begin{algorithm}[H]
  \caption{Trend-Constrained \textit{DiffTime} Sampling} \label{alg:diff_time_trend_sampling}
\begin{algorithmic}[1]
   \STATE \textbf{Input:} trained diffusion function $\rvepsilon_\theta$, trend $\boldsymbol{s}$
    \STATE \textbf{Output:} synthetic time-series $\hat{\bx}$    
    \STATE $\hat{\bx}_{T} \sim \mathcal{N}(\bmzero, \bmI)$
   \FOR{$t=\mathbf{T}$ \textbf{to} $1$}
    \STATE  $\boldsymbol{z} \sim \mathcal{N}(\bmzero, \bmI)$ if $t > 1$, else $\boldsymbol{z} = \bmzero$
\vspace{0.01em}
      \STATE \mbox{$\hat{\bx}_{t-1} = \frac{1}{\sqrt{\alpha_t}}\left( \hat{\bx}_t - \frac{1-\alpha_t}{\sqrt{1-\hat{\alpha}_t}} \rvepsilon_\theta(\hat{\bx}_t, t | \boldsymbol{s}) \right) + \sigma_t \boldsymbol{z}$}
\vspace{0.01em}
\ENDFOR
\vspace{0.3em}
\STATE \textbf{Return} $\hat{\bx}_0$ 
\vspace{0.55em}
\end{algorithmic}
\end{algorithm}
\end{minipage}
\end{figure}

\paragraph{Fixed-Value Constraint.} To guarantee the \textit{fixed point} constraints, which are hard constraints, 
we modify the \textit{reverse process} of \textit{DiffTime}, and we explicitly include them in the latent variables $\bx_{1:T}$.
The \textit{reverse process} is shown in Algorithm~\ref{alg:diff_time_fixed_values_sampling}, while the training 
procedure remains the standard one, shown in Algorithm~\ref{alg:diff_time_training}.
The sampling algorithm shows that at each diffusion step $t$ we explicitly enforce the fixed-points values
 in the noisy time-series $\hat{\bx}_t$, such that, 
$\forall \ r_{i,j} \in \mathcal{R}, \ \hat{x}_{i,j} = r_{i,j}$ where $\hat{x}_{i,j} \in \hat{\bx}_t$. 
This approach guarantees that the generated time-series have the desired fixed-point values --- 
in fact the fixed-values are enforced also for $t=0$ into $\hat{\bx}_0$ which is our final synthetic time-series.
By enforcing these fixed point at each iteration, 
we empirically found that the diffusion process better adapts the synthetic time-series to incorporate them. Figure~\ref{fig:fixed_example_ts} shows some examples of generated time-series with fixed-values constraint.

\begin{algorithm}[htb]
  \caption{\textit{DiffTime} Sampling - Fixed-Values Constraint} \label{alg:diff_time_fixed_values_sampling}
\begin{algorithmic}[1]
   \STATE \textbf{Input:} trained diffusion function $\rvepsilon_\theta$, fixed-points constraints $\mathcal{R}$
    \STATE \textbf{Output:} synthetic time-series $\hat{\bx}$    
    \STATE $\hat{\bx}_{T} \sim \mathcal{N}(\bmzero, \bmI)$
   \FOR{$t=\mathbf{T}$ \textbf{to} $1$}
    \STATE  $\boldsymbol{z} \sim \mathcal{N}(\bmzero, \bmI)$ if $t > 1$, else $\boldsymbol{z} = \bmzero$
\vspace{0.01em}
      \STATE \mbox{$\hat{\bx}_{t-1} = \frac{1}{\sqrt{\alpha_t}}\left( \hat{\bx}_t - \frac{1-\alpha_t}{\sqrt{1-\hat{\alpha}_t}} \rvepsilon_\theta(\hat{\bx}_t, t) \right) + \sigma_t \boldsymbol{z}$}
    \FOR{$r_{i,j} \in \mathcal{R}$}
        \STATE $\hat{x}_{t-1,i,j} = r_{i,j}$
    \ENDFOR
\ENDFOR
\STATE \textbf{Return} $\hat{\bx}_0$ 
\vspace{0.55em}
\end{algorithmic}
\end{algorithm}

\subsection{Loss-DiffTime}
We now discuss how to introduce a loss penalty into the diffusion model presented in the previous section, to incorporate more complex constraints.
While the sampling algorithm is the same of Algorithm~\ref{alg:diff_time_sampling}, the training now incorporates a penalty term into the loss function:
\begin{equation}
 L(\theta) := \mathbb{E}_{t, \bx_0, \rvepsilon} \left[ { \left\| \rvepsilon - \rvepsilon_\theta(\bx_t, t )  \right\|^2} + \rho f_{c}(\hat{\bx}_0) \right] \
\end{equation}
where $\bx_t = \sqrt{\hat{\alpha}_t} \bx_0 + \sqrt{1-\hat{\alpha}_t}\rvepsilon$ and $\hat{\bx}_0 = \frac{1}{\sqrt{\alpha_t}}\left( \bx_t - \frac{1-\alpha_t}{\sqrt{1-\hat{\alpha_t}}}  \rvepsilon_\theta(\bx_t, t) \right)$. 
The function $ f_{c}$ represents any differentiable constraint we want to incorporate.

The training pseudo-code is reported in Algorithm~\ref{alg:loss_diff_time_training}.

\begin{algorithm}[H]
  \caption{ \mbox{\textit{Loss-DiffTime} Training}} \label{alg:loss_diff_time_training}
\begin{algorithmic}[1]
   \STATE \textbf{Input:} input TS distribution $q(\bx_0)$, number of epochs $\mathbf{E}$, differentiable constraint $f_c$, scale parameter $\rho$
   \STATE \textbf{Output:} trained diffusion function $\rvepsilon_\theta$ 
   \FOR{$i=1$ \textbf{to} $\mathbf{E}$}
   \STATE $t \sim \textrm{Uniform}(\{1,\ldots,T\})$
   \STATE $\bx_0 \sim q(\bx_0) \ $; $\ \rvepsilon \sim \gN(\bmzero,\bmI)$
    \STATE $\bx_t = \sqrt{\hat{\alpha}_t} \bx_0 + \sqrt{1-\hat{\alpha}_t}\rvepsilon$
    \STATE $\hat{\rvepsilon} = \rvepsilon_\theta(\bx_t, t)$
    \STATE Take gradient step on \\  $\grad_\theta \left\| \rvepsilon - \hat{\rvepsilon} \right\|^2 + \rho f_{c}\left(\frac{1}{\sqrt{\alpha_t}}(\bx_t - \frac{1-\alpha_t}{\sqrt{1-\hat{\alpha_t}}} \hat{\rvepsilon} ) \right) $
\ENDFOR
\end{algorithmic}
\end{algorithm}

\subsection{Guided-DiffTime}
While \textit{Loss-DiffTime} model is able to incorporate any constraint, it requires to train a new
diffusion function $\rvepsilon_\theta$ for any new constraint.  
To overcome this limitation, we introduce \textit{Guided-DiffTime} that does not require 
re-training for new constraints --- we train a single unconstrained diffusion model using
Algorithm~\ref{alg:diff_time_training}
and then we \textit{guide} this model during inference using a differentiable constraint $f_{c}$.
We show this \textit{guided} sampling procedure in Algorithm~\ref{alg:guidingddim2}.
In particular, at each diffusion step, we get gradients from the differentiable constraints to guide (condition)
the synthetic time-series. The parameter $\rho$ weights the constraint during the generative process. 

In Section~\ref{sec:carbon_footprint} we show how \textit{Guided-DiffTime} can dramatically reduce the carbon footprint, 
by reducing the computational resources needed to handle new constraints.

\begin{algorithm}[hbt]
    \caption{\textit{Guided-DiffTime}}
    \label{alg:guidingddim2}
    \begin{algorithmic}
        \STATE \textbf{Input:} trained diffusion function $\rvepsilon_\theta$, differentiable constraint $f_c : \mathcal{X} \rightarrow \RR$, scale parameter $\rho$
        \STATE \textbf{Output:} synthetic time-series $\bx_0$
        \STATE $\hat{\bx}_T \sim \gN(\bmzero,\bmI)$
           \FOR{$t=\mathbf{T}$ \textbf{to} $1$}
        \STATE $\hat\epsilon \gets \epsilon_{\theta}(\hat{\bx}_t, t)$
            \STATE $\hat\epsilon \gets \hat\epsilon - \rho \sqrt{1-\hat{\alpha}_t} \grad_{\hat{\bx}_t} f_c(\frac{1}{\sqrt{\hat{\alpha}_t}}(\hat{\bx}_t - \hat\epsilon \sqrt{ 1 - \hat{\alpha}_t}))$
            \STATE $\hat{\bx}_{t-1} \gets \sqrt{\hat{\alpha}_{t-1}} \left( \frac{\hat{\bx}_t - \sqrt{1-\hat{\alpha}_t} \hat{\epsilon}}{\sqrt{\hat{\alpha}_t}} \right) + \sqrt{1-\hat{\alpha}_{t-1}} \hat{\epsilon}$
        \ENDFOR
        \RETURN $\hat{\bx}_0$
    \end{algorithmic}
\end{algorithm}

\subsection{Modelling the diffusion function}
\begin{figure}[!h]
    \includegraphics[trim={1.8in 0in 0in 0in},width=1\textwidth]{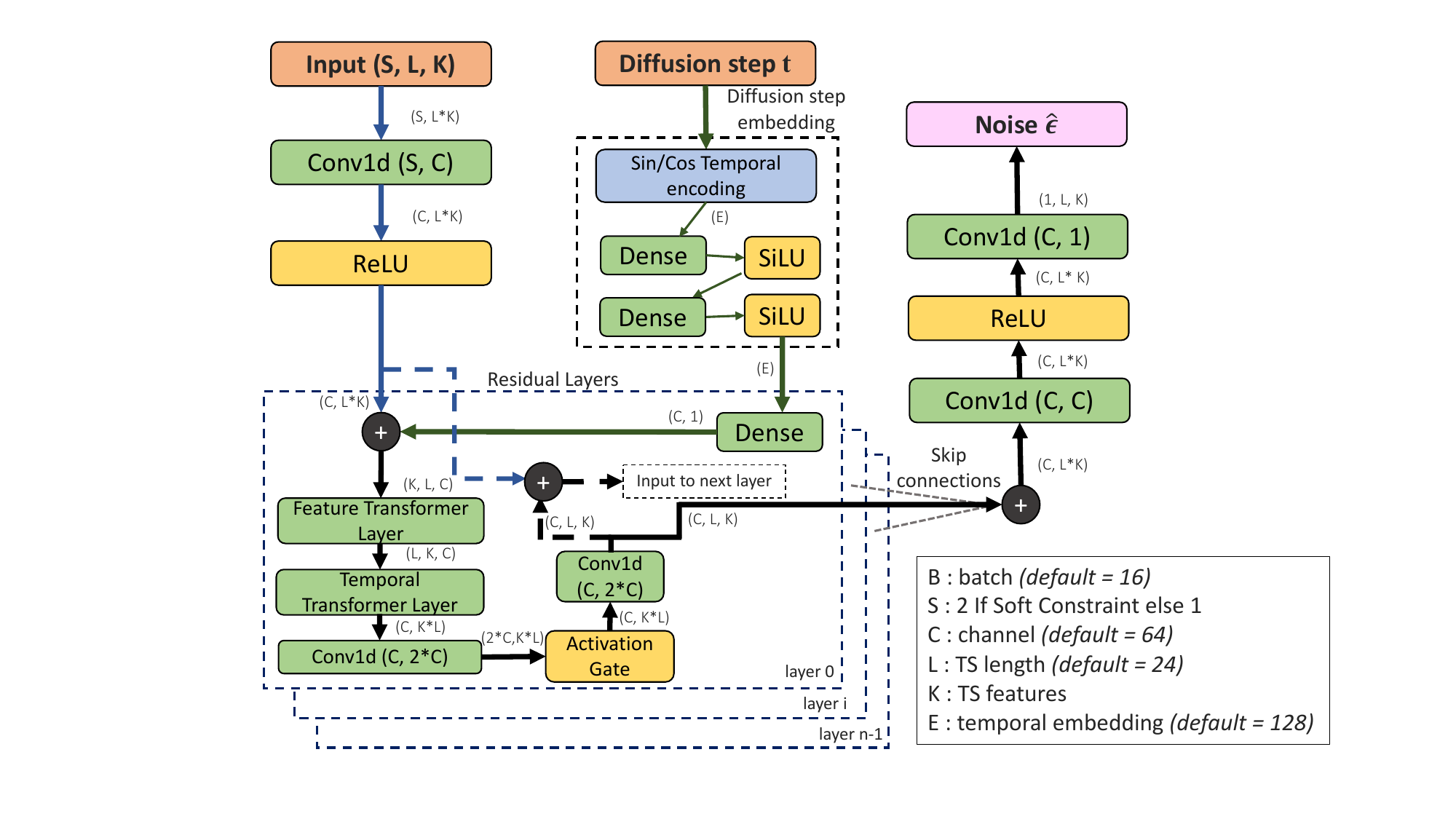}
    \caption{{\color{black} Diffusion model $e_{\theta}(\textbf{x}_t, t)$ architecture.}}
    \label{fig:diff_network}
\end{figure}

We approximate the diffusion function $\rvepsilon_\theta$ using a deep neural network, 
whose architecture is based on the groundbreaking work of~\cite{tashiro2021csdi,kong2020diffwave}. 
The architecture is composed by a 1-layer TransformerEncoder~\cite{paszke2019pytorch}, full-connected and 1d-Convolutional layers. The diffusion steps $t$ are encoded using a 128-dimensions embedding as proposed in previous work~\cite{tashiro2021csdi,vaswani2017attention,kong2020diffwave}. {\color{black} Figure~\ref{fig:diff_network} shows the neural network architecture.}

\
Since our architecture is mainly based on CSDI, we only discuss the main difference with respect to the original work~\cite{tashiro2021csdi}. In particular, we remove the \textit{side information} provided as embedding, and we incorporate all our conditionals along the input time-series $\bx$. In fact, our the conditional trend $\mathbf{s}$ has the same shape of the input time-series $\bx$. Thus, we can create an input Tensor with K features, L length, and C channels, where the first channel contains the conditional trend $\mathbf{s}$ and the second channel contains the input time-series $\bx$. We also change the kernel-size of the convolutional layers, which we found to be an important hyper-parameters to tune according the volatility and length of the input time-series. For example, sine data of length 24 requires a kernel size of $6$. Stock data requires kernel size $2$ for time-series of length 24, while the kernel size should be increased to $24$ for stock time-series with length $360$. 

For the noise level we use a \textit{Quadratic-Scheduler} which defines $\beta_t$ as follows:
$$
\beta_t = \left( \sqrt{\beta_1} + t \cdot \frac{\sqrt{\beta_T} - \sqrt{\beta_1}}{T-1} \right)^2
$$ 
where $T=50$ are the diffusion steps, and $\beta_1=1.0e-06$ and $\beta_T=0.5$.

\section{Experimental details and results} 
In this section, we report additional details about the experiments we show in the main body of the paper. 

\subsection{Unconstrained generation}\label{sec:unconstrained_append}
\begin{figure}[hbt]
\subcaptionbox{\textbf{COP-method}}{\includegraphics[width=0.19\textwidth]{images/unconditional/cop_stock.png}}%
\hfill
\subcaptionbox{\textbf{DiffTime}}{\includegraphics[width=0.19\textwidth]{images/unconditional/diff_tsne.png}}%
\hfill
\subcaptionbox{GT-GAN}{\includegraphics[width=0.21\textwidth]{images/unconditional/tsne_GT_GAN.png}}%
\hfill
\subcaptionbox{TimeGAN}{\includegraphics[width=0.19\textwidth]{images/unconditional/timegan.png}}%
\hfill
\subcaptionbox{RCGAN}{\includegraphics[width=0.19\textwidth]{images/unconditional/rcgan.png}}%
\hfill
\subcaptionbox{C-RNN-GAN}{\includegraphics[width=0.20\textwidth]{images/unconditional/crnngan.png}}%
\hfill
\subcaptionbox{T-Forcing}{\includegraphics[width=0.20\textwidth]{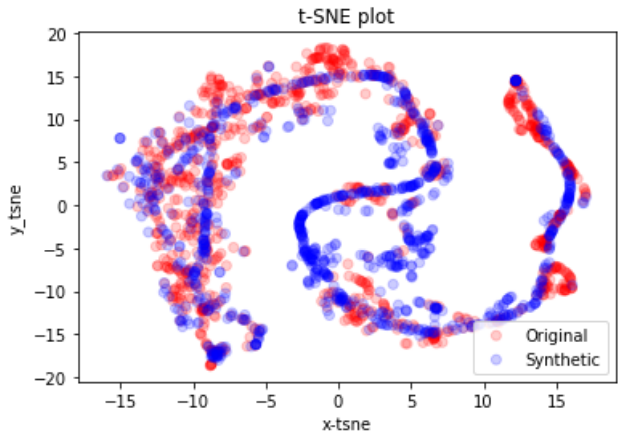}}%
\hfill
\subcaptionbox{P-Forcing}{\includegraphics[width=0.20\textwidth]{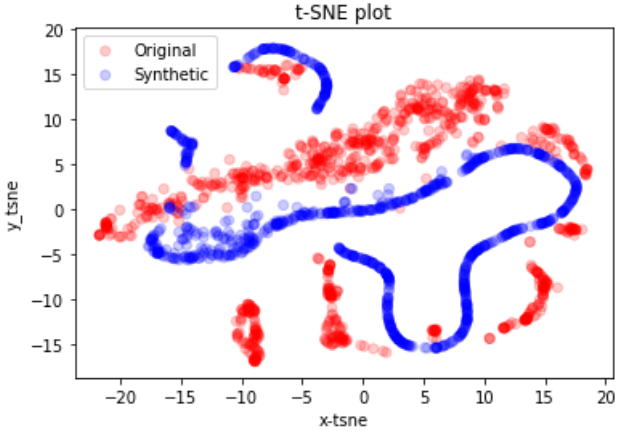}}%
\hfill
\subcaptionbox{WaveNET}{\includegraphics[width=0.20\textwidth]{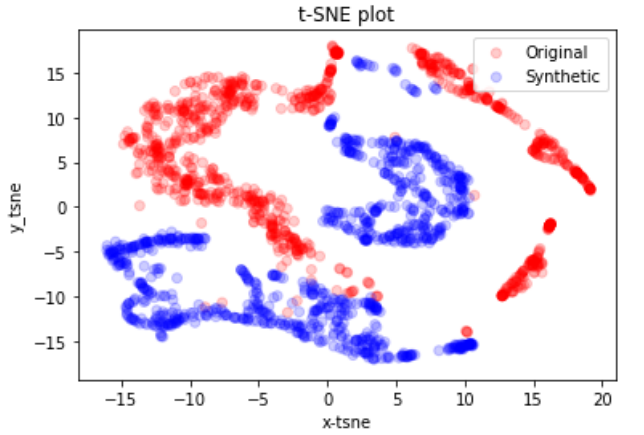}}%
\hfill
\subcaptionbox{WaveGAN}{\includegraphics[width=0.20\textwidth]{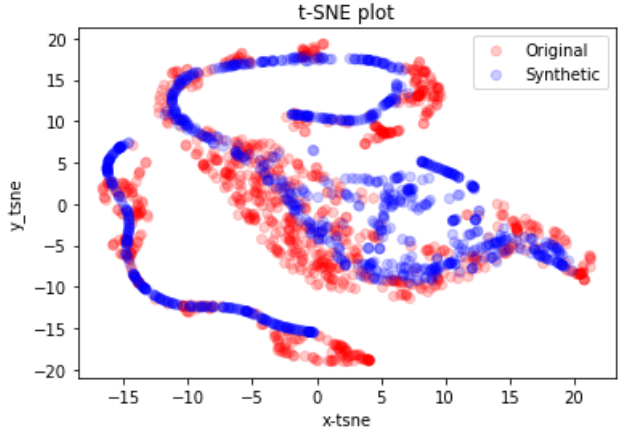}}%
\caption{t-SNE visualizations on multivariate stock data, where a greater overlap of blue and red dots shows a better distributional-similarity between the generated data and original data. Our approaches show the best performance.}\label{fig:tsnemin_unconstrained_main}
\end{figure}

In Figure~\ref{fig:tsnemin_unconstrained_main} we report the t-SNE analysis for all the approaches, which we omitted due to the limited space, in the main body of the paper. Notice that, to save computational resources, we do not recompute all the approaches but we use results from previous published work~\cite{timeGAN,jeon2022gt} for the same dataset (\textit{daily stocks}). The figure shows again that \textit{DiffTime} and \textit{COP-method} can generate realistic time-series beating existing benchmark algorithms. In particular, the figure shows that our approaches have significantly better performance with better overlap between red and blue samples.

\subsection{Trend constraint}~\label{app:trend_sine}
\begin{figure}[hbt]
\subcaptionbox{\scriptsize COP-method}{\includegraphics[width=0.20\textwidth]{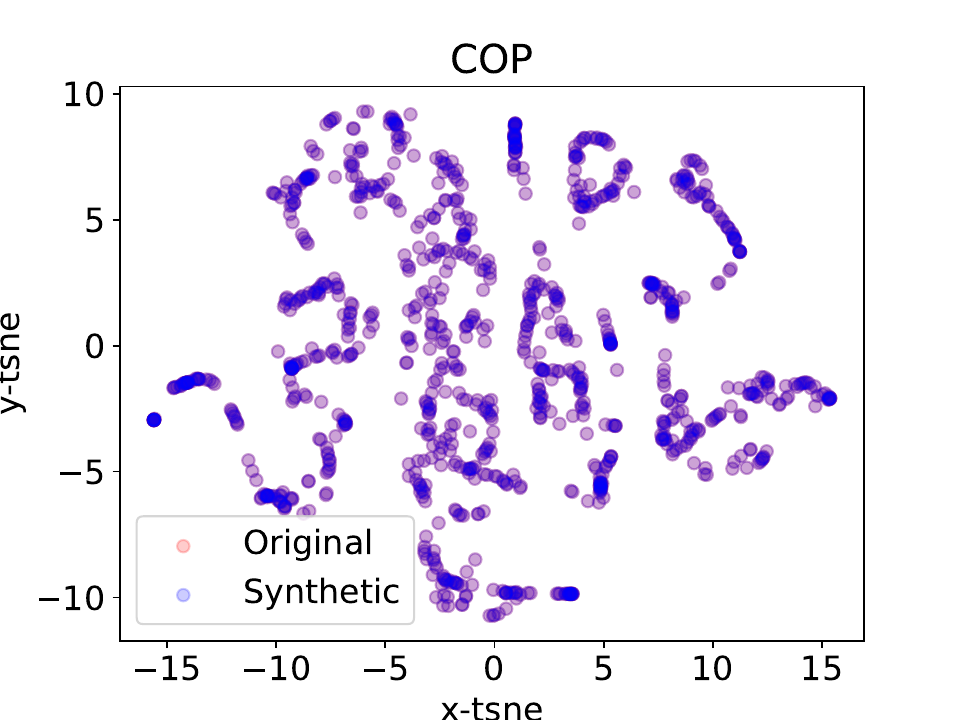}}%
\hfill
\subcaptionbox{\scriptsize DiffTime}{\includegraphics[width=0.20\textwidth]{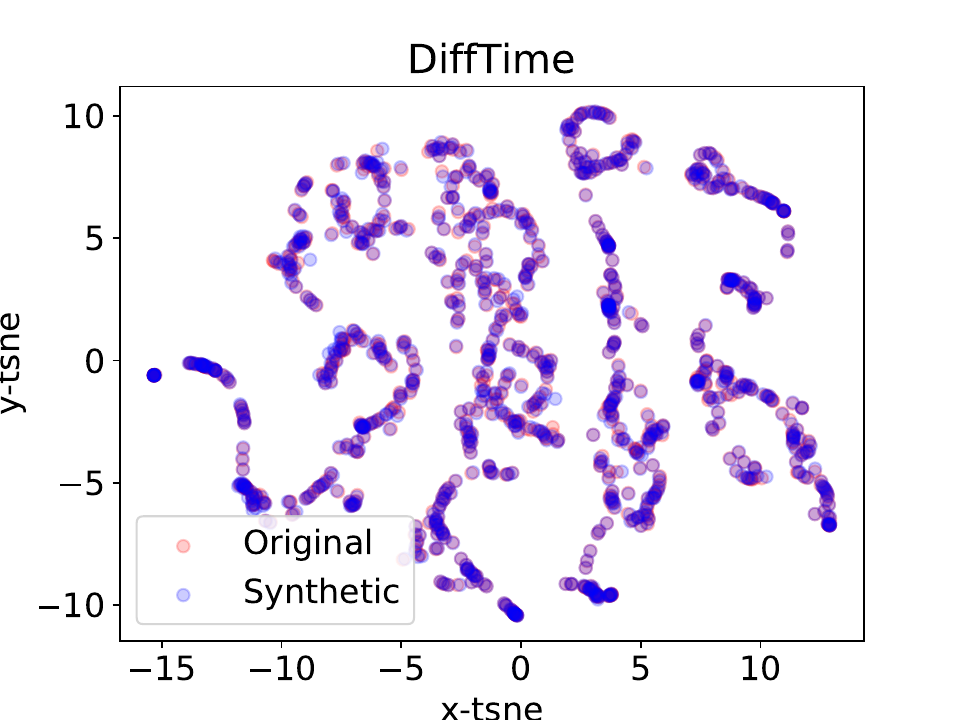}}%
\hfill
\subcaptionbox{\scriptsize GT-GAN}{\includegraphics[width=0.20\textwidth]{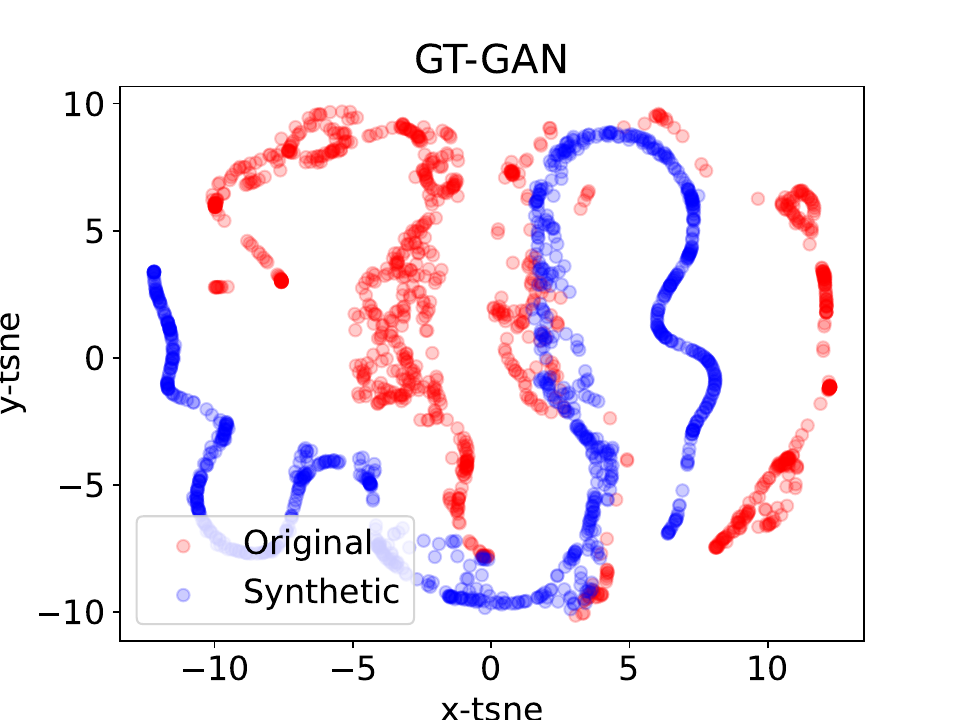}}%
\hfill
\subcaptionbox{\scriptsize TimeGAN}{\includegraphics[width=0.20\textwidth]{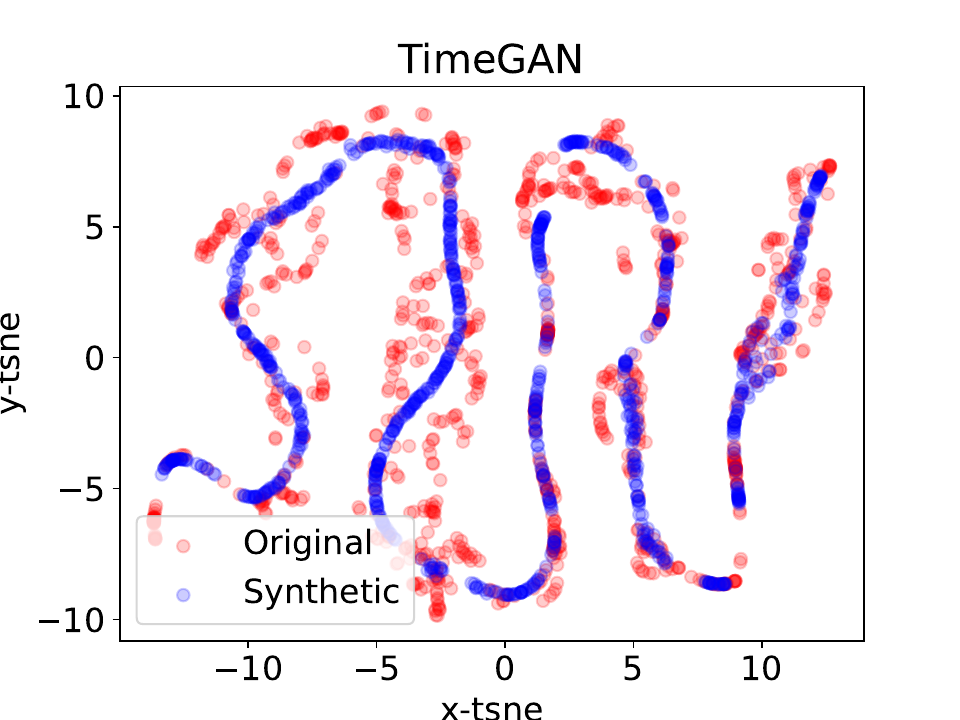}}%
\hfill
\subcaptionbox{\scriptsize RCGAN}{\includegraphics[width=0.20\textwidth]{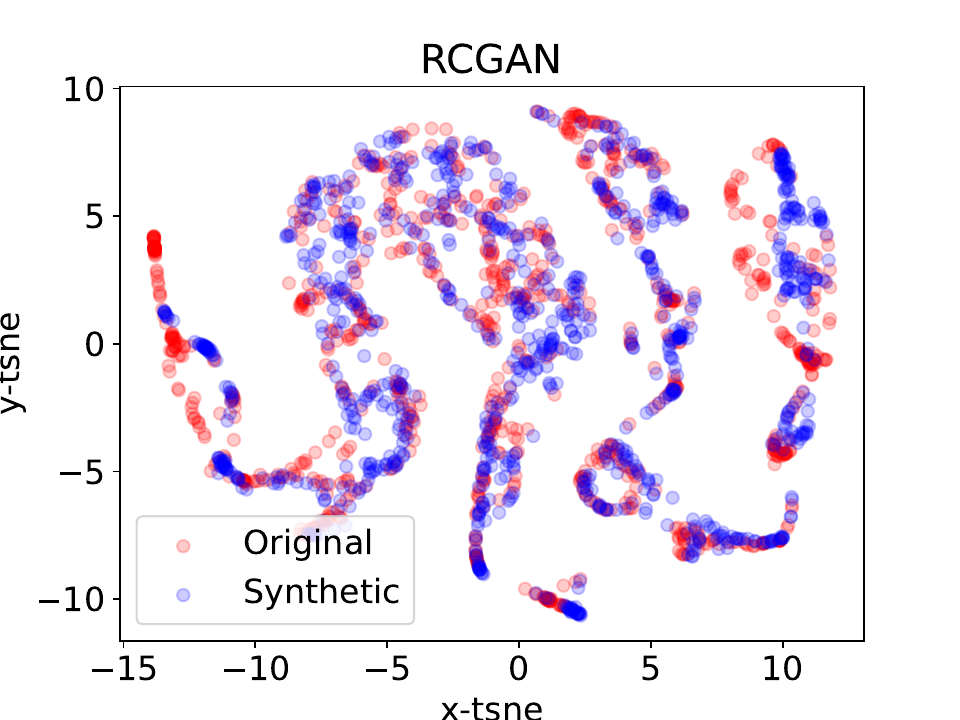}}
\caption{A t-SNE visualizations of \textit{Trend} constrained data, where a greater overlap of blue and red dots implies a better distributional-similarity between the generated data and original data. Our approaches show the best performance.}\label{fig:tsnemin_trend}
\end{figure}

We report in Figure~\ref{fig:tsnemin_trend} the t-SNE analysis which we omitted due to space limitations in the main body of the paper. This figure confirms the quantitative evaluation, with \textit{DiffTime} and \textit{COP-method} being the best models also in terms of covering the input distribution --- they show better overlap between red and blue dots.

\begin{figure}[hbt]
    \centering
    \includegraphics[width=1\textwidth]{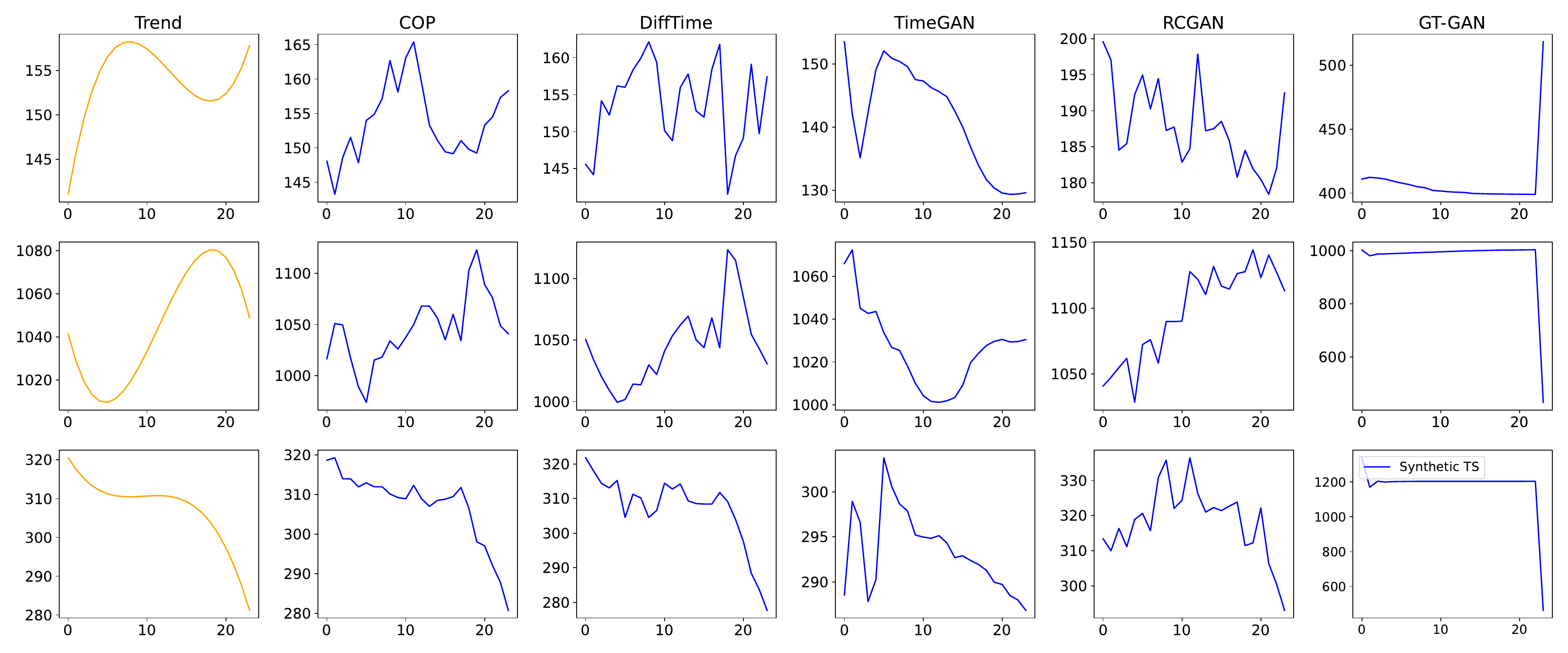}
    \caption{Example of Trend constraints and related synthetic time-series.}
    \label{fig:trend_example_syn}
\end{figure}

We report in Figure~\ref{fig:trend_example_syn} some example of the generated time-series, 
showing how our synthetic time-series are closer to the input trend. The figure also shows that \textit{GT-GAN} is only able to generate a very simple time-series matching just the upwards or downwards trend component.   

\begin{figure}[H]
\subcaptionbox{\scriptsize COP-method}{\includegraphics[width=0.20\textwidth]{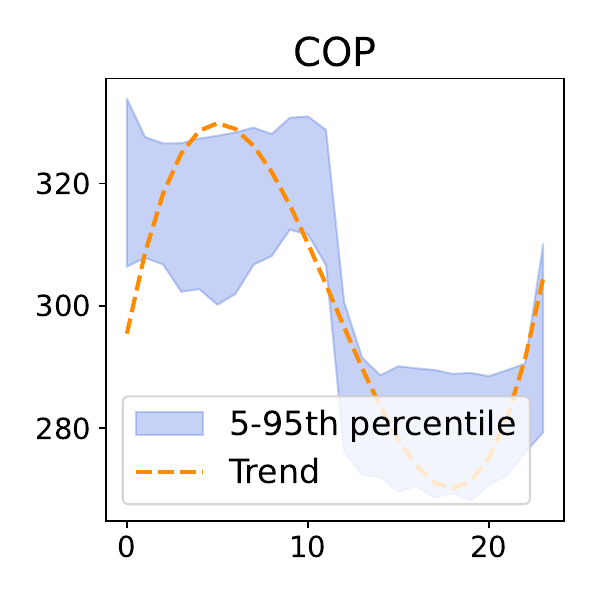}}%
\hfill
\subcaptionbox{\scriptsize DiffTime}{\includegraphics[width=0.20\textwidth]{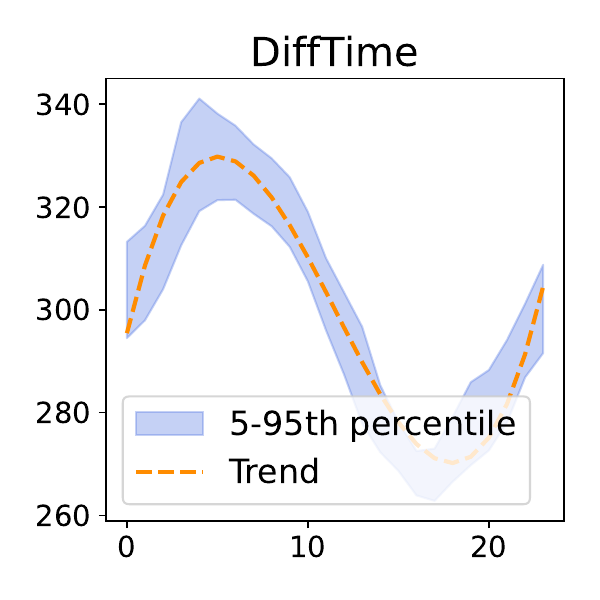}}%
\hfill
\subcaptionbox{\scriptsize GT-GAN}{\includegraphics[width=0.20\textwidth]{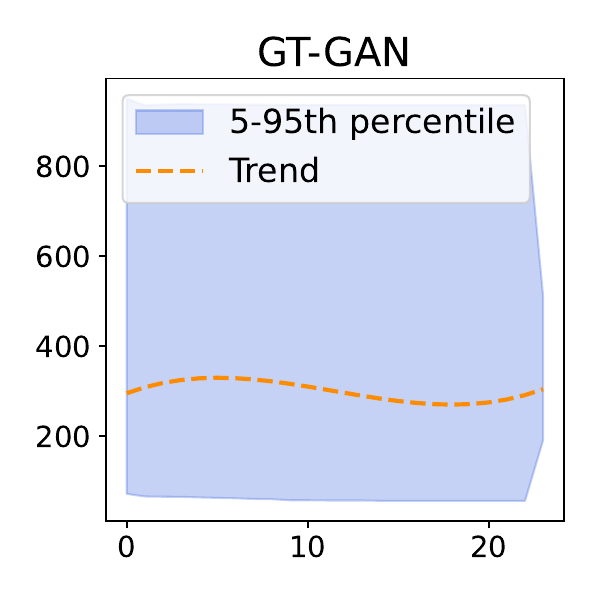}}%
\hfill
\subcaptionbox{\scriptsize TimeGAN}{\includegraphics[width=0.20\textwidth]{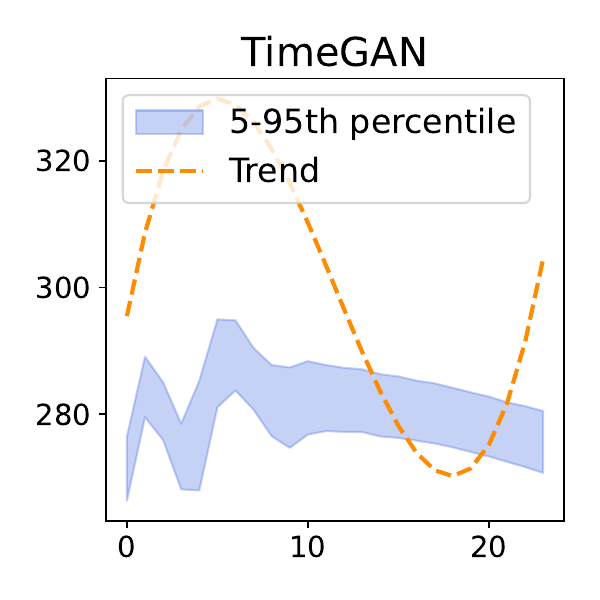}}%
\hfill
\subcaptionbox{\scriptsize RCGAN}{\includegraphics[width=0.20\textwidth]{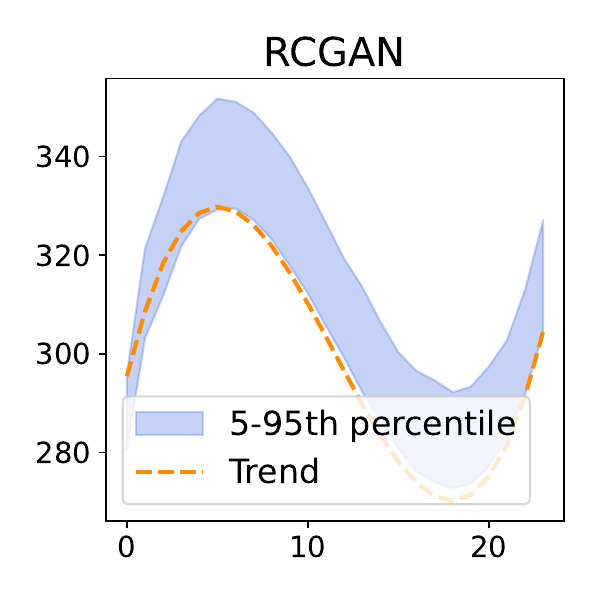}}%
\caption{A visualizations of \textit{Trend} constrained data, where the orange dotted time-series is the trend, and the shaded blue area shows the 5\% and 95\% percentiles of the generated synthetic time-series. Our approaches show the best performance with time-series closer to the input trend.}\label{fig:trend_perc}
\vspace{-0.1in}
\end{figure}

{\color{black}
\paragraph{\textbf{Sinusoidal Trend}} Finally we evaluate the case of a sinusoidal trend, i.e., the trend is provided as a sine wave, computed similarly to Eq.~\ref{eq:sine}. Considering the peculiar properties of a sinusoidal trend, i.e., its periodicity, we investigate additional metrics, including: \textbf{L2 distance} and \textbf{DTW distance} which measures how much the synthetic data follows a trend constraint by evaluating the distance between the TS and the trend using L2 norm and Dynamic Time Warping (DTW) approach~\cite{berndt1994using}, respectively; \textbf{Fourier distance} which applies a Fourier transformation and compares the basis of the periodic trend and the synthetic TS generated. 

Table~\ref{tab:trend_new} shows the evaluated quantitative metrics. The table confirms the results shown in the main body of the paper: our approaches achieve the best performance in terms of Discriminative and Predictive score; they also have the closest distance w.r.t. to the input trend. It is interesting to note that the DTW and spectral transformation techniques effectively capture any latent similarity patterns with the trends. For example, the spectral transformation highlights how some methods, like RCGAN, are able to somehow capture the trend even if shifted (which is also visible on Figure~\ref{fig:trend_perc}).

\begin{table}[h]
    \caption{Soft Constraints (Sinusoidal Trend) - Time-Series Generation}
    \label{tab:trend_new}
        \centering
        \resizebox{1\linewidth}{!}{
\begin{tabular}{l|l|l|l|l|l|l}
\hline
Algo & Discr-Score & Pred-Score & Inference-Time & L2 Distance & {\color{black} DTW Distance }
& {\color{black} Fourier-based distance}\\  \hline
COP (Ours) & \textbf{0.01±0.01} & \textbf{0.20±0.00} & 0.73 ± 0.05 & 46.3±32.9 & 35.8±25.8 & 0.57±0.57  \\
DiffTime (Ours) & \textbf{0.01±0.01} & \textbf{0.20±0.00} & 0.02 ± 0.00 & \textbf{35.57±16.99} &      \textbf{27.57±13.12} &  \textbf{0.49±0.57}
\\
GT-GAN & 0.04 ± 0.03 & 0.22 ± 0.00 & \textbf{0.00±0.00} & 1699.4±1253.1 &  1692.5±1253.9 &  1.74±2.51
\\
TimeGAN & 0.02 ± 0.02 & \textbf{0.20±0.00} & \textbf{0.00±0.00} &  121.35±61.30 &      87.29±50.25 &  1.06±1.11
\\ 
RCGAN & 0.02 ± 0.01 &\textbf{0.20±0.00}& \textbf{0.00±0.00} & 124.82±83.29 &      95.73±72.62 &  0.70±0.75
\\ \hline
\end{tabular}}
\end{table}

In Figure~\ref{fig:trend_perc} we fixed a trend for all the approaches, and we sample 1000 time-series to evaluate the generated time-series. The blue shaded area shows the 5\% and 95\% percentiles of the generated synthetic time-series.
}

\newpage
\subsection{Fixed-values constraint}\label{sec:trend_exp}
\begin{figure}[hbt]
\subcaptionbox{\scriptsize COP-method}{\includegraphics[width=0.20\textwidth]{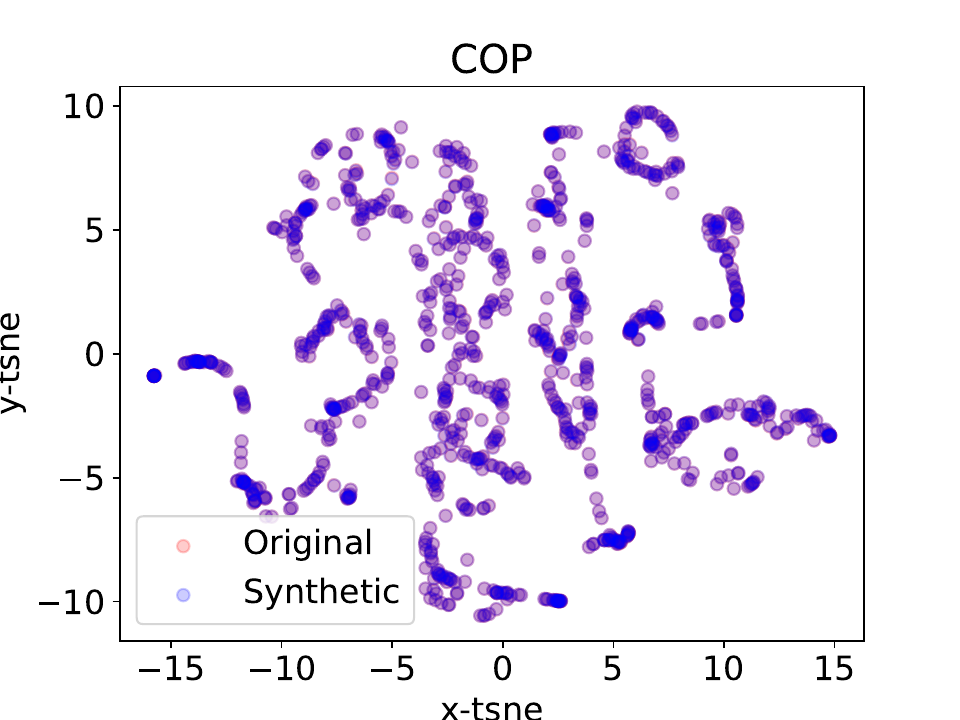}}%
\hfill
\subcaptionbox{\scriptsize DiffTime}{\includegraphics[width=0.20\textwidth]{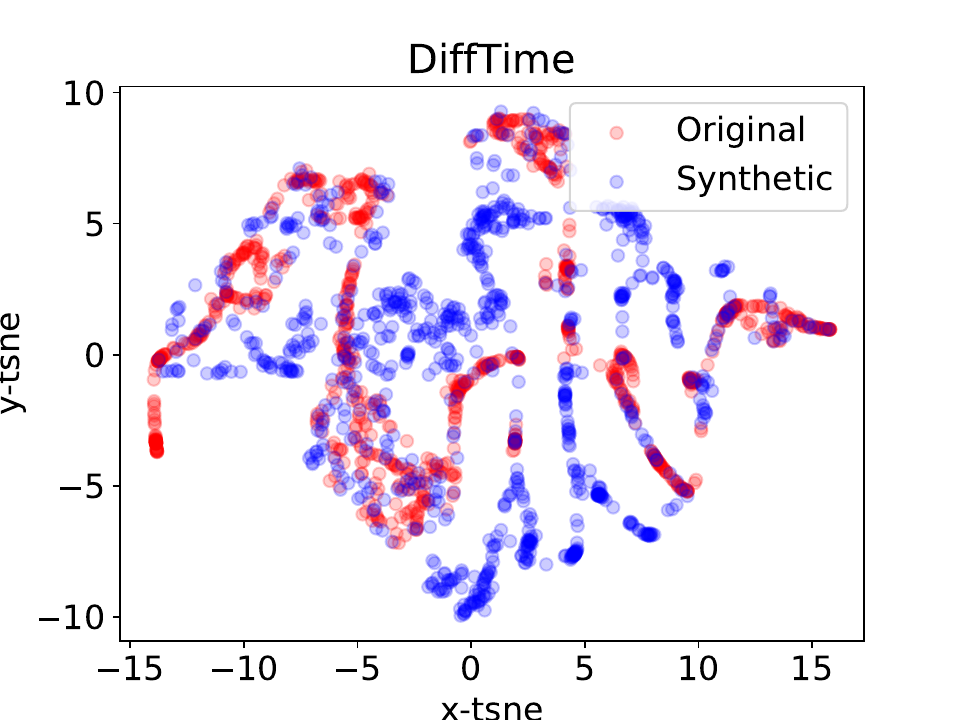}}%
\hfill
\subcaptionbox{\scriptsize GT-GAN}{\includegraphics[width=0.20\textwidth]{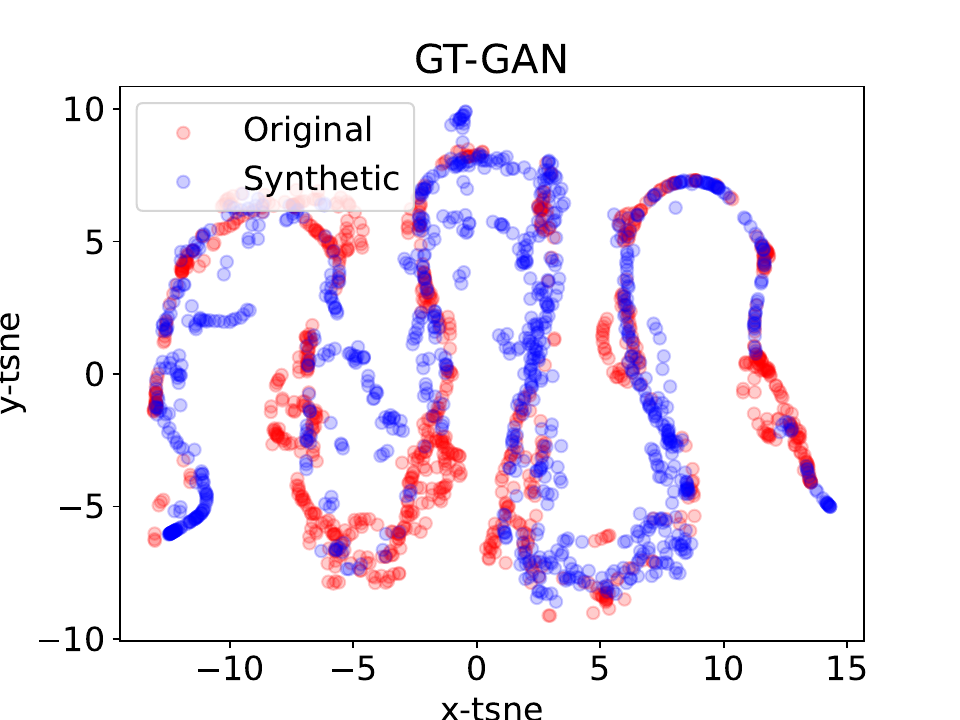}}%
\hfill
\subcaptionbox{\scriptsize TimeGAN}{\includegraphics[width=0.20\textwidth]{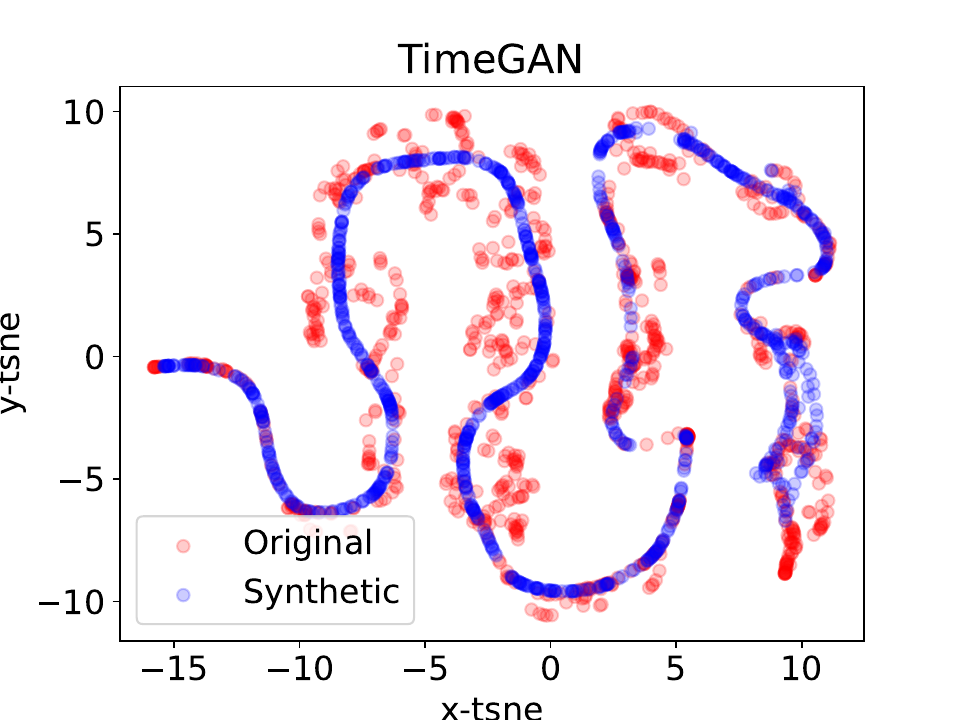}}%
\hfill
\subcaptionbox{\scriptsize RCGAN}{\includegraphics[width=0.20\textwidth]{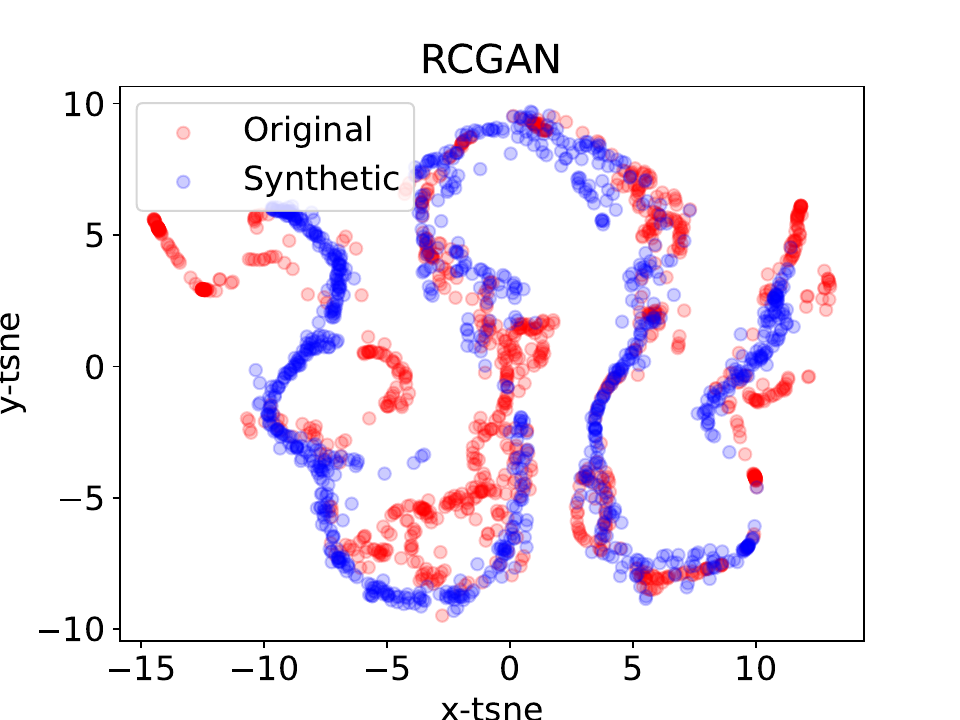}}
\caption{A t-SNE visualizations of \textit{Fixed-values} constrained data, where a greater overlap of blue and red dots implies a better distributional-similarity between the generated data and original data. Our approaches are among the best models.}\label{fig:tsnemin_fixed}
\end{figure}

For the fixed-values constraint we consider two fixed-value points at index $6$ and $18$ of the 
input time-series, which represent the points at $25\%$ and $75\%$ positions, respectively.  
We report in Figure~\ref{fig:tsnemin_fixed} the t-SNE analysis which we omitted due to space limitations in the main body of the paper. 
This figure confirms the quantitative evaluation, with \textit{DiffTime} and \textit{COP-method} being the 
best models also in terms of covering the input distribution --- they show better overlap between red and blue dots. In particular, we recall that while \textit{DiffTime} is not perfectly covering the input distribution, it always guarantee (i.e., 100$\%$ of the time) that the synthetic time-series pass through the two input fixed-points.

\begin{figure}[hbt]
    \centering
    \includegraphics[width=1\textwidth]{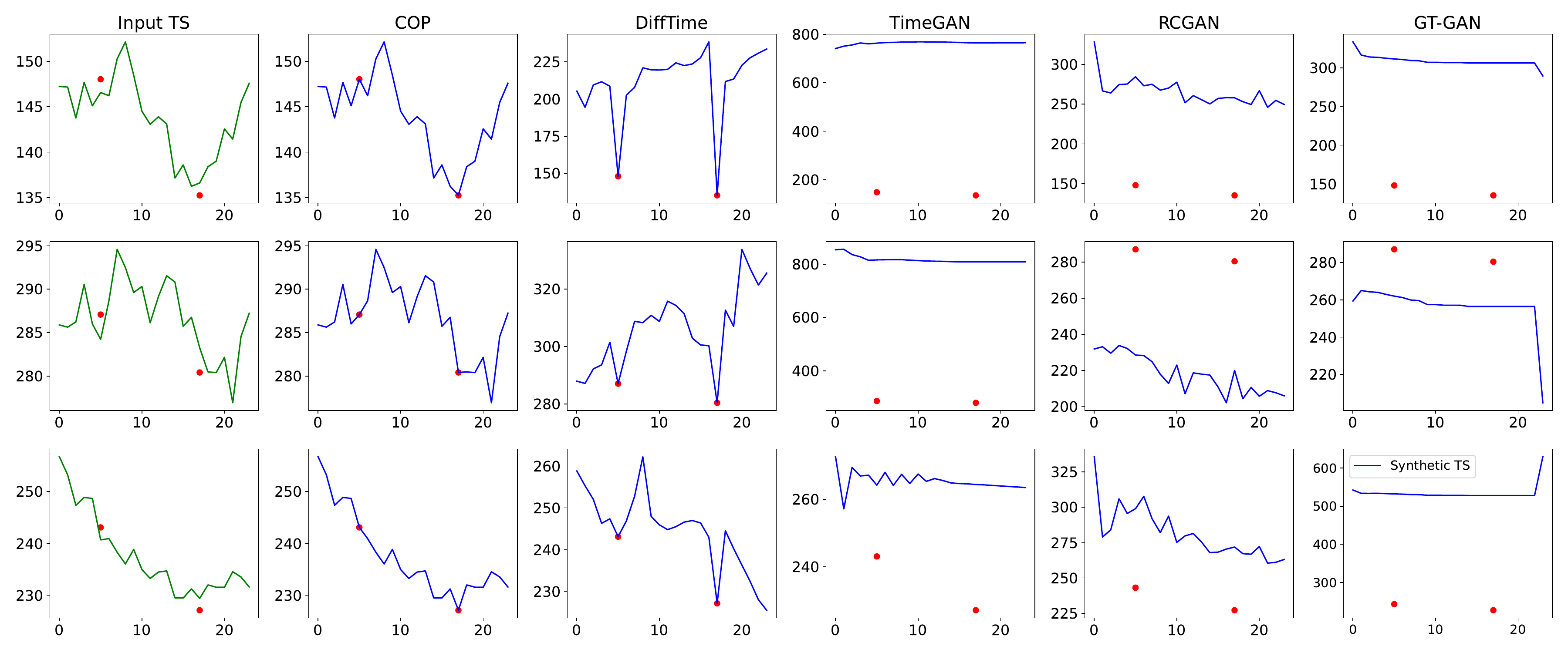}
    \caption{Example of Fixed-values constraint and related synthetic time-series.}
    \label{fig:fixed_example_ts}
\end{figure}

We report in Figure~\ref{fig:fixed_example_ts} some example of the generated time-series.
This picture highlights the ability of \textit{DiffTime} to generate reasonable time-series passing through the two fixed-points. COP-method shows the best results in this case, although it doesn't change the TS much from the input TS given to the COP-method. On the other hand, our DiffTime method does a better job of generating more different TS while satisfying the constraints.

\newpage
\subsection{Global Minimum constraint}
\begin{figure}[H]
\subcaptionbox{\scriptsize COP-method}{\includegraphics[width=0.32\textwidth]{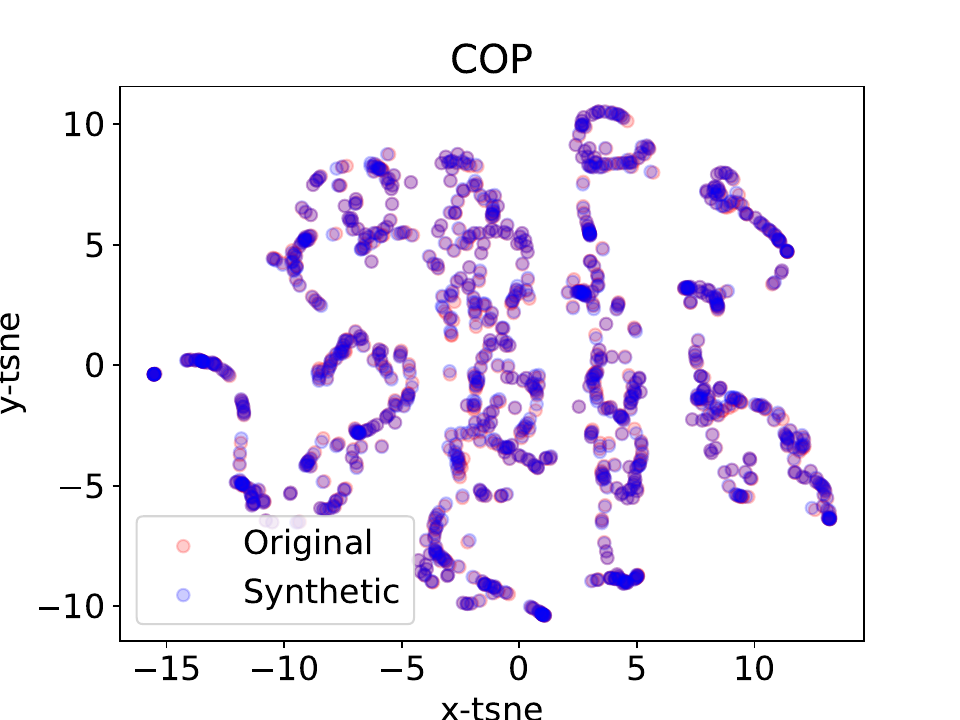}}%
\hfill
\subcaptionbox{\scriptsize Guided-DiffTime}{\includegraphics[width=0.32\textwidth]{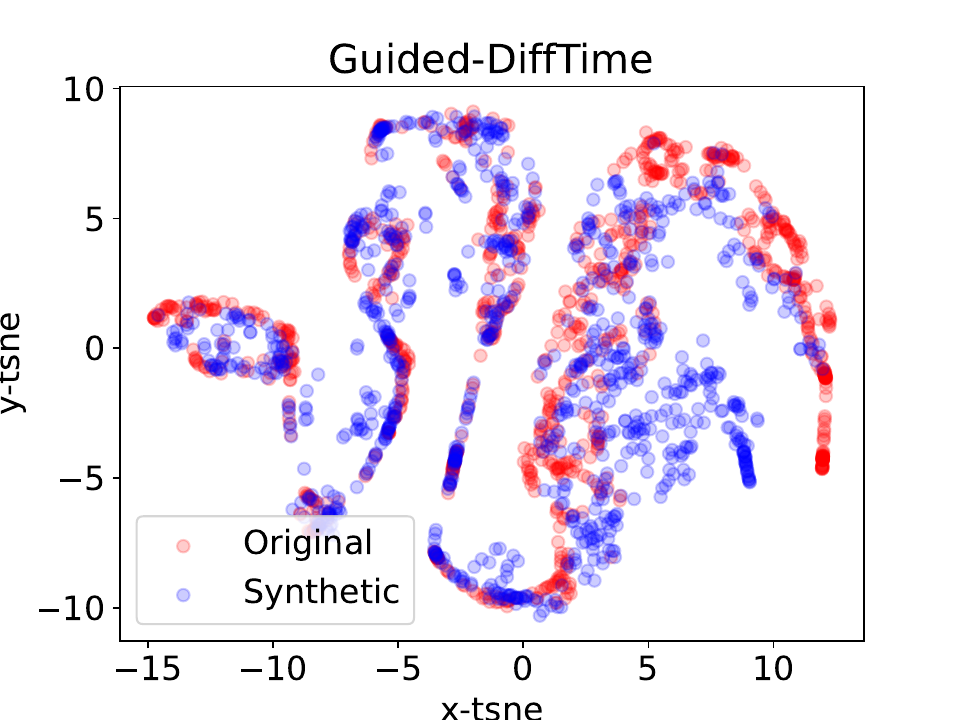}}%
\hfill
\subcaptionbox{\scriptsize Loss-DiffTime}{\includegraphics[width=0.32\textwidth]{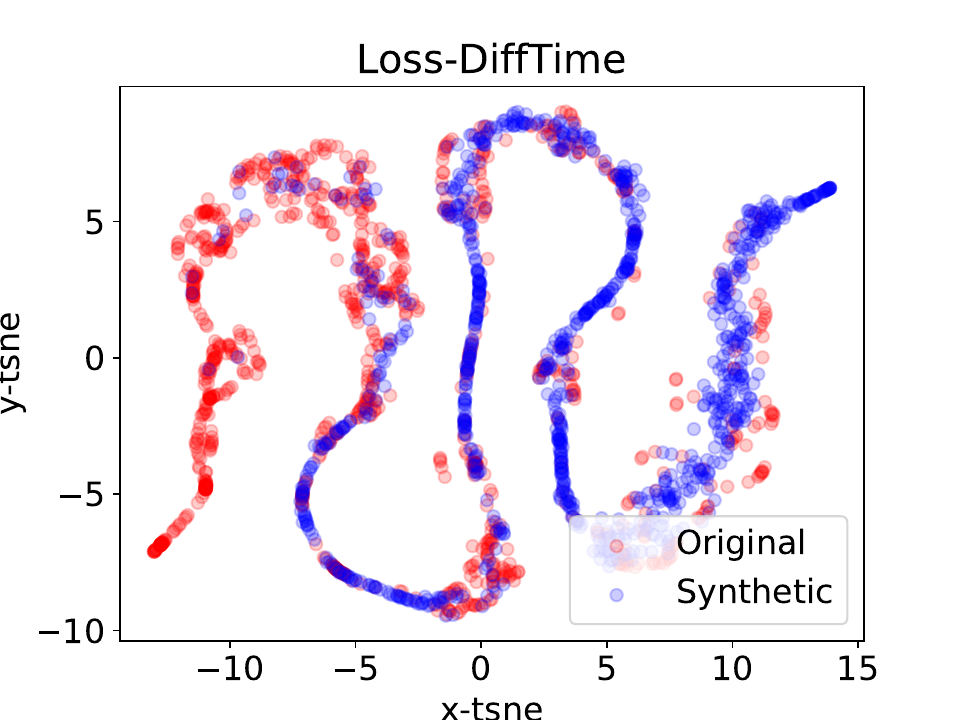}}%
\hfill
\subcaptionbox{\scriptsize GT-GAN}{\includegraphics[width=0.32\textwidth]{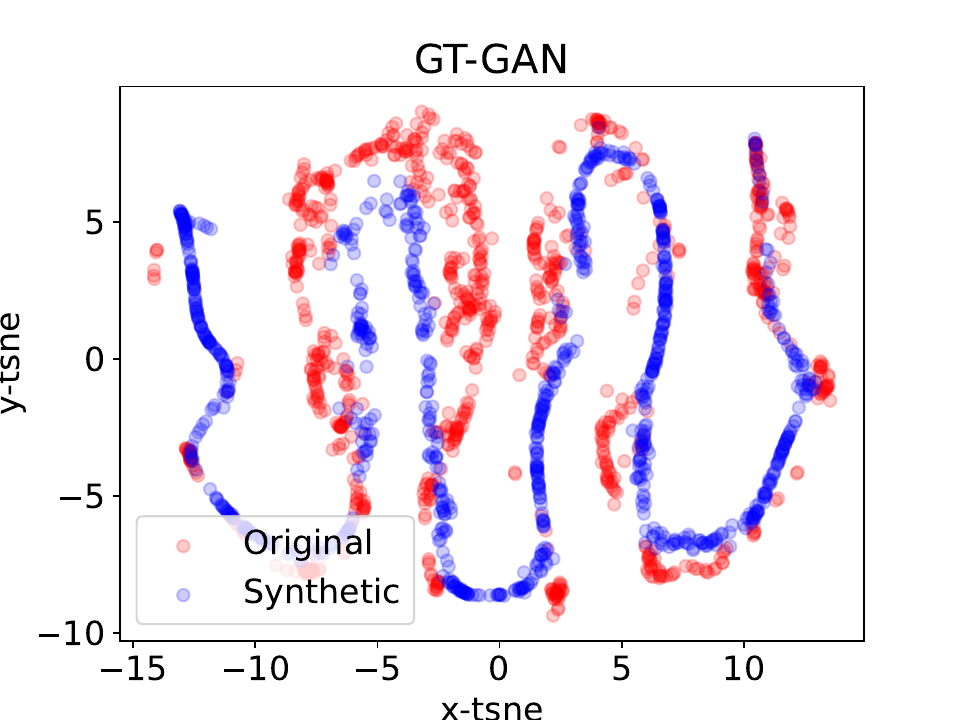}}%
\hfill
\subcaptionbox{\scriptsize TimeGAN}{\includegraphics[width=0.32\textwidth]{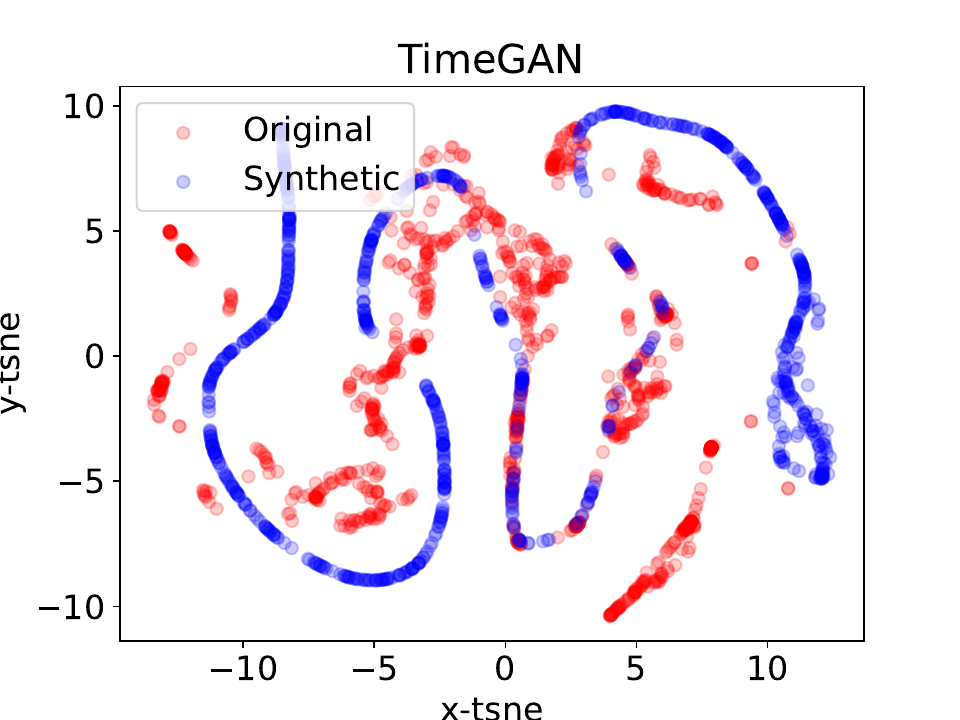}}%
\hfill
\subcaptionbox{\scriptsize RCGAN}{\includegraphics[width=0.32\textwidth]{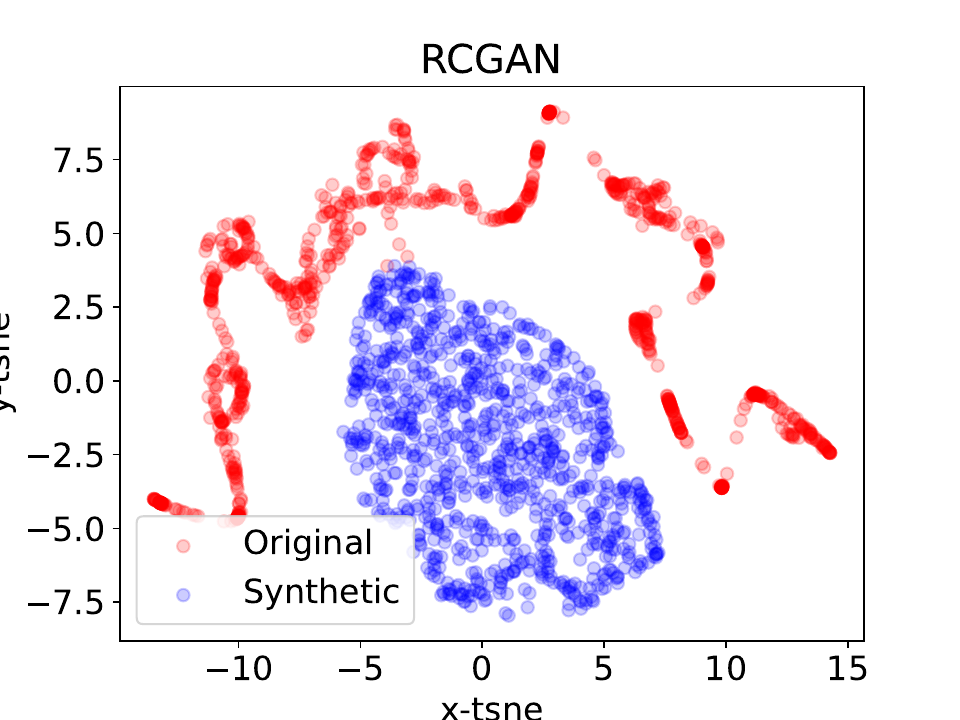}}
\caption{A t-SNE visualizations of \textit{Global-Min} constrained data, where a greater overlap of blue and red dots implies a better distributional-similarity between the generated data and original data.  Our approaches shows the best performance.}\label{fig:tsnemin_global_min}
\end{figure}

For the global minimum generation, we enfored the time-series to have a global minimum at index $10$. For \textit{Guided-DiffTime} we use $\rho=2$ while for \textit{Loss-DiffTime} we use $\rho=3.5$. In Figure~\ref{fig:tsnemin_global_min} we report the t-SNE analysis. 

\begin{figure}[H]
    \centering
    \includegraphics[width=1\textwidth]{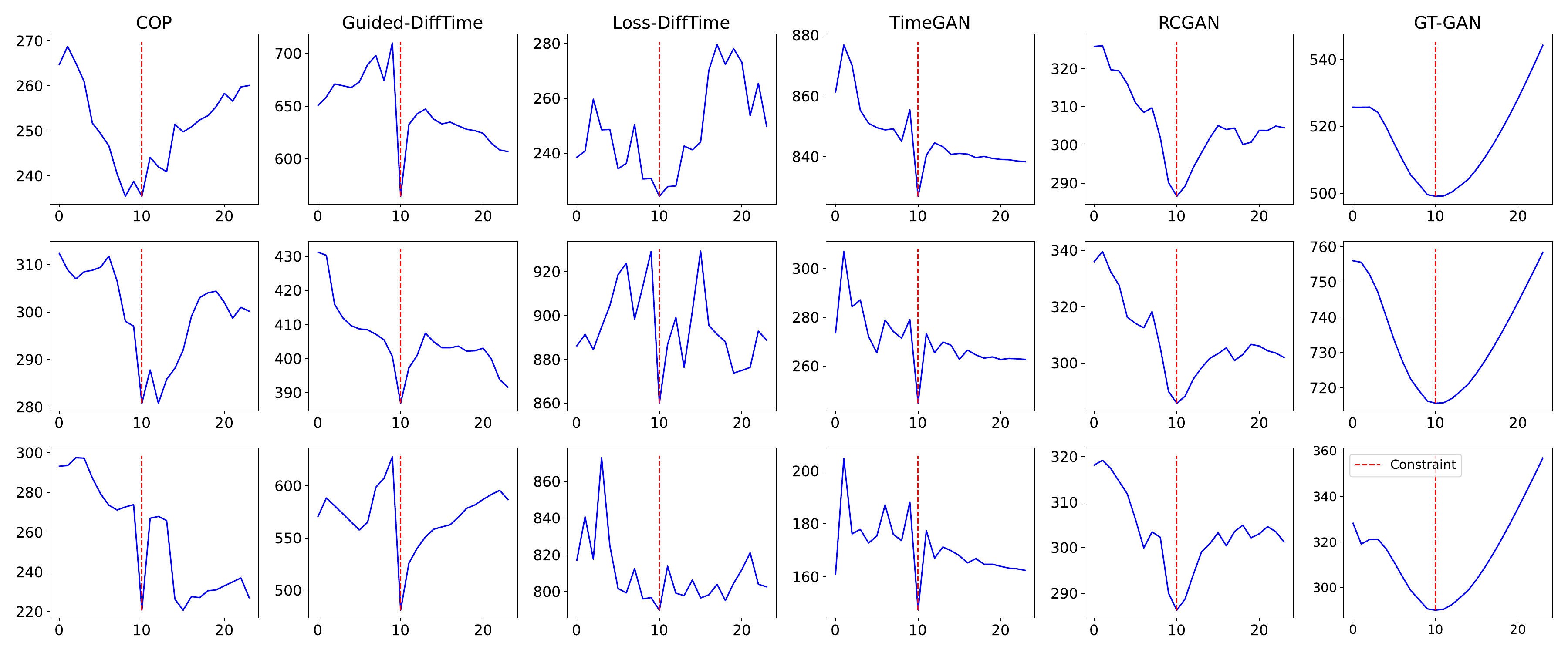}
    \caption{Example of Global-min constraint and related synthetic time-series.}
    \label{fig:global_min_example_ts}
\end{figure}

We report in Figure~\ref{fig:global_min_example_ts} some example of the generated time-series. While most of the approaches generate  synthetic time-series that respect the global minimum constraint, our methods better cover the input distribution (see  Figure~\ref{fig:tsnemin_global_min}), i.e., more fidelity in the generated data. In fact, in all the benchmarks the generated time-series are very similar, while our approaches have more diverse time-series.

\subsection{Multivariate constraint}\label{sec:ohlc}
\begin{figure}[hbt]
\subcaptionbox{\scriptsize COP-method}{\includegraphics[width=0.32\textwidth]{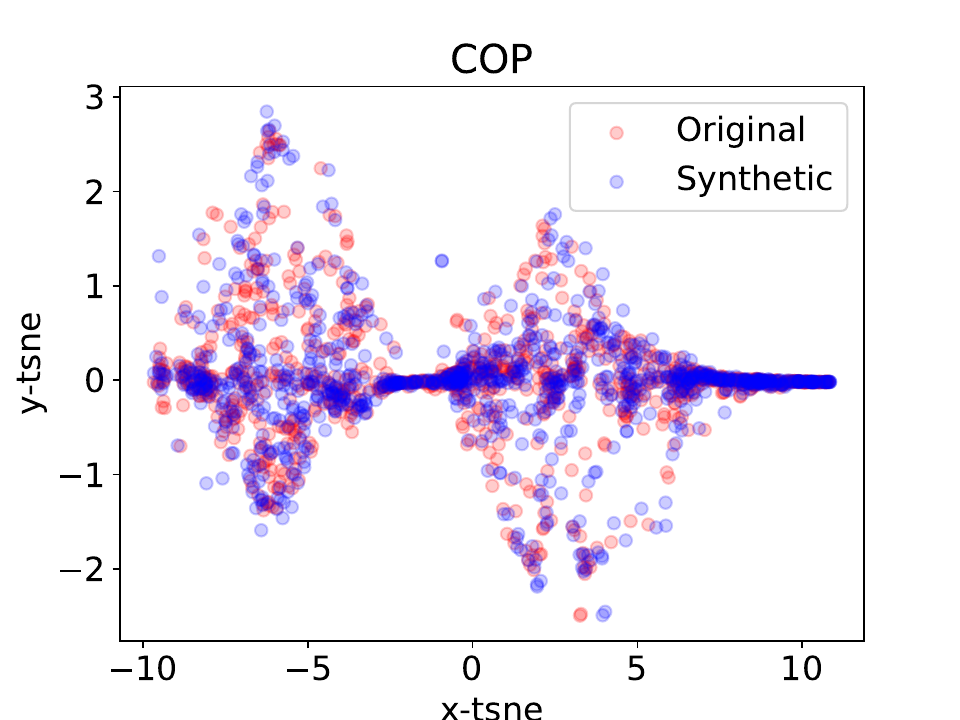}}%
\hfill
\subcaptionbox{\scriptsize Guided-DiffTime}{\includegraphics[width=0.32\textwidth]{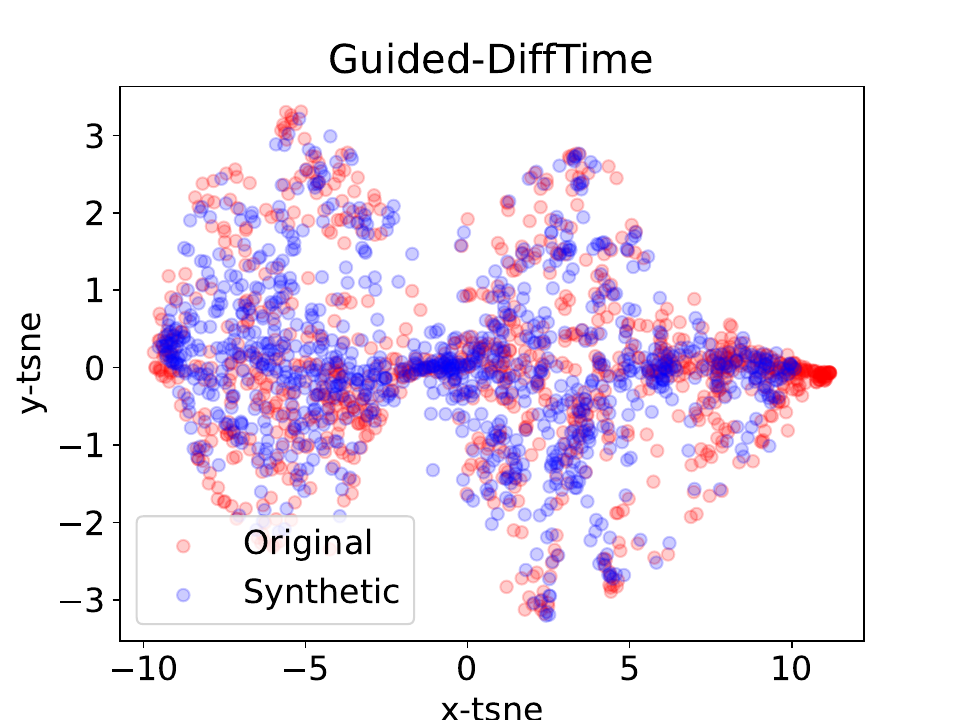}}%
\hfill
\subcaptionbox{\scriptsize Loss-DiffTime}{\includegraphics[width=0.32\textwidth]{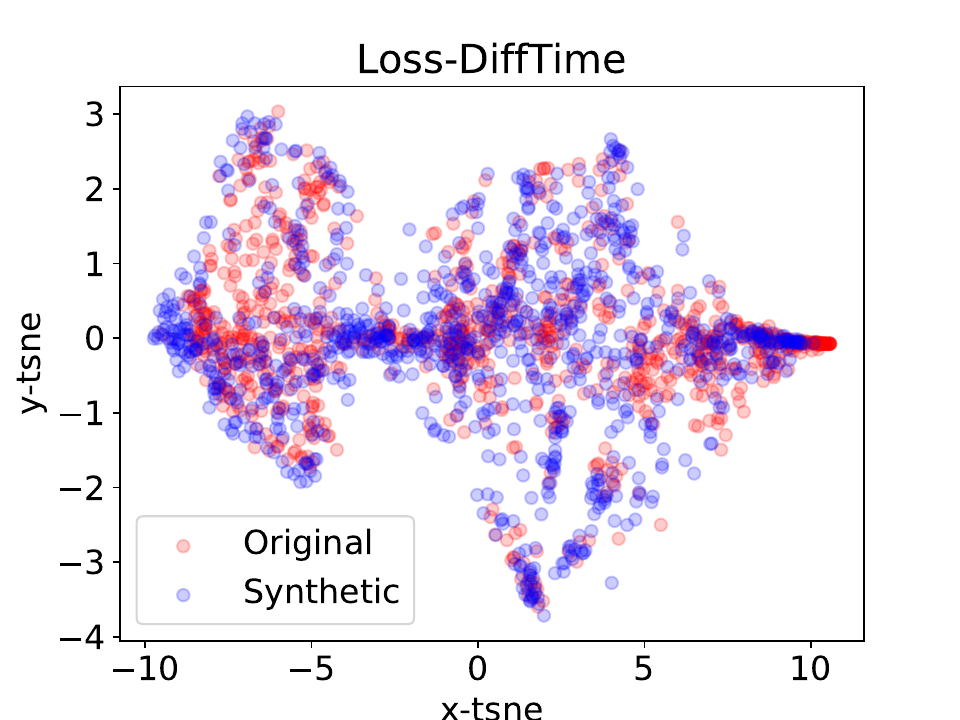}}%
\hfill
\subcaptionbox{\scriptsize GT-GAN}{\includegraphics[width=0.32\textwidth]{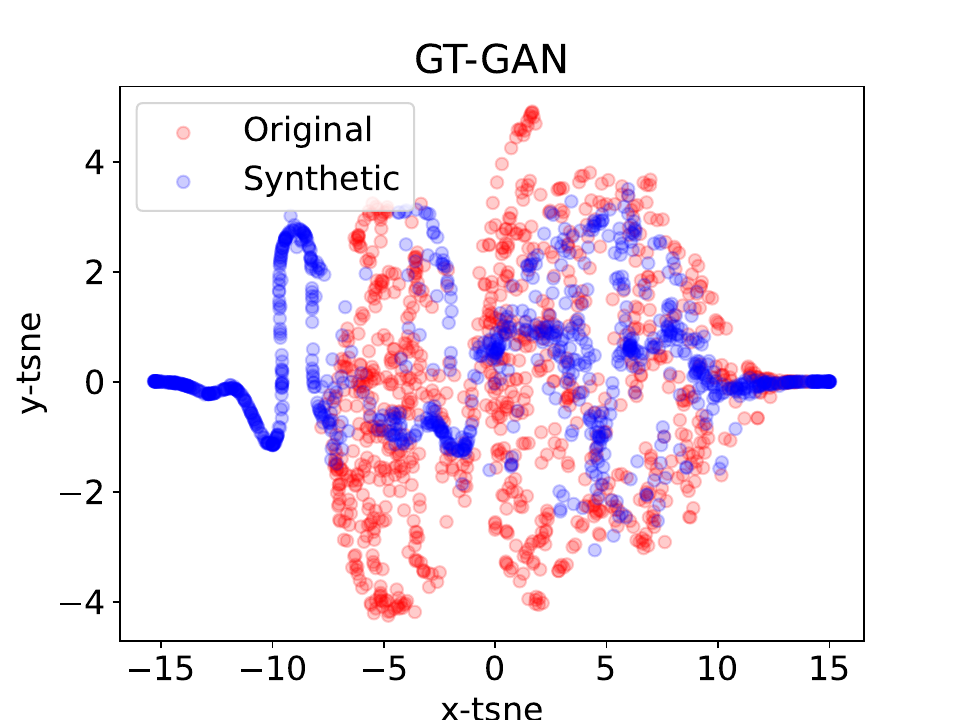}}%
\hfill
\subcaptionbox{\scriptsize TimeGAN}{\includegraphics[width=0.32\textwidth]{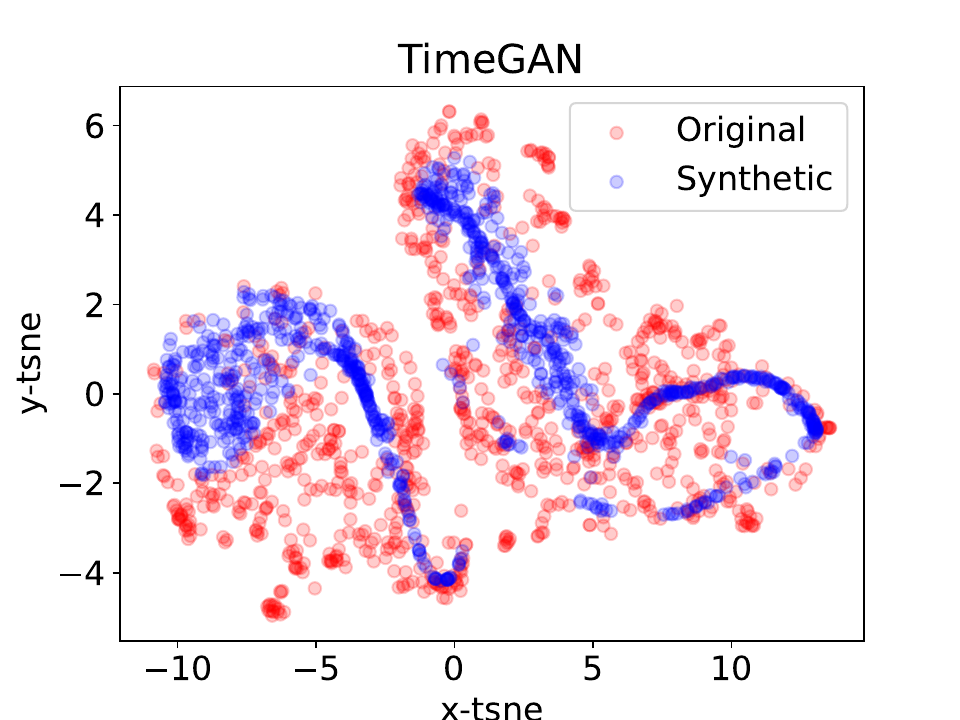}}%
\hfill
\subcaptionbox{\scriptsize RCGAN}{\includegraphics[width=0.32\textwidth]{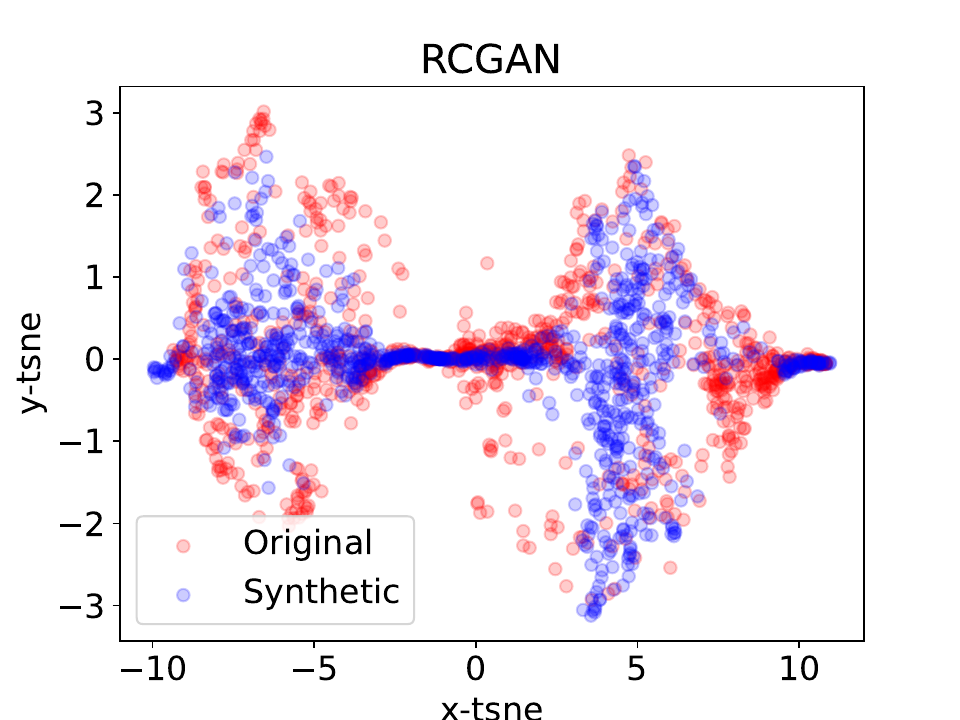}}
\caption{A t-SNE visualizations of \textit{Multivariate} constrained data, where a greater overlap of blue and red dots implies a better distributional-similarity between the generated data and original data. Our approaches shows the best performance.}\label{fig:tsnemin_ohlc}
\end{figure}

Finally, we report the multivariate constraint using the multivariate Google stock data. 
This constraint guarantees a well known financial data property where: 
the feature \textit{High} has always the highest value w.r.t. to the other features;
and the feature \textit{Low} has always the lowest value w.r.t. to the other features. For \textit{Guided-DiffTime} we use $\rho=0.001$ while for \textit{Loss-DiffTime} we use $\rho=3.5$.
We report in Figure~\ref{fig:tsnemin_ohlc} the t-SNE analysis which we omitted due to space limitations in the main body of the paper.
This figure confirms the quantitative evaluation, with \textit{DiffTime} and \textit{COP-method},
and our approaches show a better coverage of the input distribution, with a higher overlap between red and blue dots.

\begin{figure}[hbt]
    \centering
    \includegraphics[width=1\textwidth]{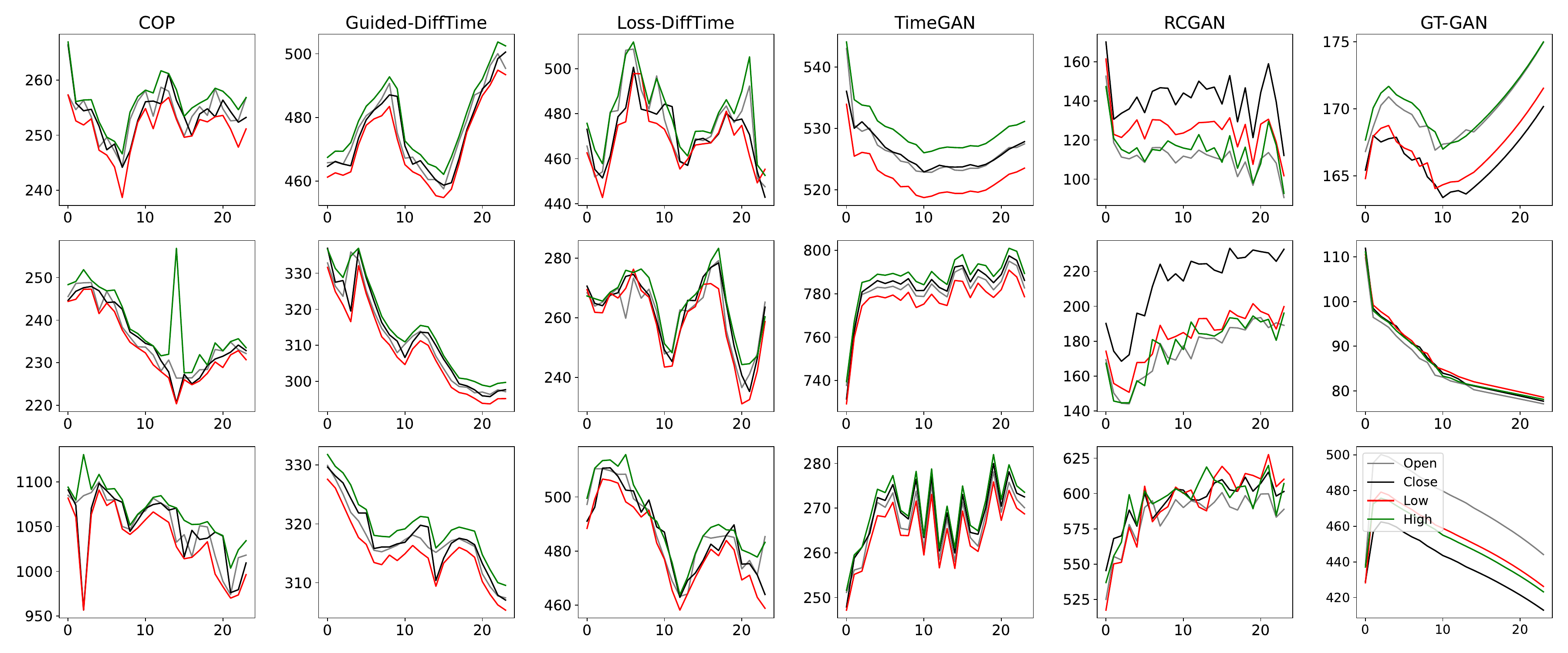}
    \caption{Example of multivariate constraint and related synthetic time-series.}
    \label{fig:ohlc_example_ts}
\end{figure}

We report in Figure~\ref{fig:ohlc_example_ts} some example of the generated time-series. In this case, it's worth noticing that \textit{Guided-DiffTime} and \textit{COP-method} have among the best performance, showing time-series that respect the multivariate  constraints (i.e., high feature has always the maximum value, while low feature is the lowest). The Figure also shows that the generated time-series from the GT-GAN have not exactly the common statistical properties of stock data~\cite{bouchaud2018trades}; while RCGAN and TimeGAN have a huge difference between High and Low features, which is unlikely in real data and in the training set. 

\section{Ablation Study}
In this section we carried out an ablation study of the proposed approaches. Where not otherwise stated, we consider univariate stock-data. 

\subsection{Diffusion steps}\label{sec:diff_step}
\begin{figure}[hbt]
\subcaptionbox{\scriptsize  DiffTime $T=50$}{\includegraphics[width=0.32\textwidth]{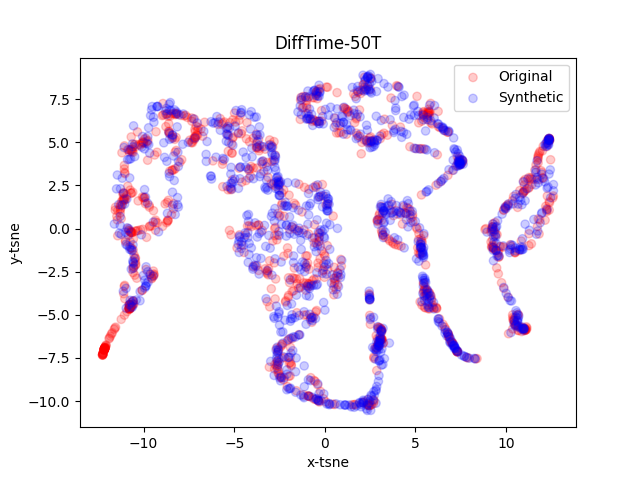}}%
\hfill
\subcaptionbox{\scriptsize  DiffTime $T=100$}{\includegraphics[width=0.32\textwidth]{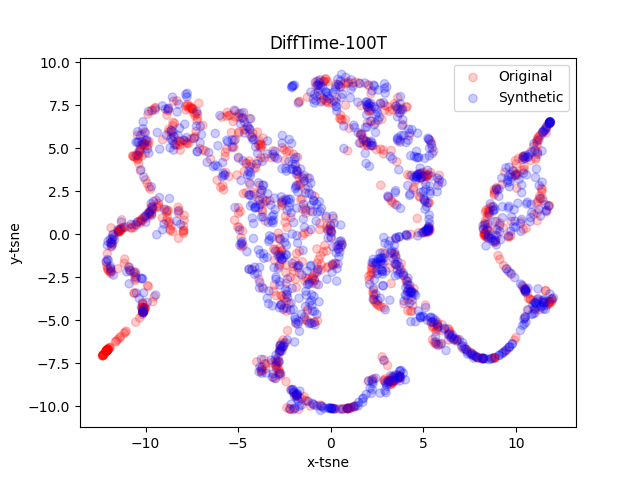}}%
\hfill
\subcaptionbox{\scriptsize DiffTime $T=200$}{\includegraphics[width=0.32\textwidth]{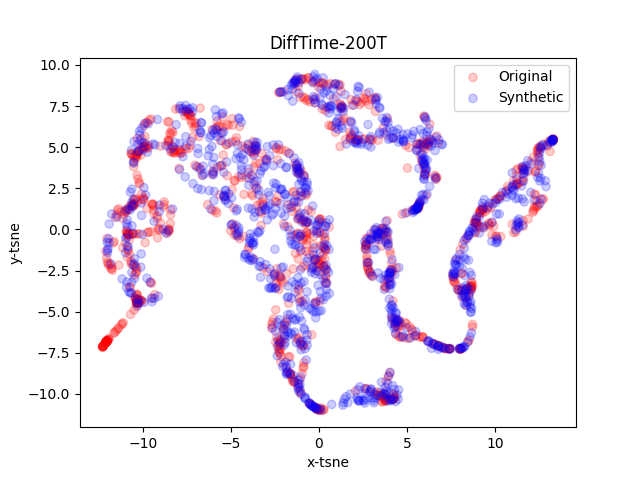}}
\caption{A t-SNE visualizations of \textit{DiffTime} for different diffusion steps $T \in [50,100,200]$.}\label{fig:tsen_steps}
\end{figure}

Here we evaluate the impact of a different number of diffusion steps in the diffusion models. We vary the diffusion steps using $T \in [50, 100, 200]$.
Figure~\ref{fig:tsen_steps} shows the t-SNE comparison for the different diffusion steps, which show all the same performance. Therefore, in all our experiments we considered the most economic setup of $T=50$.
In table~\ref{tab:T_vary} we evaluate the impact of the different diffusion steps in the model according the quantitative metrics. Also in this table, we notice that the increasing the diffusion steps do not improve the results.

\begin{table}[hbt]
\centering
\caption{ \textit{DiffTime} with different diffusion steps $T \in [50,100,200]$.}\label{tab:T_vary}
\begin{tabular}{llll}
\toprule
Algo & Discr-Score & Pred-Score & Inference-Time \\
\midrule
DiffTime $T=50$  &   \textbf{0.05±0.03} &  \textbf{0.21±0.00} &     \textbf{0.020±0.00} \\
DiffTime $T=100$ &    0.07±0.02  &  0.22±0.00 &     0.049±0.02 \\
DiffTime $T=200$ &    0.06±0.01 &  \textbf{0.21±0.00} &     0.091±0.01 \\
\bottomrule
\end{tabular}
\end{table}

\subsection{Noise Variance}\label{sec:noise_var}
Here we evaluate the impact of a different noise variance scheduler in the diffusion models.
We recall that we consider $T=50$ diffusion steps, and we set the minimum noise level $\beta_1=1.0e-06$ , the maximum level to $\beta_{T} = 0.5$. Following recent work in diffusion models~\cite{tashiro2021csdi,nichol2021improved,song2020denoising}, we define $\beta_t$ by consider the following schedulers:

$\bullet$ \textit{Linear-Scheduler}:
$$
\beta_t = \left( \beta_1 + t \cdot \frac{\beta_T - \beta_1}{T-1} \right)
$$

$\bullet$ \textit{Quadratic-Scheduler}:
$$
\beta_t = \left( \sqrt{\beta_1} + t \cdot \frac{\sqrt{\beta_T} - \sqrt{\beta_1}}{T-1} \right)^2
$$

$\bullet$ \textit{Cosine-Scheduler}:
$$
\beta_t = \beta_1 + (\beta_T - \beta_1) \cdot \frac{1}{2}\left(1 + cos\left(\frac{\pi * t}{T}\right)\right)
$$

\begin{figure}[hbt]
\subcaptionbox{\scriptsize  DiffTime-Quad}{\includegraphics[width=0.32\textwidth]{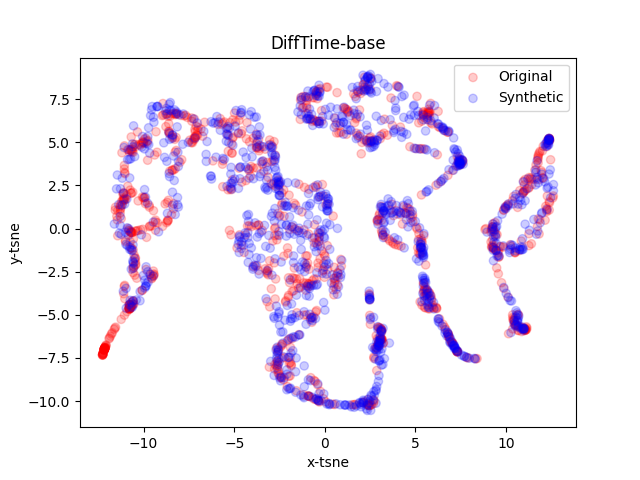}}%
\hfill
\subcaptionbox{\scriptsize  DiffTime-Linear}{\includegraphics[width=0.32\textwidth]{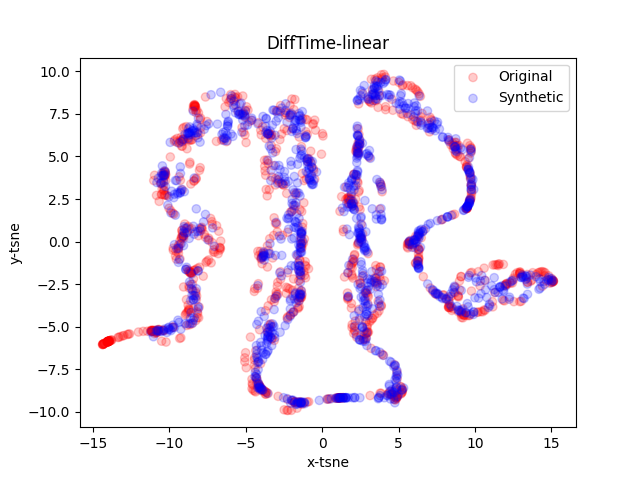}}%
\hfill
\subcaptionbox{\scriptsize DiffTime-Cosine}{\includegraphics[width=0.32\textwidth]{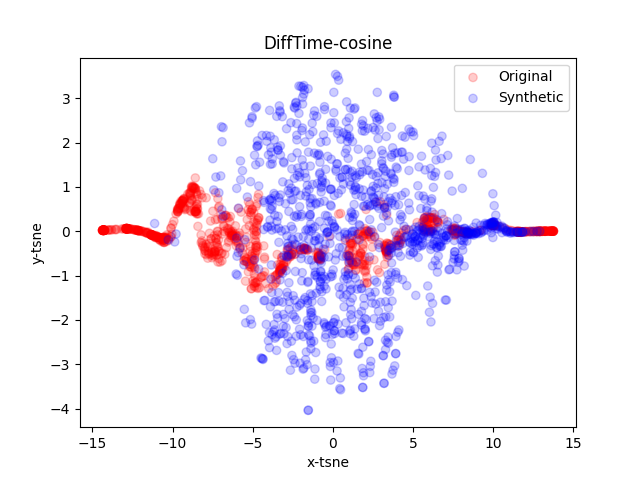}}
\caption{A t-SNE visualizations of \textit{DiffTime} for different noise variance scheduler.}\label{fig:tsen_noise}
\end{figure}

In Figure~\ref{fig:tsen_noise} we show the t-SNE comparison for the different schedulers. The figure shows that in our case \textit{Cosine} scheduler does not achieve a good performance, while both linear and quad scheduler better cover the input data distribution. 

\begin{table}[hbt]
\centering
\caption{ \textit{DiffTime} with different noise variance scheduler.}\label{tab:sched_vary}
\begin{tabular}{llll}
\toprule
Algo & Discr-Score & Pred-Score & Inference-Time \\
\midrule
DiffTime-quad   &   \textbf{0.05±0.03} &  \textbf{0.21±0.00} &     \textbf{0.021±0.00} \\
DiffTime-linear &   0.06±0.02 &  \textbf{0.21±0.00} &     \textbf{0.021±0.01} \\
DiffTime-cosine &   0.25±0.02 &  0.23±0.00 &     \textbf{0.021±0.00} \\
\bottomrule
\end{tabular}
\end{table}

{\color{black}
\subsection{Diffusion model architecture}\label{sec:model_architecture}
We now evaluate the impact of different model layers, and hyper-parameters, on DiffTime performance. We introduce the following variants of DiffTime:
\begin{itemize}
    \item \textit{DiffTime-K-Heads} - we change the number of \textit{attention heads}, from 1 to 8;
    \item \textit{DiffTime-LSTM} - we replace the convolutional layers using recurrent layers (i.e., LSTM) along the attention mechanism, which is particularly successful for imputation and interpolation of TS~\cite{shukla2020multi};
    \item \textit{DiffTime-full-LSTM} - we replace all the convolutional and transformer layers by using LSTM layers, which is common for time-series generation~\cite{mogren2016c};
    \item \textit{DiffTime-full-CNN} - we replace the transformer layers using convolutional layers;
\end{itemize}

\begin{table}[hbt]
\centering
\caption{\textit{DiffTime} using different layers and hyper-parameters}\label{tab:ablation_diff}
\begin{tabular}{l|lll}
\toprule
Algo & Discr-Score & Pred-Score & Inference-Time \\
\midrule
DiffTime-1Heads    &   \textbf{0.03±0.02} &  \textbf{0.21±0.00} &      \textbf{0.02±0.01} \\
DiffTime-4Heads    &   0.06 ± 0.02 &  \textbf{0.21±0.00} &      0.04±0.01 \\
DiffTime-8Heads    &   0.05 ± 0.03 &  \textbf{0.21±0.00} &      \textbf{0.02±0.01} \\
DiffTime-LSTM      &   0.06 ± 0.01 &  \textbf{0.21±0.00} &      \textbf{0.02±0.01} \\
DiffTime-full-LSTM &   0.50 ± 0.00 &  \textbf{0.21±0.00} &      \textbf{0.02±0.01} \\
DiffTime-full-CNN       &   0.14 ± 0.04 &  \textbf{0.21±0.00} &      0.03±0.01 \\
\bottomrule
\end{tabular}
\end{table}

The results are shown in Table~\ref{tab:ablation_diff}. The table highlights the performance of the current architecture, which uses transformer and convolutional layers. Moreover, the table shows that the number of attention heads should be tuned according to the input dataset to achieve better results. 
}

\subsection{COP-method Initial seed}\label{sec:cop_initial_seed}
\begin{figure}[hbt]
\subcaptionbox{\scriptsize  COP-Original}{\includegraphics[width=0.32\textwidth]{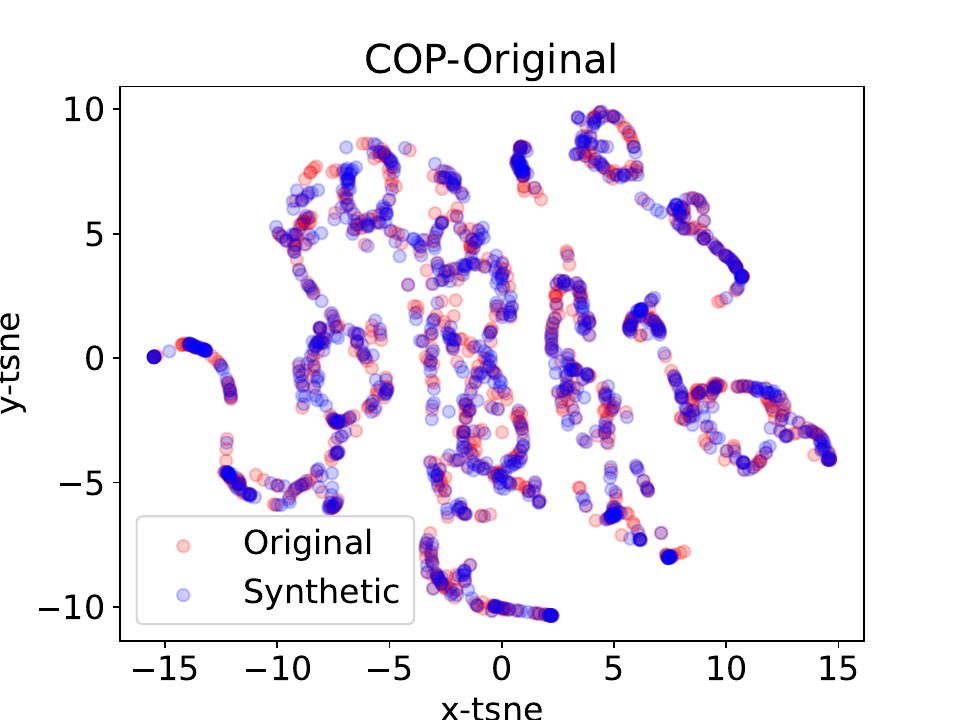}}%
\hfill
\subcaptionbox{\scriptsize  COP-Brownian}{\includegraphics[width=0.32\textwidth]{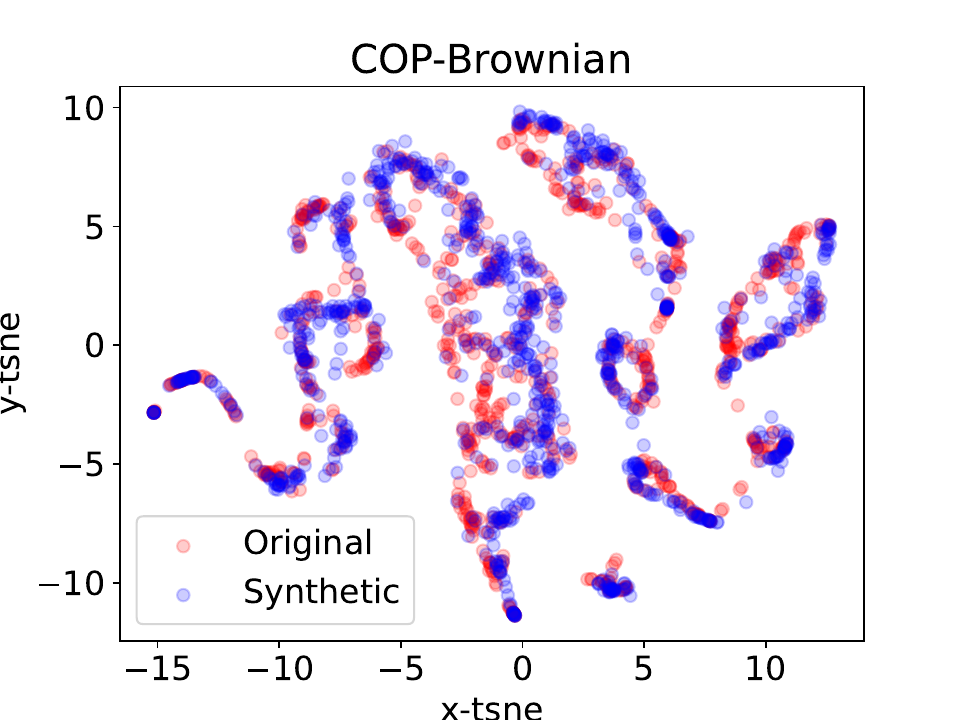}}%
\hfill
\subcaptionbox{\scriptsize  COP-Blended}{\includegraphics[width=0.32\textwidth]{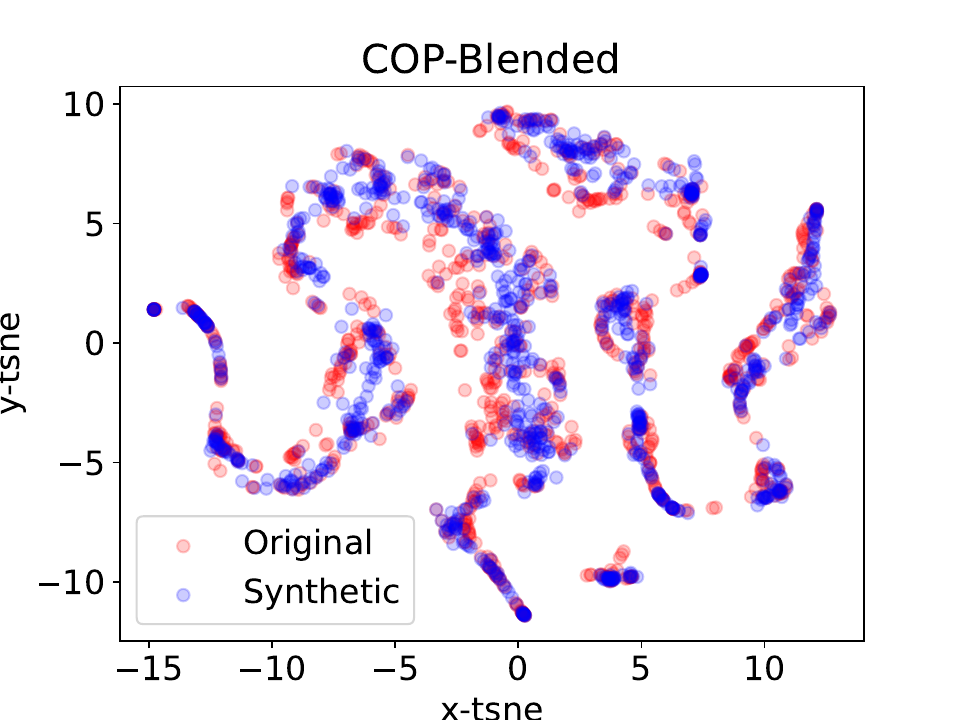}}%
\caption{A t-SNE visualizations \textit{COP-method} using different input seed data.}\label{fig:tsnemin_cop}
\end{figure}

Here we evaluate the impact of different initial seed into COP-method framework. We test the following: a) the input time-series distribution $q(\bx)$; b) Brownian random noise that is scaled to a real TS sample; c) \textit{Blended} time-series where we add brownian noise to the input time-series from $q(\bx)$. Figure~\ref{fig:tsnemin_cop} shows the t-SNE results which show that all the different approaches achieve realistic results, covering the input data distribution. Quantitative metrics are shown in Table~\ref{tab:cop_vary}, and confirm the applicability of COP-method to the different input seed data. 

\begin{table}[H]
\centering
\caption{ \textit{COP-method} with different input seed data.}\label{tab:cop_vary}
\begin{tabular}{llll}
\toprule
Algo & Discr-Score & Pred-Score & Inference-Time \\
\midrule
COP-Blended  &   \textbf{0.01±0.01} &  \textbf{0.20±0.00} &     0.81±0.02 \\
COP-Brownian &   0.02±0.02 &  \textbf{0.20±0.00} &     0.70±0.05 \\
COP-Original &   0.02±0.01 &  \textbf{0.20±0.00} &     0.63±0.01 \\
\bottomrule
\end{tabular}
\end{table}

{\color{black}
\subsection{COP performance using different distance metrics}
COP maximizes a L2 distance as objective, to obtain diversity and create new synthetic samples starting from the input initial seeds. However, L2 distance may not necessarily be the best proxy for diversity, and we can use other distance-based metrics. In this ablation experiment, we compare the performance of COP comparing two different distance metrics. In particular, we empirically evaluated L2 distance and L1 distance. Figure~\ref{fig:cop_l1_vs_l2} shows that both the distance metrics preserve distributional similarity in the synthetic data, which we empirically evaluated using t-SNE. However, the L2 distance achieves slightly better quantitative results, a shown in Table \ref{tab:cop_l1_vs_l2_distance}.

\begin{table}[H]
\centering
\caption{COP using L1 vs L2 distance to generate synthetic samples.}\label{tab:cop_l1_vs_l2_distance}
\begin{tabular}{l|ll}
\toprule
Algo & Discr-Score & Pred-Score \\ 
\midrule
 \text{COP L2-distance} & \textbf{0.017$\pm$0.006} & \textbf{0.203$\pm$0.001} \\ 
 \text{COP L1-distance} & 0.021$\pm$0.012 & \textbf{0.203$\pm$0.002}  \\ 
\bottomrule
\end{tabular}
\end{table}

\begin{figure}[hbt]
\subcaptionbox{\scriptsize  L2-distance}{\includegraphics[width=0.4\textwidth]{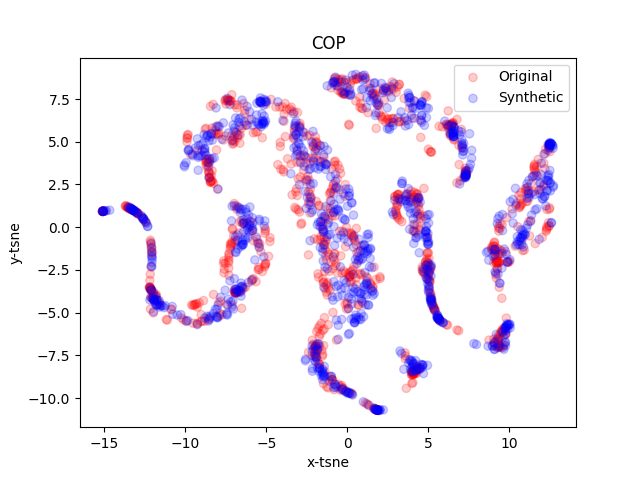}}%
\hfill
\subcaptionbox{\scriptsize  L1-distance}{\includegraphics[width=0.4\textwidth]{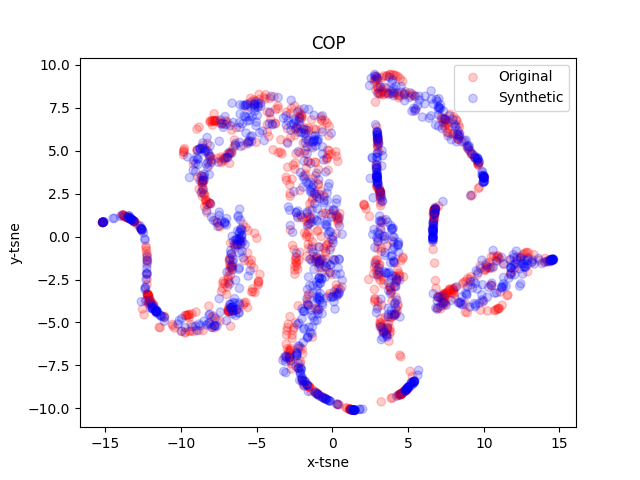}}%
\caption{A t-SNE visualizations of the time-series generated by COP by maximizing the L2 or L1 distance w.r.t. initial seed samples.}\label{fig:cop_l1_vs_l2}
\end{figure}

}

\subsection{The impact of the scale parameter in \textit{Guided-DiffTime}}
\begin{figure}[hbt]
\subcaptionbox{\scriptsize  Guided-DiffTime $\rho=0.4$}{\includegraphics[width=0.32\textwidth]{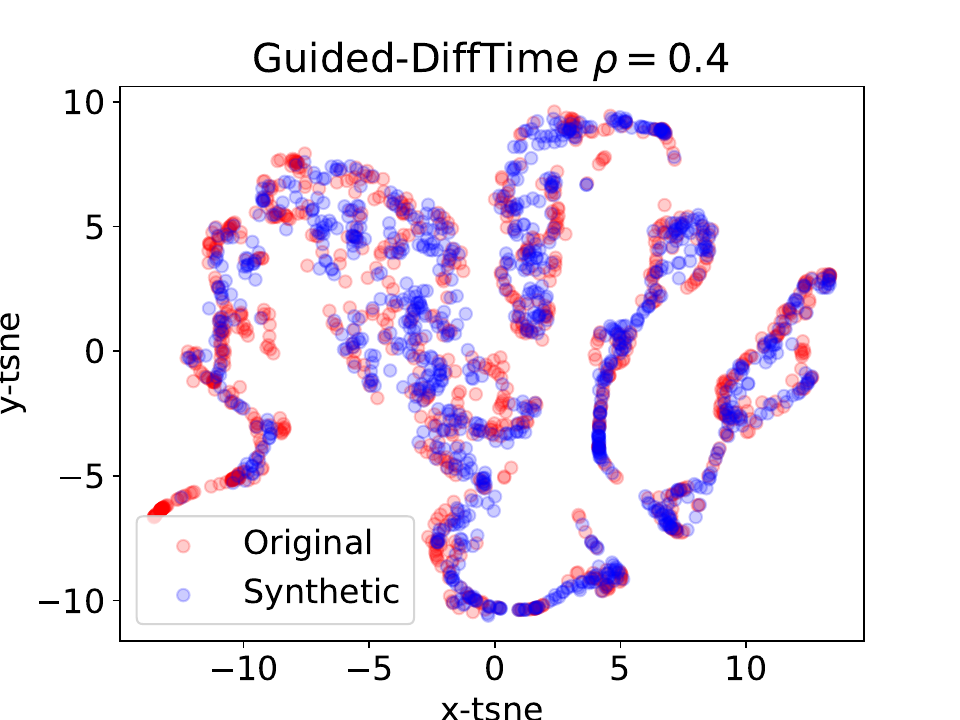}}%
\hfill
\subcaptionbox{\scriptsize  Guided-DiffTime $\rho=0.8$}{\includegraphics[width=0.32\textwidth]{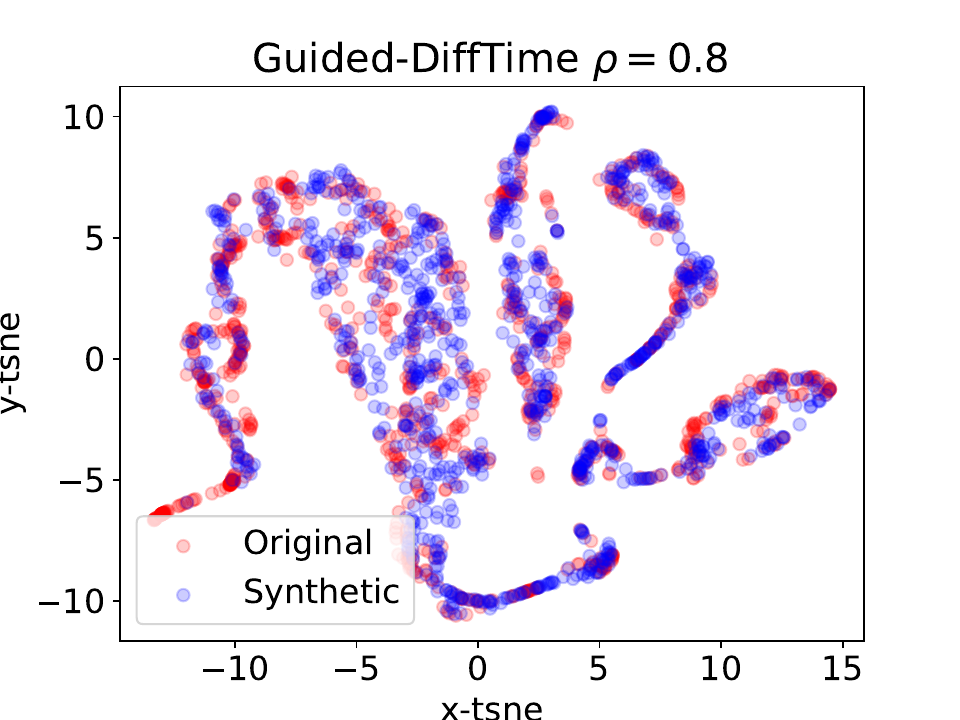}}%
\hfill
\subcaptionbox{\scriptsize  Guided-DiffTime $\rho=1.0$}{\includegraphics[width=0.32\textwidth]{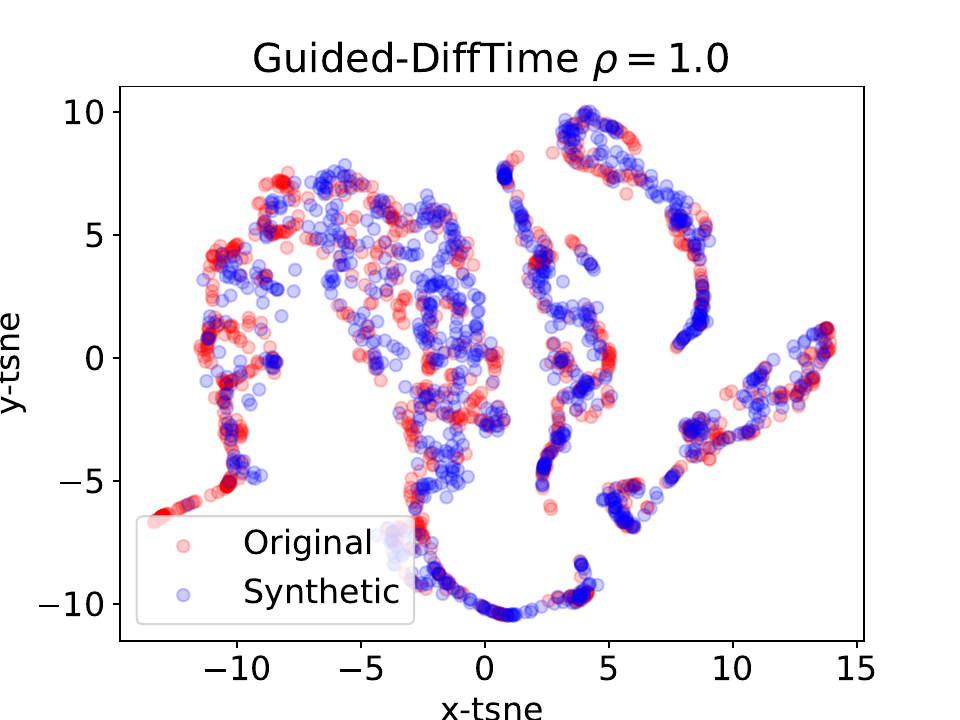}}%
\hfill
\subcaptionbox{\scriptsize  Guided-DiffTime $\rho=1.2$}{\includegraphics[width=0.32\textwidth]{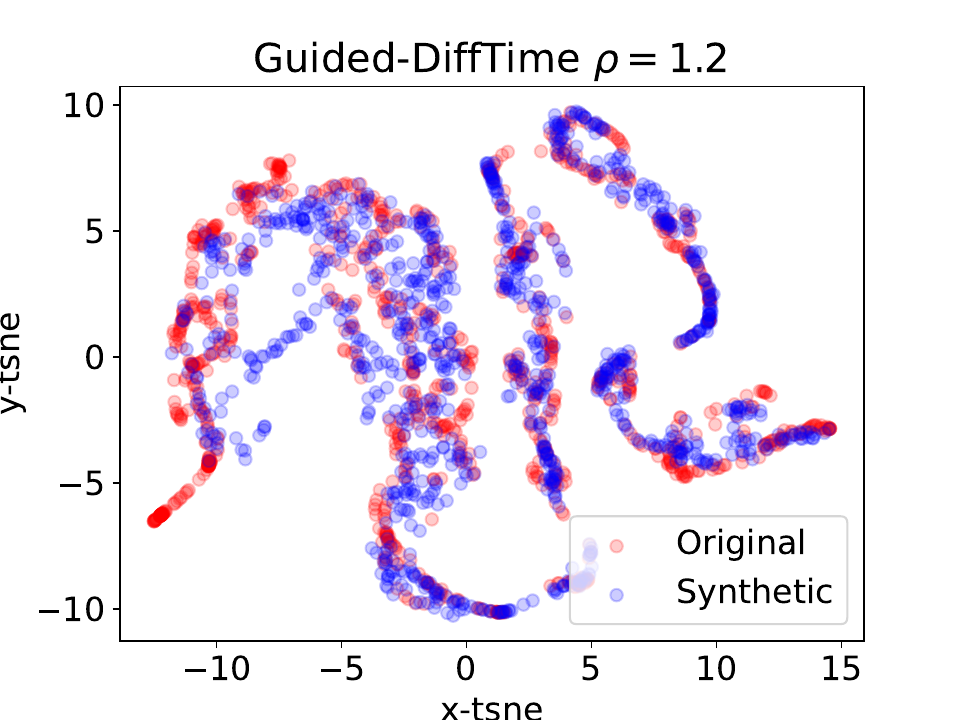}}%
\hfill
\subcaptionbox{\scriptsize  Guided-DiffTime $\rho=1.4$}{\includegraphics[width=0.32\textwidth]{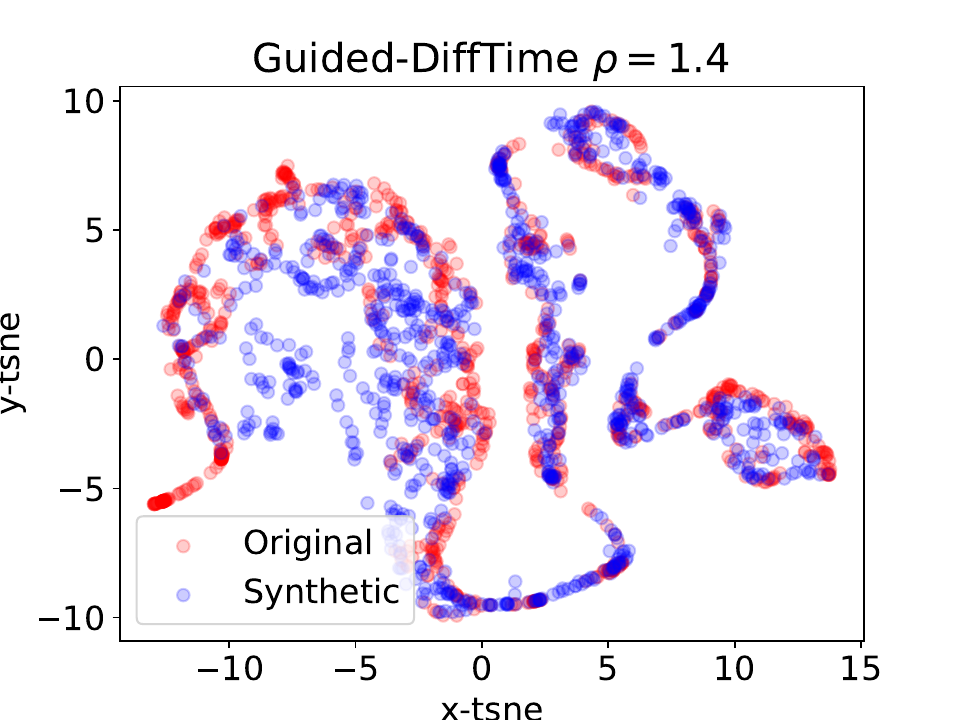}}%
\hfill
\subcaptionbox{\scriptsize  Guided-DiffTime $\rho=2.0$}{\includegraphics[width=0.32\textwidth]{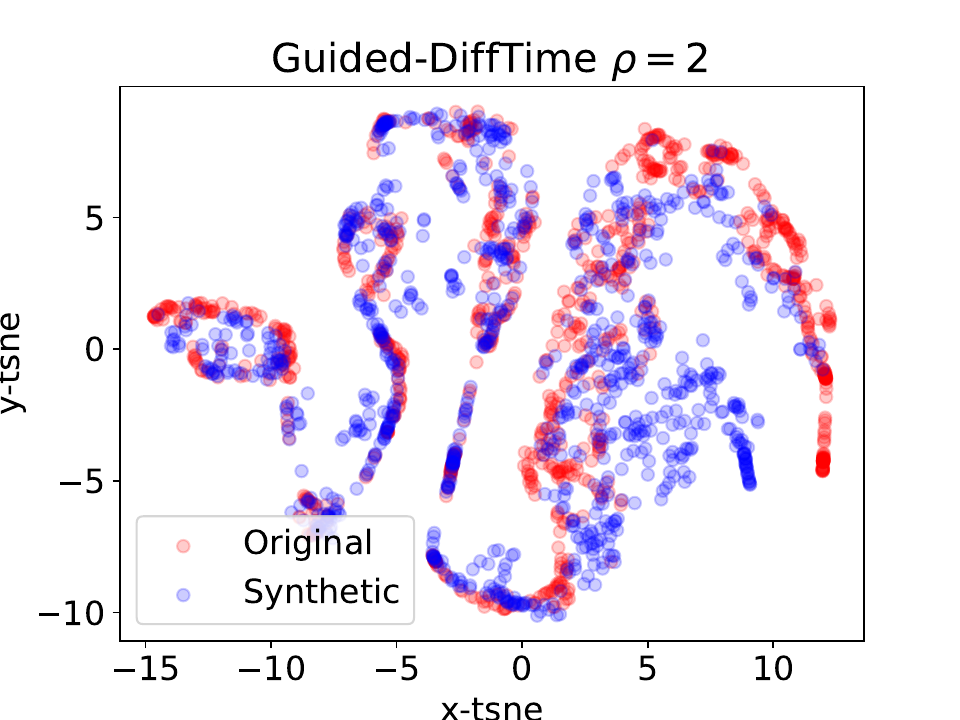}}
\caption{A t-SNE visualizations of \textit{Global Min} constrained data at varying of the scale parameter $\rho$ for \textit{Guided-DiffTime}.}\label{fig:tsnemin_weights}
\end{figure}

\begin{figure}[hbt]
    \centering
    \includegraphics[width=1\textwidth]{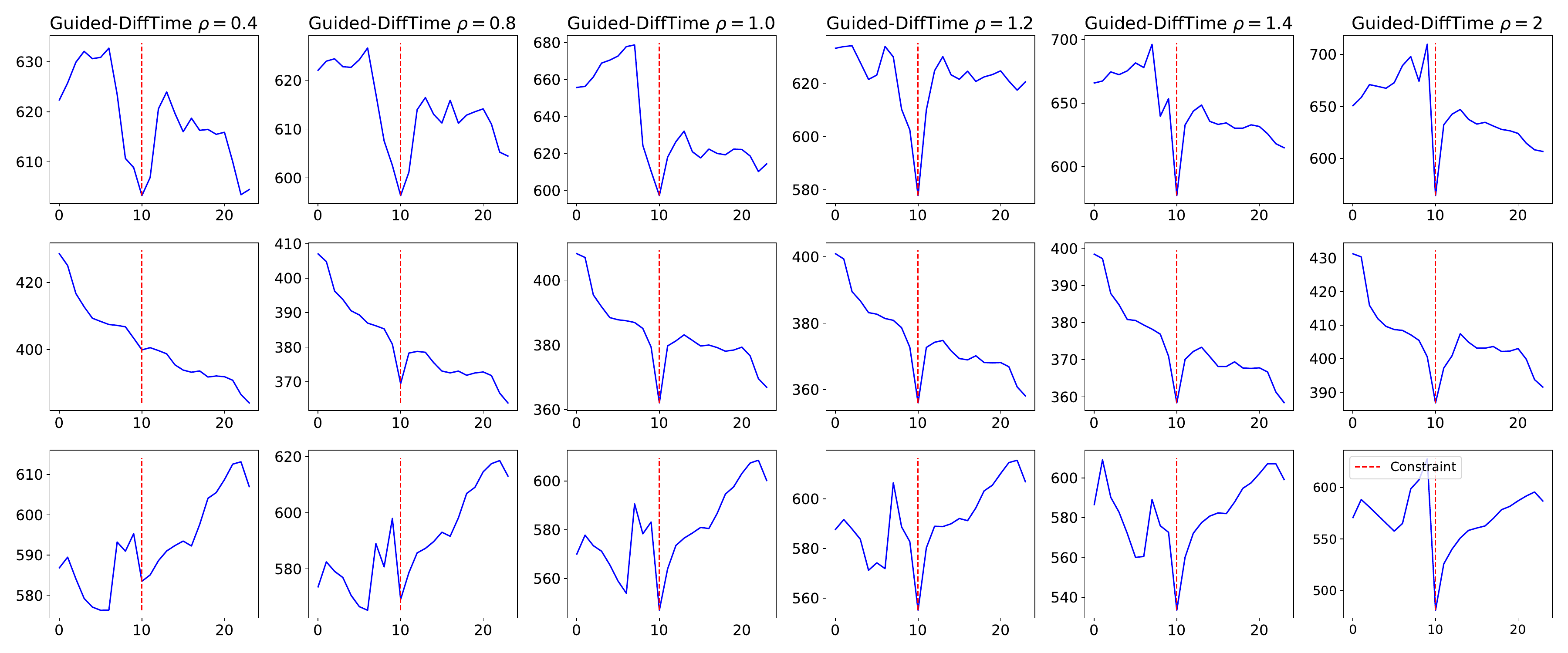}
    \caption{Example of \textit{Global Min} constrained time-series at varying of the scale parameter $\rho$ for \textit{Guided-DiffTime}.}
    \label{fig:vary_example_ts}
\end{figure}

We evaluate the impact of the scale parameter $\rho$ to the \textit{Guided-DiffTime} when applied to \textit{Global Min} constraint.
In Figure~\ref{fig:tsnemin_weights} we report the t-SNE analysis, while in Figure~\ref{fig:vary_example_ts} we show some examples of generated synthetic time-series. The quantitative metrics are reported in Table~\ref{tab:weights_vary}. It's worth noticing that (as expected) increasing of the scale parameter $\rho$, results in the model trading-off realism to guarantee the constraints for all the synthetic time-series. 

\begin{table}[H]
\centering
\caption{ \textit{Guided-DiffTime} global-min constrained generation at varying of $\rho$.}\label{tab:weights_vary}
\vspace{0.2in}
\begin{tabular}{lllll}
\toprule
Algo & Discr-Score & Pred-Score & Inference-Time & Satisfaction Rate \\
GuidedDiffTime              &             &            &                &                   \\
\midrule
$\rho=$0.4 &   \textbf{0.04±0.03} &  \textbf{0.21±0.00} &     0.034±0.02 &         0.36±0.00 \\
$\rho=$0.8 &   \textbf{0.04±0.03} &  \textbf{0.21±0.00} &     0.033±0.02 &         0.70±0.00 \\
$\rho=$1.0 &   0.05±0.02 &  \textbf{0.21±0.00} &     \textbf{0.032±0.01} &         0.81±0.00 \\
$\rho=$1.2 &   0.06±0.02 &  \textbf{0.21±0.00} &     \textbf{0.032±0.01} &         0.88±0.00 \\
$\rho=$1.4 &   0.06±0.03 &  \textbf{0.21±0.00} &     0.033±0.04 &         0.90±0.00 \\
$\rho=$2.0 &   0.07±0.02 &  \textbf{0.21±0.00} &     0.034±0.02 &         \textbf{0.94±0.00} \\
\bottomrule
\end{tabular}
\end{table}

\section{Additional Experiments}\label{sec:add_exp}
In this section we present additional experiments which we omitted in the main body of the paper due to limited space. Where not otherwise stated, we consider univariate stock-data. 

\subsection{The computational cost of constrained-generation}\label{sec:carbon_footprint}
First we evaluate the impact of adding a new constraint on the proposed models. We evaluate this in terms of computational cost, i.e., the computational resources and time needed to incorporate the new constraints and sample $N=1000$ time-series for each constraint. For this experiment, we compute the \textit{Global Min} constraint and vary the global minimum index $i \in [0, 23]$, i.e., along all the time-series. Therefore, we have 24 different constraints.

In Table~\ref{tab:comp_cost}\footnote{The presented values are estimated using available experimental data, to reduce the computational cost.} we show the training, inference and total computation times required for all the 24 constraints. The table shows that \textit{COP-method} does not require any training, however has a large sampling (inference) time, due to the complexity of the optimization problem. Instead, \textit{Guided-DiffTime} only requires that we train a single unconstrained \textit{DiffTime} model used to handle all the different constraints. Therefore, \textit{Guided-DiffTime} has a very low computational cost with respect to other approaches that have to be re-trained for each new constraint. The table shows that \textit{Guided-DiffTime} is estimated to reduce the emission of around $60\%$ w.r.t. to \textit{COP-method} and around $92\%$ w.r.t. other Deep Generative models. All the deep generative models are trained on a NVIDIA T4 GPU, with 4 cores and 16gb or RAM. To compare the computational times, the inference is done on a 4 core 3rd generation AMD EPYC processors for all the models including COP. Experiments were conducted using \href{https://docs.aws.amazon.com/ec2/index.html}{AWS cloud service} in Ohio region, where the total emissions are estimated using a \href{https://mlco2.github.io/impact#compute}{Machine Learning Impact calculator} presented in \cite{lacoste2019quantifying}. 

\begin{table}[H]
\caption{Constrained Generation - Estimated total computational cost}\label{tab:comp_cost}
\vspace{0.2in}
\centering
\resizebox{1\linewidth}{!}{
\begin{tabular}{lrrrr}
\toprule
      Algorithm &  Training-Time (hrs) &  Inference-Time (hrs) &  Total-Time (hrs) &  Emissions (kgCO$_2$eq) \\
\midrule
              COP-method &                  0.0 &            127.8 &                      127.8 &                1.25 \\
Guided-DiffTime &                 12.0 &              0.2 &                       12.2 &                0.52 \\
    Loss-DiffTime &                312.0 &              0.1 &                      312.1 &               12.45 \\
          TimeGAN &                400.0 &              0.0 &                      400.0 &               15.96 \\
            RCGAN &                156.0 &              0.0 &                      156.0 &                6.22 \\
           GT-GAN &                192.0 &              0.0 &                      192.0 &                7.66 \\
\bottomrule
\end{tabular}
}
\end{table}

\newpage
\color{black}
\subsection{Longer time-series using DiffTime}\label{sec:longer-ts}
We now evaluate the impact of the different time-series lengths on the generative models for un-constrained generation. Notice that, while this is not the goal of our work, DiffTime and COP shows consistently higher performance for longer time-series, while maintaining a stable training/inference procedure. On the other hand, GANs-based methods, which have inherently unstable training, show decreased performance for longer time-series. We consider daily stock-data with three different lengths $\in [36, 72, 360]$ (i.e., days). For these experiments we keep all the same hyper-parameters and we only change: 
\begin{itemize}
    \item the kernel-size $ks$ of CNN layers in the diffusion model, being $ks \in [3,6,24]$ for the different lengths $\in [36,72,360]$, respectively; 
\item the hidden-dimension of \textit{TimeGAN}, \textit{RCGAN}, and \textit{GT-GAN}, which is set to be the time-series length, as suggested by authors and empirically evaluated; 
\item the window size $\theta_w$ of COP, being the time-series length divided by 2. 
\end{itemize}

As mentioned, we found that the training time highly increase for \textit{TimeGAN} and \textit{GT-GAN}, especially with time series of length equal to 360. For \textit{RCGAN} and \textit{DiffTime} the training time is only slightly increased. 

\begin{table}[h]
\caption{Un-Constrained Generation - Longer TS}\label{tab:longer_Ts}
\centering
\begin{tabular}{lllll}
\toprule
Length TS & Algo & Discr-Score & Pred-Score & Inference-Time \\
\midrule
\textbf{36}  & COP-Brownian &  \textbf{0.01±0.01} &  \textbf{0.20±0.00} &      0.33±0.00 \\
    & COP-Original &  \textbf{0.01±0.01} &  \textbf{0.20±0.00} &      0.04±0.00 \\
    & DiffTime &  0.04±0.03 &  0.21±0.00 &      0.05±0.00 \\
    & GT-GAN &  0.03±0.02 &  0.21±0.00 &      0.00±0.00 \\
    & RCGAN &   \textbf{0.01±0.01} &  \textbf{0.20±0.00} &      0.00±0.00 \\
    & TimeGAN &   0.03±0.02 &  \textbf{0.20±0.00} &      0.00±0.00 \\ \hline
\textbf{72}  & COP-Brownian &   \textbf{0.02±0.01} &  \textbf{0.21±0.00} &      0.63±0.00 \\
    & COP-Original &   \textbf{0.02±0.01} &  \textbf{0.21±0.00} &      0.07±0.00 \\
    & DiffTime &   0.04±0.02 &  0.22±0.00 &      0.15±0.00 \\
    & GT-GAN &   0.09±0.05 &  0.22±0.00 &      0.00±0.00 \\
    & RCGAN &   0.03±0.02 &  \textbf{0.21±0.00} &      0.00±0.00 \\
    & TimeGAN &   0.06±0.02 &  0.24±0.00 &      0.00±0.00 \\ \hline
\textbf{360} & COP-Brownian &   0.06±0.04 &  \textbf{0.20±0.00} &      2.39±0.00 \\
    & COP-Original &   \textbf{0.03±0.01} &  \textbf{0.20±0.00} &      0.38±0.00 \\
    & DiffTime &   0.06±0.06 &  \textbf{0.20±0.00} &      0.04±0.00 \\
    & GT-GAN &   0.18±0.05 &  \textbf{0.20±0.00} &      0.03±0.00 \\
    & RCGAN &   0.09±0.06 &  0.21±0.00 &      0.00±0.00 \\
    & TimeGAN &   0.10±0.09 &  0.22±0.00 &      0.00±0.00 \\
\bottomrule
\end{tabular}
\end{table}

The quantitative metrics are reported in Table~\ref{tab:longer_Ts}. It's worth noticing that (as expected) increasing the length of the time-series results in lower performance, as the models have more difficulty to capture the longer statistical properties of the time-series. However, \textit{DiffTime} and \textit{COP} have the lower degradation: the Discr. Score of \textit{DiffTime} and \textit{COP} increases only of $0.02$ when the time-series length increases from 36 to 360, while the other methods have at least $0.08$ ($400\%$ more) increase in Discr. Score.

\begin{figure}[hbt]
\subcaptionbox{\scriptsize COP-Original}{\includegraphics[width=0.32\textwidth]{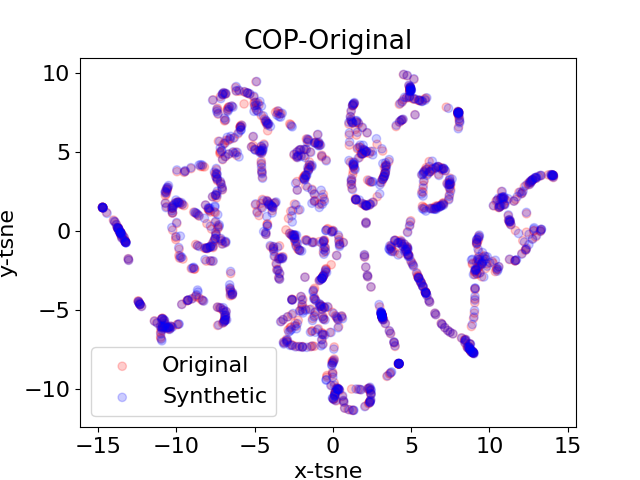}}%
\hfill
\subcaptionbox{\scriptsize COP-Brownian}{\includegraphics[width=0.32\textwidth]{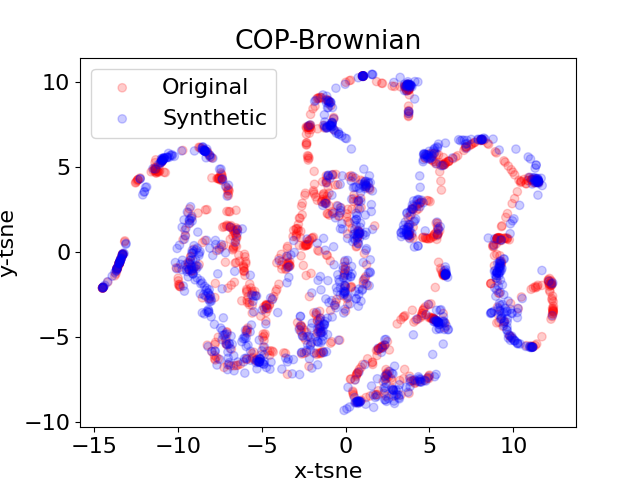}}%
\hfill
\subcaptionbox{\scriptsize DiffTime}{\includegraphics[width=0.32\textwidth]{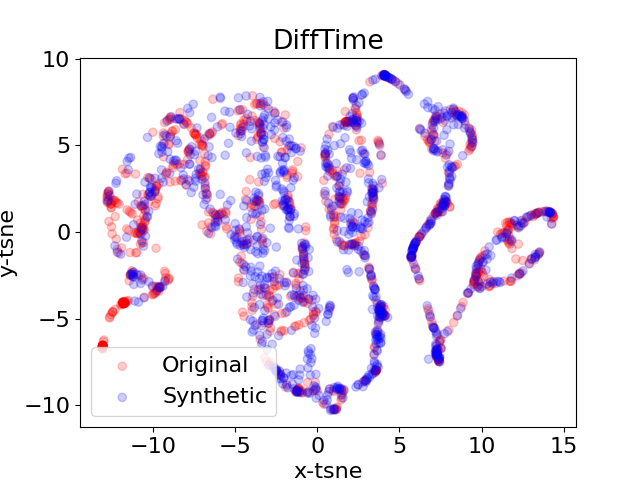}}%
\hfill
\subcaptionbox{\scriptsize GT-GAN}{\includegraphics[width=0.32\textwidth]{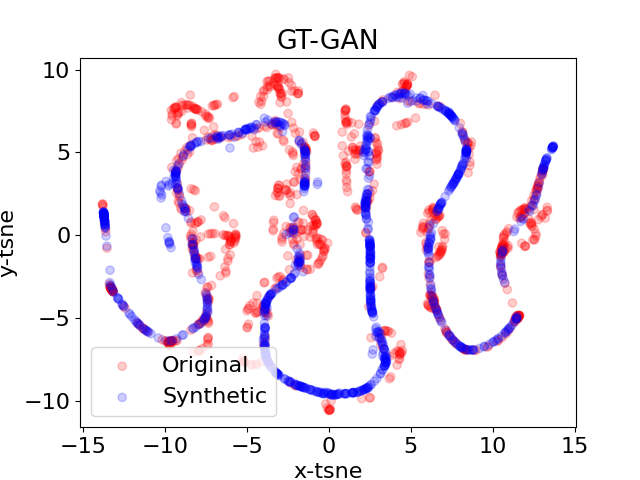}}%
\hfill
\subcaptionbox{\scriptsize TimeGAN}{\includegraphics[width=0.32\textwidth]{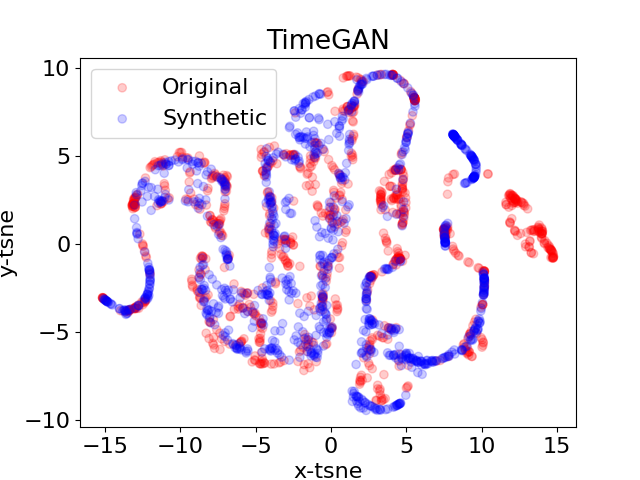}}%
\hfill
\subcaptionbox{\scriptsize RCGAN}{\includegraphics[width=0.32\textwidth]{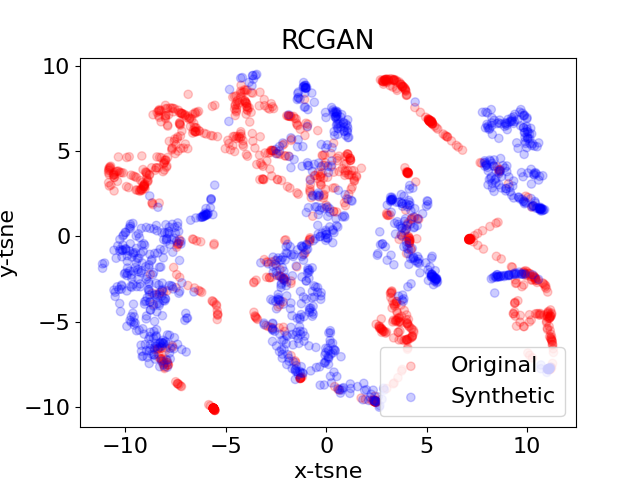}}
\caption{A t-SNE visualization of un-constrained time-series with length equal to 36. A greater overlap of blue and red dots implies a better distributional-similarity between the generated data and original data. Our approaches shows the best performance.}\label{fig:longer_ts_tsne}
\end{figure}

\begin{figure}[hbt]
    \centering
    \vspace{-0.2in}
    \includegraphics[width=1\textwidth]{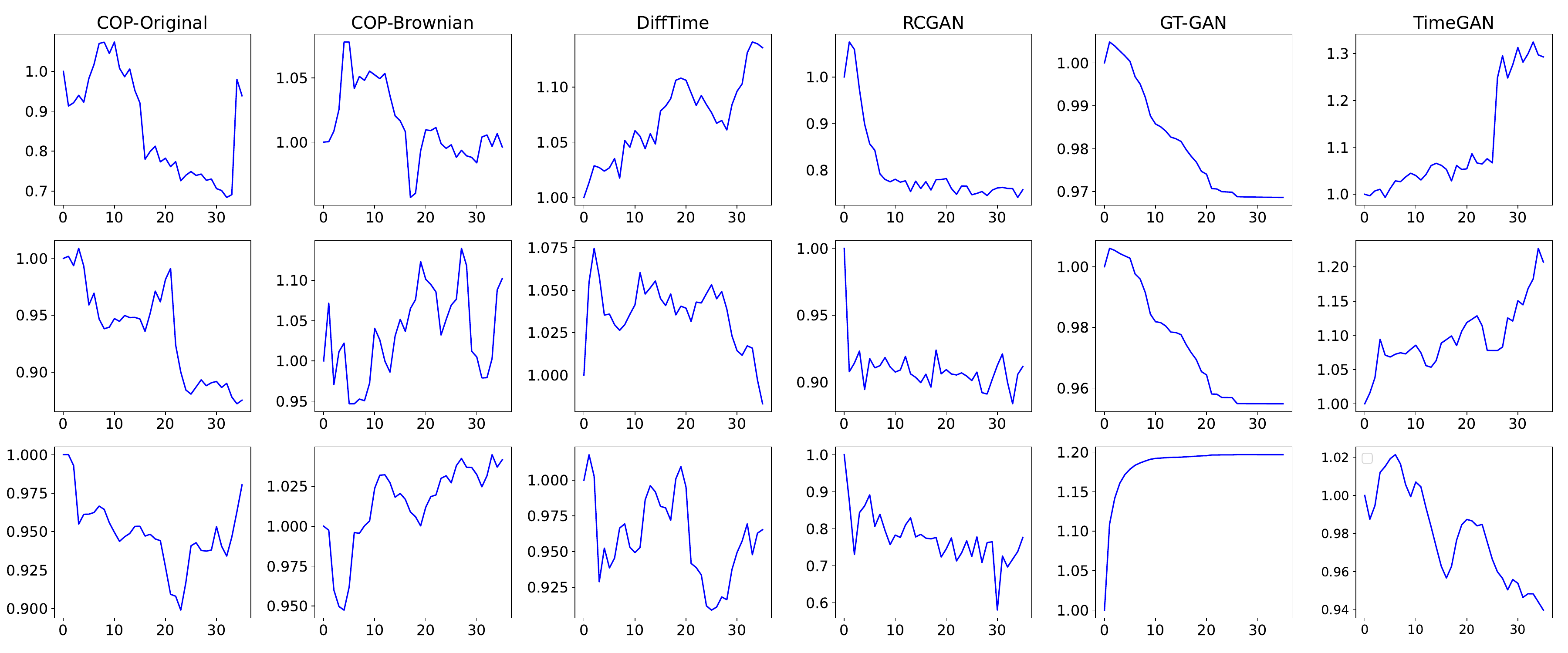}
    \caption{Example of un constrained time-series with length 36.}
    \label{fig:vary_example_ts_longer}
    \vspace{-0.2in}
\end{figure}

In Figure~\ref{fig:longer_ts_tsne} we report the t-SNE analysis for length 36, while in Figure~\ref{fig:vary_example_ts_longer} we show some examples of generated synthetic time-series, normalized w.r.t. their first values.  While the generated time-series in Figure~\ref{fig:vary_example_ts_longer} may seem reasonable, some of them exhibit very unusual volatility (e.g., RCGAN and TimeGAN generate time-series with more than $30\%$ price changes in 36 days), while others samples have not much diversity (i.e., first two time-series generated by GT-GAN). Importantly, it is also the case that professional traders can easily distinguish between real stock price series and synthetic price series generated by simple price models~\cite{mandelbrot2010mis}.

In the next section we better investigate some specific financial properties, called stylized facts~\cite{vyetrenko2019get}, to show that  our approaches outperform the benchmarks in preserving real data properties. 

\subsection{Financial properties}\label{sec:fin-properties}
In this section we investigate three specific financial properties of price series, showing that synthetic time-series generated by our approaches better preserve such properties w.r.t. existing benchmarks. For example, as asset daily returns usually have fat tail distribution and long-range dependence, we expected the same properties (or \textit{stylized facts}) from artificial markets. To have a fair comparison, we choose the case of time-series with length equal to 36, as 
the existing benchmarks have the closest performance to us when the length is 36 (seeTable~\ref{tab:longer_Ts}). 

We evaluate the following three stylized facts \textit{auto-correlations}, \textit{heavy tails distribution}, and \textit{long range dependence}, to evaluate asset return properties. We refer the reader to the work in \cite{vyetrenko2019get} and \cite{bouchaud2018trades} for a more detailed introduction to stylized facts.

\begin{figure}[hbt]
    \centering
    \includegraphics[width=1\textwidth]{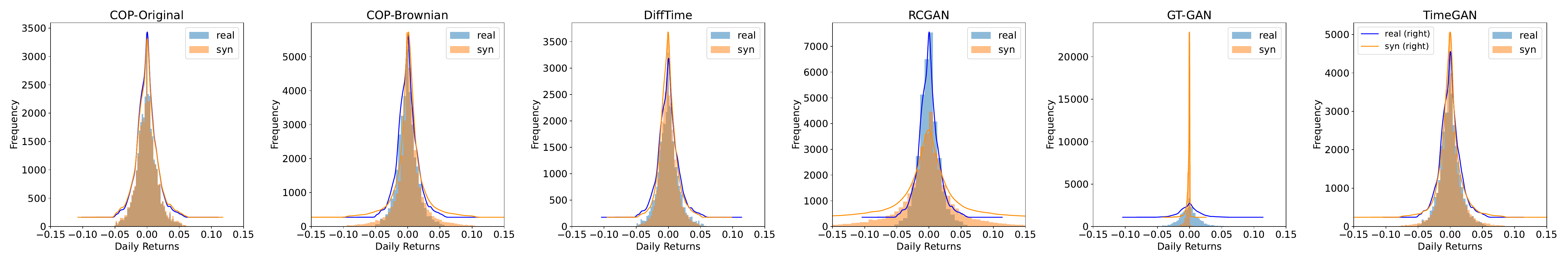}
    \caption{Returns distribution of un-constrained time-series with length 36.}
    \label{fig:returns_ts_longer}
\end{figure}

The first Figure~\ref{fig:returns_ts_longer} shows the return distribution of real and synthetic time-series, for all the approaches. Our approaches show better overlap between orange and blue distributions, as the synthetic time-series better resemble the real data returns. Is it interesting to note that RCGAN synthetic data has a too much fat-tailed distribution, although in table~\ref{tab:longer_Ts} it has among the best performance in Discr. Score.

\begin{figure}[hbt]
    \centering
    \includegraphics[width=1\textwidth]{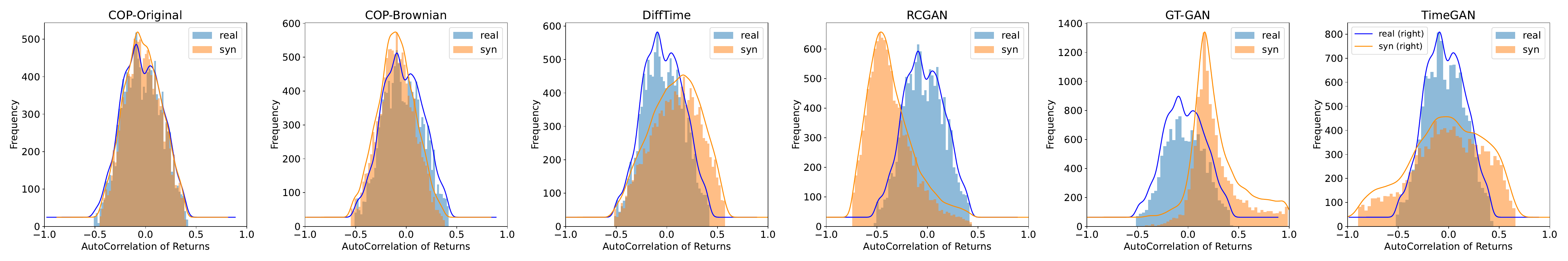}
    \caption{Auto-correlation distribution of un-constrained time-series with length 36.}
    \label{fig:autocorr_ts_longer}
\end{figure}

\begin{figure}[h!]
    \centering
    \includegraphics[width=1\textwidth]{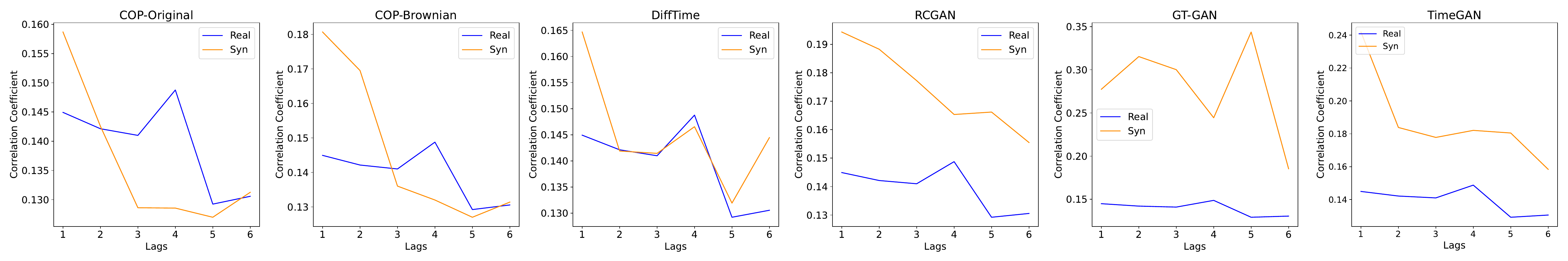}
    \caption{Volatility-clustering and long-range dependence for time-series with length 36.}
    \label{fig:long_range_ts_longer}
\end{figure}

Figure~\ref{fig:autocorr_ts_longer} confirms the superiority of our approaches as the auto-correlation of synthetic returns have much more similarity to those of real data: our approaches show better overlap between orange and blue distributions. Finally, in Figure~\ref{fig:long_range_ts_longer} we show the long-range correlation/dependence of returns, with different lags from 1 to 6 days. The charts show that the volatility decays at increasing number of the days, and that DiffTime has the best performance in preserving this property: orange and blue lines are closer.   

\subsection{Time-Series fine-tuning using COP}\label{sec:fine-tuning-COP}
In this section we show that COP can be used to fine-tune synthetic samples and enforce constraints, for any deep learning model. In particular, we recall that COP can take as input synthetic samples that do not respect a given constraint, and it can slightly alter them (see Algorithm~\ref{alg:COP_repeated_optimization_search} and Figure~\ref{fig:COP_sliding_window}) to meet the required properties and comply with the input constraint. We consider again the multivariate constraint of Section~\ref{sec:ohlc}, using the multivariate Google stock data.  Notice that, COP fine-tuning procedure minimizes the L2 distance between the input samples and the generated ones, i.e., it minimizes the number of changes needed to satisfy the constraints. 

\begin{figure}[hbt]
\subcaptionbox{\scriptsize Guided-DiffTime}{\includegraphics[width=0.24\textwidth]{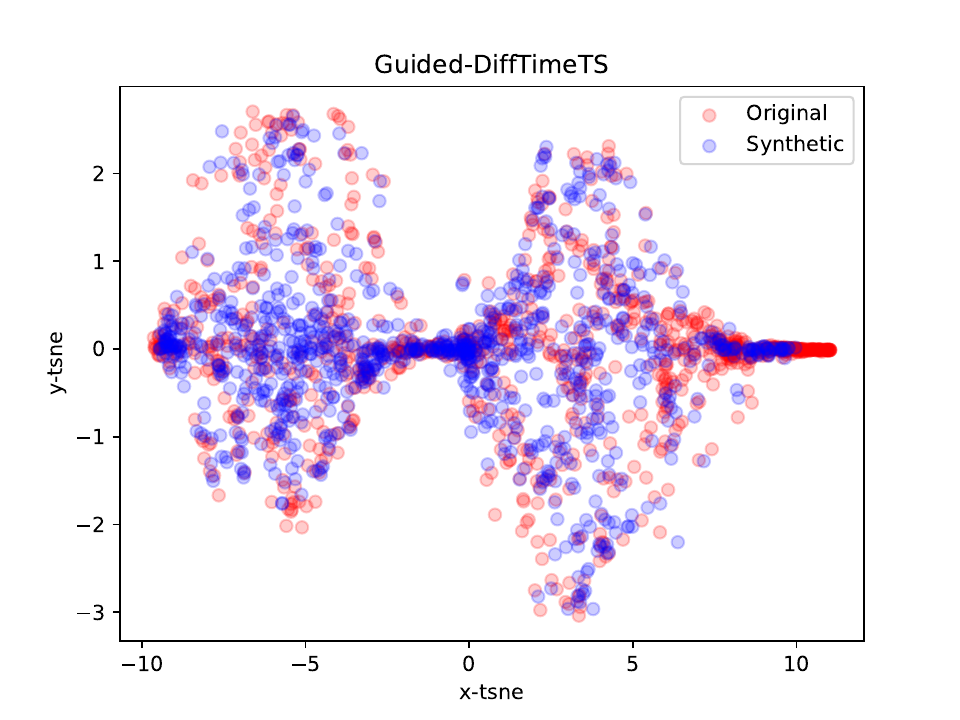}}%
\hfill
\subcaptionbox{\scriptsize COP fine-tuned Guided-DiffTime}{\includegraphics[width=0.24\textwidth]{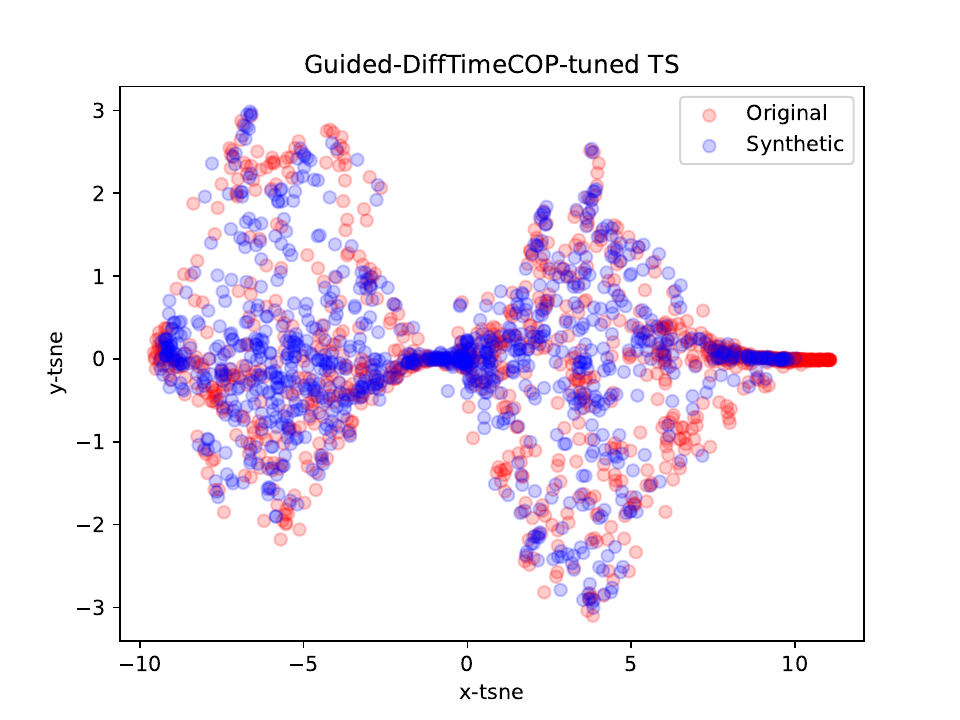}}%
\hfill
\subcaptionbox{\scriptsize TimeGAN}{\includegraphics[width=0.24\textwidth]{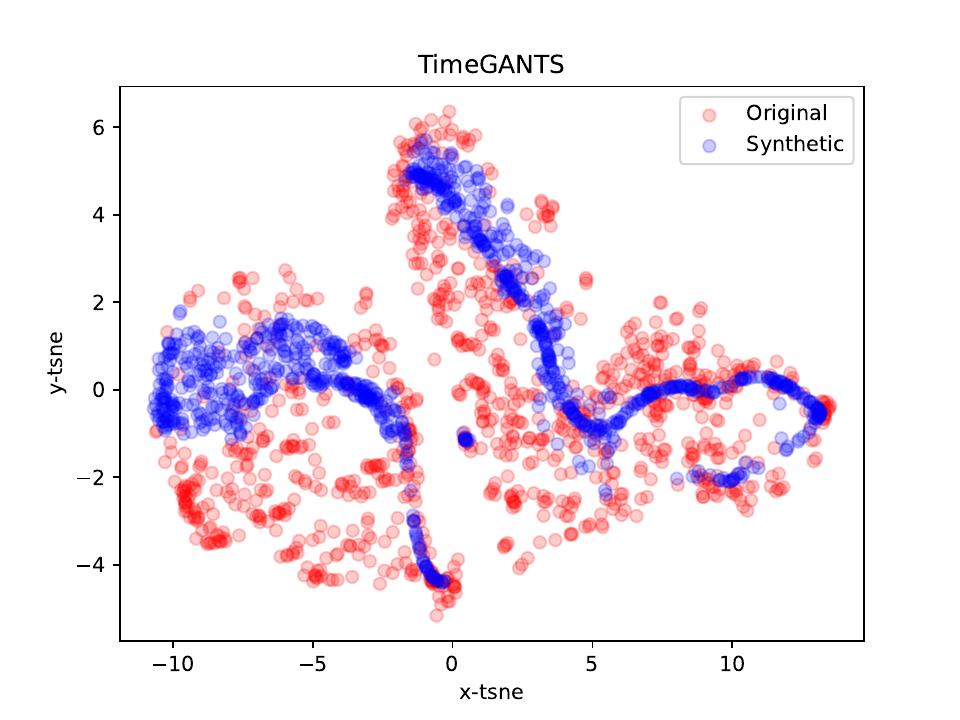}}%
\hfill
\subcaptionbox{\scriptsize COP fine-tuned TimeGAN}{\includegraphics[width=0.24\textwidth]{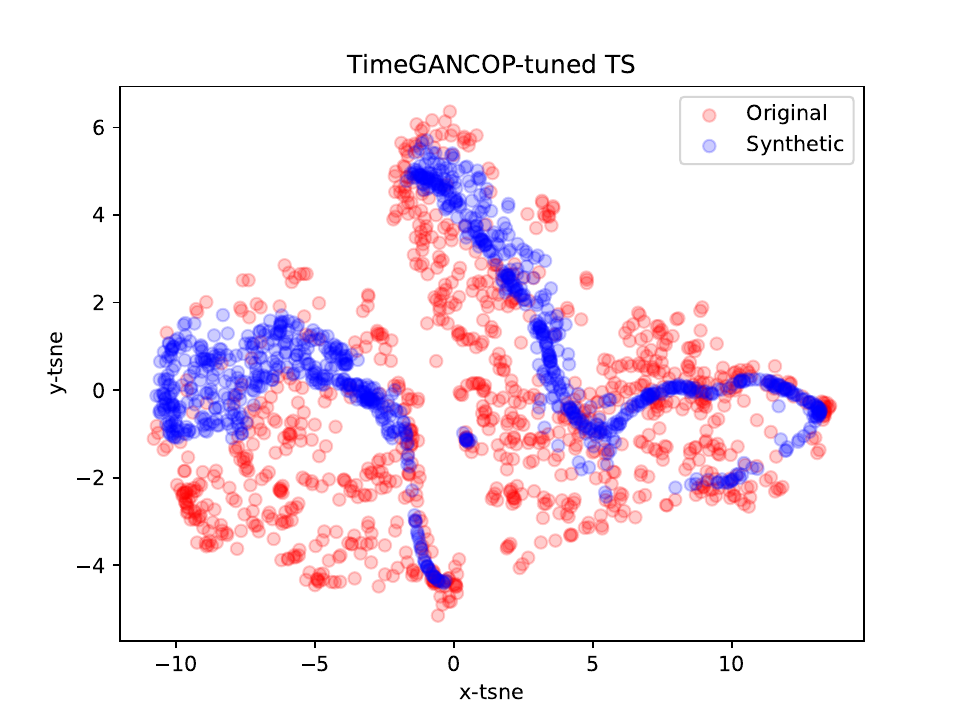}}%

\caption{A t-SNE visualizations of time-series generated by Guided-DiffTime and TimeGAN for OHLC, before and after applying COP for fine-tuning. The fine-tuning does not alter or deteriorate the data distribution.}\label{fig:Cop_fine_tune}
\end{figure}

In Table~\ref{tab:cop_finetuned} we show that COP can fine-tune generated samples and highly improve the percentage of TS that respect the input OHLC constraint. Notice that, COP does not guarantee 100\% of satisfaction rate, as for some samples it is not able to guarantee the constraints (under current settings) without destroying original data properties (e.g., autocorrelation), thus it fails. However, COP almost doubles the satisfaction rate, and with different settings it can guarantee even higher satisfaction rate. In particular, while Guided-DiffTime and TimeGAN achieve a satisfaction rate of 72\% and 51\%, respectively, after the fine-tuning they achieve 97.3\% and 89.7\%.  Importantly, Figure~\ref{fig:Cop_fine_tune} confirms that the data distribution learn by the model is not highly affected by COP fine-tuning procedure.

\begin{table}[h]
\caption{OHLC-Constrained Generation}\label{tab:cop_finetuned}
\centering
\begin{tabular}{lllll}
\toprule
Algo & COP Fine-Tuning & Satisfaction-Rate  \\
\midrule
Guided-DiffTime &  False &  0.72±0.02  \\
TimeGAN &  False &  0.51±0.02 \\
\hline
Guided-DiffTime &  True  &  \textbf{0.97±0.03} \\
TimeGAN &   True  &  \textbf{0.90±0.02} \\
\bottomrule
\end{tabular}
\end{table}

\newpage
\section{Extended related work comparison}
In recent years, there has been a growing body of research dedicated to the exploration of synthetic data, with particular emphasis on its application within the financial and healthcare domain~\cite{van2023synthetic,timeGAN,jeon2022gt,esteban2017real,mogren2016c,coletta2022learning,chen2021synthetic,coletta2021towards}. This surge in interest can be attributed to the escalating utility demonstrated by synthetic data across a diverse array of studies, particularly in scenarios where access to genuine data is restricted due to privacy constraints~\cite{alaa2022faithful,van2023beyond,coletta2023k,esteban2017real}. 

In this section we survey additional related work for synthetic time-series generation. In particular, we consider the following state-of-art approaches: COSCI-GAN~\cite{seyfi2022generating}, RTSGAN~\cite{pei2021towards}, and LS4~\cite{zhou2023deep}. COSCI-GAN is a promising GAN-based approach that focuses mostly on synthetic multivariate time series, which originates from a single source (i.e., biometric measurements from a medical patient; or open-high-low-close time-series from financial markets). We consider such work as it shows promising results, especially for the preservation of inter-channel/feature dynamics: we may expect such work to easily capture the OHLC constraint from data itself. The second work, namely RTSGAN, focuses on real-world time series, where sequences can have variable lengths, missing data, and noisy observations. The work proposes a novel generative framework where an encoder-decoder module learns a mapping between a time series instance and a fixed dimension latent vector, and the generative model works on such lower dimensional latent space. 
To the best of our knowledge, this work shows state-of-art results on multivariate stock data. Finally, LS4 is a generative model that uses latent variables evolving according to a state space ODE to increase modeling capacity. However, differently from us, it focuses on long-sequence modelling and continuous time-series. Therefore, we do not consider this last work as benchmark in our extended comparison. 

Furthermore, we recall that none of the above mentioned models directly support constrained generation. Thus, we first consider them within the domain of unconstrained time-series (TS) generation. Then, we modified the training procedure of such models by introducing a penalty loss, which penalizes the generative models proportional to how much the generated time-series violate the input constraint. For the constrained generation we specifically focus on the Open-High-Low-Close (OHLC) constraint. We chose OHLC constraint for comparing the new baselines since COSCI-GAN is intended for multivariate time series and OHLC is a constraint on the relative values between 4 time series. For both COSCI-GAN and RTSGAN we follow the official authors' implementation.

\subsection{Un-Constrained Generation}
\begin{figure}[hbt]
\subcaptionbox{\scriptsize COP}{\includegraphics[width=0.24\textwidth]{images/unconditional/cop_stock.png}}%
\hfill
\subcaptionbox{\scriptsize DiffTime}{\includegraphics[width=0.24\textwidth]{images/unconditional/diff_tsne.png}}%
\hfill
\subcaptionbox{\scriptsize RTSGAN}{\includegraphics[width=0.24\textwidth]{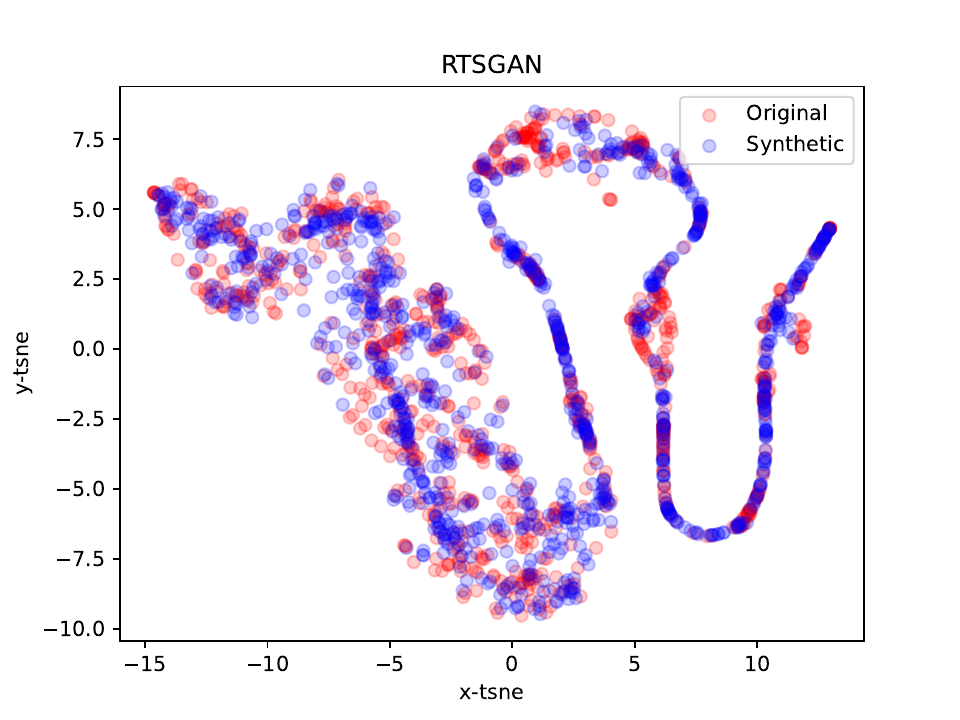}}%
\hfill
\subcaptionbox{\scriptsize COSCI-GAN}{\includegraphics[width=0.24\textwidth]{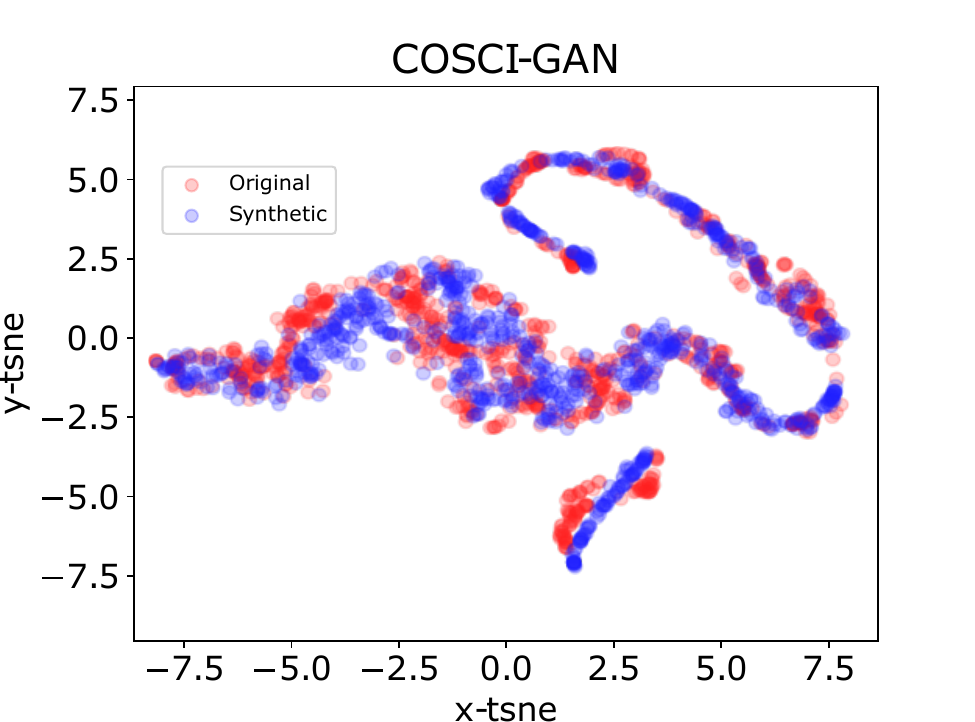}}%
\caption{A t-SNE visualizations of unconstrained time-series generation. Our models show among the best performance. }\label{fig:unconstrained_tsne_new_models}
\end{figure}

\begin{figure}[h!]
    \centering
    \includegraphics[width=1\textwidth]{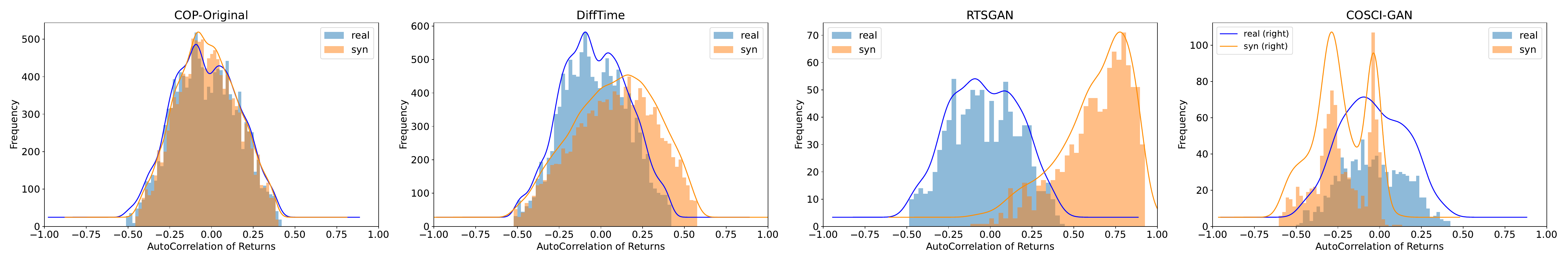}
    \caption{Autocorrelation of returns distribution for un-constrained time-series.}
    \label{fig:new_autoccr}
\end{figure}

We first focus on uncontrained time-series scenarios for multivariate stock-data. We report the quantitative metrics, Discr. and Pred. Score, in Table~\ref{tab:unconstrained_add} , while t-SNE analysis is shown in Figure~\ref{fig:unconstrained_tsne_new_models}. From the results, COP still shows the best distributional similarity w.r.t. to real data, which is empirically evaluated in the t-SNE plot, where blue and red dots almost always overlap. RTSGAN achieves notable performance in terms of Discr. and Pred. scores, with good distributional similarity in the t-SNE chart. However, with respect to properties pertinent to financial data introduced in Section~\ref{sec:fin-properties}, RTSGAN shows higher autocorrelation than real data, potentially stemming from multiple GRU layers (see Figure~\ref{fig:new_autoccr}); and more shallow return distribution.

\begin{table}[h]
    \caption{Unconstrained Time-Series Generation - Stock data}
    \label{tab:unconstrained_add}
    \centering
\begin{tabular}{l|llll}
\hline
Algo &  Discr-Score &   Pred-Score & Inference-Time \\  \hline
COP (Ours) & .050 ± .017 & .041 ± .001 & 1.01 ± 0.00   \\
DiffTime (Ours) & .097 ± .016 & .038 ± .001 & 0.02 ± 0.00  \\
COSCI-GAN      &  .412 ± .002 &  .088 ± .000 &  \textbf{0.00±0.00} \\
RTSGAN         &  \textbf{.024 ± .007} &  \textbf{.036 ± .000} &    \textbf{0.00±0.00} \\
\hline
\end{tabular}
\end{table}

\subsection{OHLC-Constrained Generation}
\begin{figure}[h!]
\subcaptionbox{\scriptsize COP}{\includegraphics[width=0.19\textwidth]{images/suppl/ohlc/tsne_COP.pdf}}%
\hfill
\subcaptionbox{\scriptsize Loss-DiffTime}{\includegraphics[width=0.19\textwidth]{images/suppl/ohlc/tsne_Loss-DiffTime.pdf}}%
\hfill
\subcaptionbox{\scriptsize Guided-DiffTime}{\includegraphics[width=0.19\textwidth]{images/suppl/ohlc/tsne_Guided-DiffTime.pdf}}%
\hfill
\subcaptionbox{\scriptsize RTSGAN}{\includegraphics[width=0.19\textwidth]{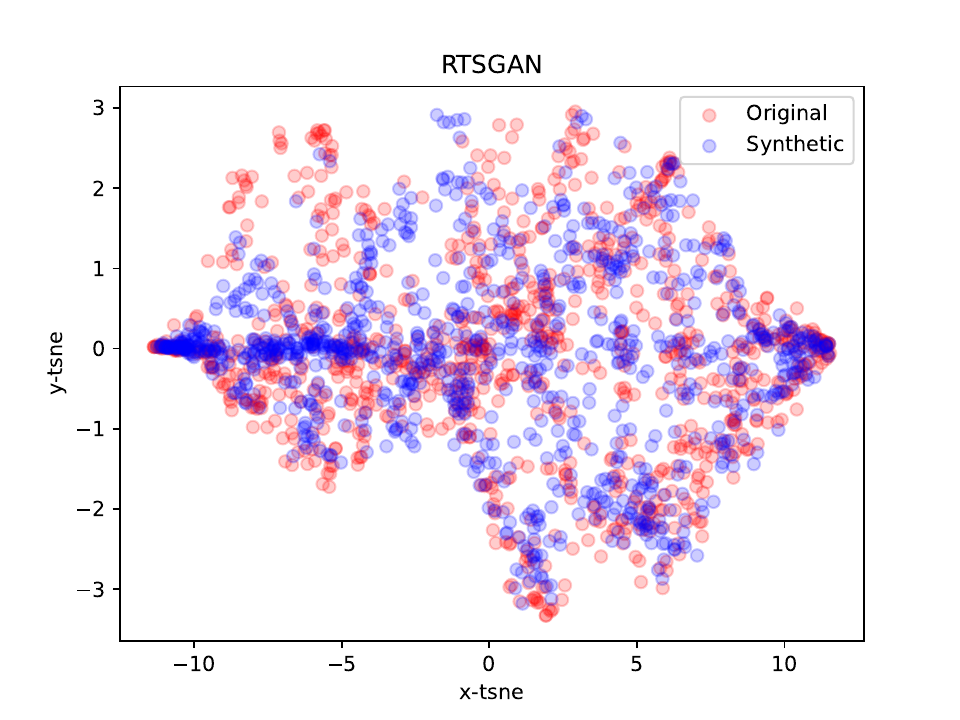}}%
\hfill
\subcaptionbox{\scriptsize COSCI-GAN}{\includegraphics[width=0.19\textwidth]{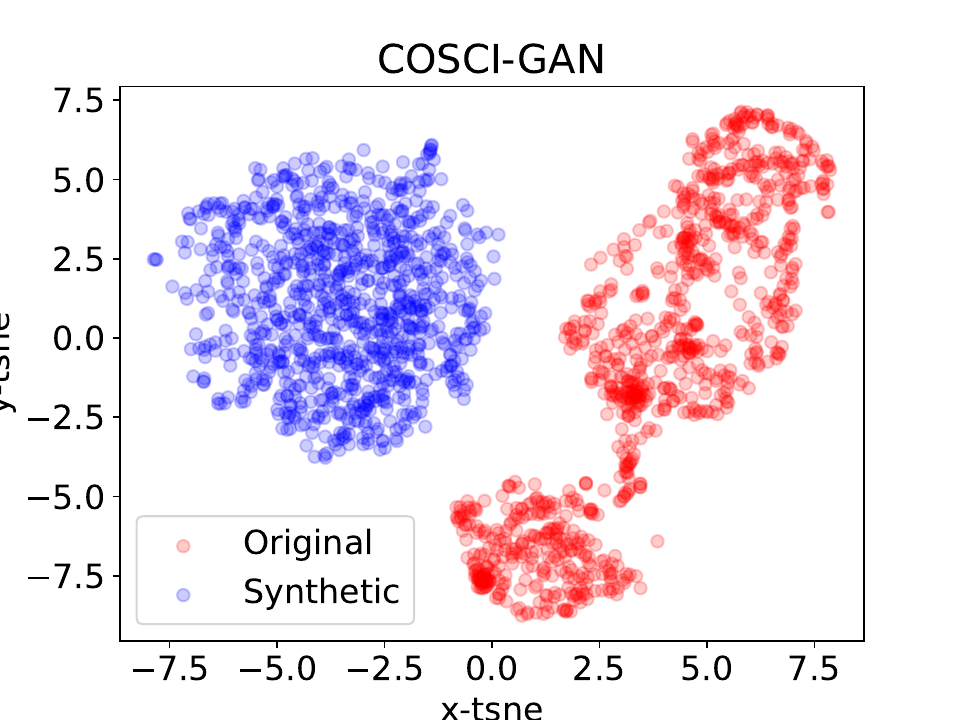}}%
\caption{A t-SNE visualizations of OHLC constrained time-series generation. Our models show among the best performance. }\label{fig:OHLC_constrained_tsne_new_models}
\end{figure}

We now focus on OHLC constrained time-series scenarios for multivariate stock-data. Table~\ref{tab:multivariate_add} shows the quantitative results. Figure~\ref{fig:OHLC_constrained_tsne_new_models} shows the distributional similarity of the new approaches, empirically evaluated through the t-SNE plot. From the results we can observe similar performance as in the unconstrained setting for COSCI-GAN and RTSGAN, both in terms of distributional similarity, discr. and pred. scores. However, looking at the satisfaction rate (i.e., percentage of time-series respecting the input constraint), our methods outperform the two benchmarks. Most importantly, our Guided-DiffTime model stands out for its remarkable capacity to accommodate new constraints without any retraining, constituting a fundamental innovative contribution to the literature on generating TS data.

\begin{table}[h!]
    \caption{OHLC Constrained - Stock data}
    \label{tab:multivariate_add}
        \centering
\begin{tabular}{l|llll}
\hline
Algo &  Discr-Score &   Pred-Score & Inference-Time & Satisfaction Rate \\        \hline
  COP (Ours) & 0.04 ± 0.02 & \textbf{0.04±0.00} & 2.17 ± 0.10 & \textbf{1.00±0.00}  \\
  GuidedDiffTime (Ours) & 0.08 ± 0.00 & \textbf{0.04±0.10} & 0.15 ± 0.00 &  0.72 ± 0.02  \\
  LossDiffTime (Ours) & 0.35 ± 0.04 & \textbf{0.04±0.01} & 0.14 ± 0.00 &  0.69 ± 0.01  \\
 COSCI-GAN &  0.45 ± 0.01 &  0.09 ± 0.00 &  \textbf{0.00±0.00} & 0.02 ± 0.00 \\
 RTSGAN &  \textbf{0.02±0.01} &  \textbf{0.04±0.00} & \textbf{0.00±0.00} & 0.54 ± 0.02 \\ \hline
\end{tabular}
\end{table}

\newpage

\section{Data Description}
We now report all the statistical properties of used datasets.

\begin{table}[h]
\caption{Stock Dataset (GOOG)}\label{tab:goog}
\vspace{0.2in}
\centering
\resizebox{1\linewidth}{!}{
\begin{tabular}{lrrrrrrr}
\toprule
Feature &          mean &         std &      min &         25\% &         50\% &         75\% &          max \\
\midrule
Open      &        453.23 &      305.02 &    49.27 &      233.25 &      306.95 &      621.22 &      1271.00 \\
High      &        457.33 &      307.45 &    50.54 &      235.40 &      309.35 &      627.55 &      1273.89 \\
Low       &        448.81 &      302.55 &    47.67 &      230.75 &      304.51 &      612.40 &      1249.02 \\
Close     &        453.15 &      305.13 &    49.68 &      233.44 &      306.44 &      622.69 &      1268.33 \\
Adj-Close &        453.15 &      305.13 &    49.68 &      233.44 &      306.44 &      622.69 &      1268.33 \\
Volume    &    7391935.77 &  8197565.12 &  7900.00 &  1959200.00 &  4674500.00 &  9723900.00 &  82768100.00 \\
\bottomrule
\end{tabular}
}
\end{table}

\begin{table}[h]
\centering

\caption{Synthetic Dataset (Sine)}\label{tab:sine_data}
\vspace{0.2in}
\begin{tabular}{lrrrrrrr}
\toprule
Feature &       mean &   std &  min &   25\% &   50\% &   75\% &  max \\
\midrule
Sine-1 &    0.49 &  0.32 &  0.0 &  0.20 &  0.46 &  0.80 &  1.0 \\
Sine-2 &    0.50 &  0.32 &  0.0 &  0.20 &  0.46 &  0.80 &  1.0 \\
Sine-3 &    0.50 &  0.32 &  0.0 &  0.19 &  0.46 &  0.81 &  1.0 \\
Sine-4 &    0.50 &  0.32 &  0.0 &  0.20 &  0.46 &  0.80 &  1.0 \\
Sine-5 &    0.49 &  0.32 &  0.0 &  0.19 &  0.46 &  0.80 &  1.0 \\
\bottomrule
\end{tabular}
\end{table}

\begin{table}[h]
\centering
\caption{Energy Dataset}\label{tab:energy}
\begin{tabular}{lrrrrrrr}
\toprule
Feature &       mean &     std &     min &     25\% &     50\% &     75\% &      max \\
\midrule
Appliances  &     97.69 &  102.52 &   10.00 &   50.00 &   60.00 &  100.00 &  1080.00 \\
lights      &      3.80 &    7.94 &    0.00 &    0.00 &    0.00 &    0.00 &    70.00 \\
T1          &     21.69 &    1.61 &   16.79 &   20.76 &   21.60 &   22.60 &    26.26 \\
RH\_1        &     40.26 &    3.98 &   27.02 &   37.33 &   39.66 &   43.07 &    63.36 \\
T2          &     20.34 &    2.19 &   16.10 &   18.79 &   20.00 &   21.50 &    29.86 \\
RH\_2        &     40.42 &    4.07 &   20.46 &   37.90 &   40.50 &   43.26 &    56.03 \\
T3          &     22.27 &    2.01 &   17.20 &   20.79 &   22.10 &   23.29 &    29.24 \\
RH\_3        &     39.24 &    3.25 &   28.77 &   36.90 &   38.53 &   41.76 &    50.16 \\
T4          &     20.86 &    2.04 &   15.10 &   19.53 &   20.67 &   22.10 &    26.20 \\
RH\_4        &     39.03 &    4.34 &   27.66 &   35.53 &   38.40 &   42.16 &    51.09 \\
T5          &     19.59 &    1.84 &   15.33 &   18.28 &   19.39 &   20.62 &    25.80 \\
RH\_5        &     50.95 &    9.02 &   29.82 &   45.40 &   49.09 &   53.66 &    96.32 \\
T6          &      7.91 &    6.09 &   -6.06 &    3.63 &    7.30 &   11.26 &    28.29 \\
RH\_6        &     54.61 &   31.15 &    1.00 &   30.02 &   55.29 &   83.23 &    99.90 \\
T7          &     20.27 &    2.11 &   15.39 &   18.70 &   20.03 &   21.60 &    26.00 \\
RH\_7        &     35.39 &    5.11 &   23.20 &   31.50 &   34.86 &   39.00 &    51.40 \\
T8          &     22.03 &    1.96 &   16.31 &   20.79 &   22.10 &   23.39 &    27.23 \\
RH\_8        &     42.94 &    5.22 &   29.60 &   39.07 &   42.38 &   46.54 &    58.78 \\
T9          &     19.49 &    2.01 &   14.89 &   18.00 &   19.39 &   20.60 &    24.50 \\
RH\_9        &     41.55 &    4.15 &   29.17 &   38.50 &   40.90 &   44.34 &    53.33 \\
T\_out       &      7.41 &    5.32 &   -5.00 &    3.67 &    6.92 &   10.41 &    26.10 \\
Press\_mm\_hg &    755.52 &    7.40 &  729.30 &  750.93 &  756.10 &  760.93 &   772.30 \\
RH\_out      &     79.75 &   14.90 &   24.00 &   70.33 &   83.67 &   91.67 &   100.00 \\
Windspeed   &      4.04 &    2.45 &    0.00 &    2.00 &    3.67 &    5.50 &    14.00 \\
Visibility  &     38.33 &   11.79 &    1.00 &   29.00 &   40.00 &   40.00 &    66.00 \\
Tdewpoint   &      3.76 &    4.19 &   -6.60 &    0.90 &    3.43 &    6.57 &    15.50 \\
rv1         &     24.99 &   14.50 &    0.01 &   12.50 &   24.90 &   37.58 &    50.00 \\
rv2         &     24.99 &   14.50 &    0.01 &   12.50 &   24.90 &   37.58 &    50.00 \\
\bottomrule
\end{tabular}
\end{table}

\end{document}